\documentclass[a4paper,11.5pt,tablecaptionabove]{scrreprt}
\usepackage[left= 2.5cm,right = 2.5cm, bottom = 4.3 cm]{geometry}
\usepackage[onehalfspacing]{setspace}
\usepackage{multirow}
\usepackage{multicol}
\usepackage{rotating}
\usepackage[table,xcdraw,dvipsnames]{xcolor}
\usepackage{listofsymbols}
\usepackage{tikz,pgfplots}
\usepackage[linesnumbered,ruled,vlined]{algorithm2e}
\usepackage{svg}
\usepackage{dirtytalk}

\usepackage[
	pdftitle={Deep Learning for Robust Moving Target Indication with a Moving Camera },
	pdfsubject={},
	pdfauthor={Markus Bosch},
	pdfkeywords={Deep Learning,MTI},	
	hidelinks
]{hyperref}

\usepackage[utf8]{inputenc}
\usepackage[english]{babel}
\usepackage[T1]{fontenc}
\usepackage{graphicx, subfig}
\graphicspath{{img/}}
\usepackage{fancyhdr}
\usepackage{lmodern}
\usepackage[printonlyused]{acronym}
\usepackage{amsfonts}
\usepackage{amsmath}
\usepackage{hyperref}
\usepackage[section]{placeins}
\usepackage{verbatim}
\usepackage{xr}
\usepackage{pdfpages}
\usepackage{longtable}
\usepackage{booktabs}
\usepackage{colortbl}%
  
\begin{document}

\pagestyle{empty}

\begin{center}
\begin{tabular}{p{\textwidth}}
\begin{center}
\includegraphics[scale=0.2]{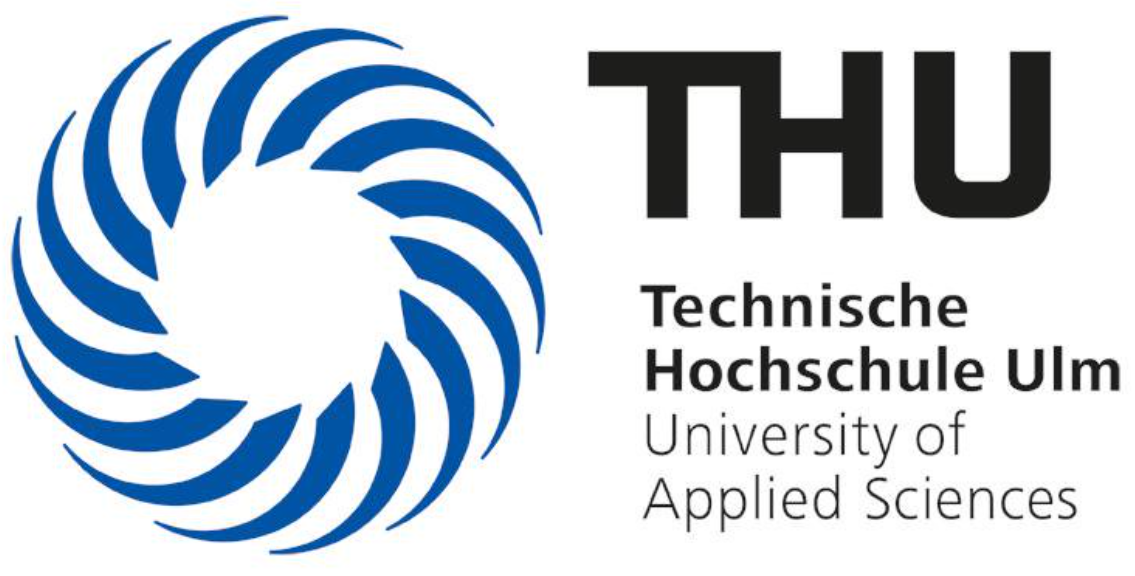}
\end{center}
\\
\begin{center}
\LARGE{\textsc{\textbf{Deep Learning for Robust Motion Segmentation with Non-Static Cameras \\
}}}
\end{center}

\\

\begin{center}
Ulm University of Applied Sciences \\
Department of Electrical Engineering and Information Technology\\

\Large{Systems Engineering and Management - Electrical Engineering}
\end{center}
\\

\begin{center}
\textbf{\LARGE{Master's Thesis}}
\end{center}

\begin{center}
in order to obtain the academic degree\\
Master of Engineering
\end{center}

\begin{center}
submitted by
\end{center}

\begin{center}
\large{\textbf{Markus Bosch}} \\
\end{center}

\begin{center}
in March 2020
\end{center}

\\

\\

\begin{center}
\begin{tabular}{lll}

\textbf{Examiner:} & &Prof. Dr.rer.nat. Roland Münzner\\
                   & &Prof. Dr.-Ing. Anestis Terzis\\
\textbf{Supervisor:} & &Dr.-Ing. Michael Teutsch\\
\end{tabular}
\end{center}

\end{tabular}
\end{center}

\newpage
\large{Version \today}

\vfill
\large{\textcopyright\ 2020 Markus Bosch} 

\newpage

\pagenumbering{Roman}
\setcounter{page}{1}

\pagestyle{fancy}

\addcontentsline{toc}{section}{Table of Contents}
\tableofcontents
\addcontentsline{toc}{section}{List of Figures}
\listoffigures
\addcontentsline{toc}{section}{List of Tables}
\listoftables
\thispagestyle{plain}
\chapter*{Acronyms}
\addcontentsline{toc}{section}{Acronyms}
\label{Abkvz}
\begin{acronym}[DCNN]
\acro{Adam}[Adam]{Adaptive Moment Estimation}
\acro{API}[API]{Application Programming Interface}
\acro{AUC}[AUC]{Area Under Curve}
\acro{AUROC}[AUROC]{Area under the Receiver Operating Characteristics}
\acro{CNN}[CNN]{Convolutional Neural Network}
\acroplural{CNN}[CNNs]{Convolutional Neural Networks}
\acro{ConvLSTM}[ConvLSTM]{Convolutional Long Short-Term Mermory}
\acro{CT}[CT]{Computer Tomography}
\acro{CUDA}[CUDA]{Compute Unified Device Architecture}
\acro{DCNN}[DCNN]{Deep Convolutional Neural Network}
\acroplural{DCNN}[DCNNs]{Deep Convolutional Neural Networks}
\acro{FCN}[FCN]{Fully Convolutional Networks}
\acro{FN}[FN]{False Negative}
\acroplural{FN}[FNs]{False Negatives}
\acro{FNR}[FNR]{False Negateive Rate}
\acro{FP}[FP]{False Positive}
\acroplural{FP}[FPs]{False Positives}
\acro{FPM}[FPM]{Feature Pooling Module}
\acro{FPR}[FPR]{False Positive Rate}
\acro{GPU}[GPU]{Graphics Processing Unit}
\acro{GT}[GT]{Groundtruth}
\acro{HDF}[HDF]{Hirarchical Data Format}
\acro{IOM}[IOM]{Intermitted Object Motion}
\acro{LSTM}[LSTM]{Long Short-Term Memory}
\acroplural{LSTM}[LSTMs]{Long Short-Term Memories}
\acro{MOD}[MOD]{Moving Object Detection}
\acro{ORB}[ORB]{Oriented Fast and rotated Brief}
\acro{PRE}[PRE]{Precision}
\acro{PTZ}[PTZ]{Pan-Tilt-Zoom}
\acro{PWC}[PWC]{Percentage of Wrong Classification}
\acro{ReLu}[ReLU]{Rectified Linear Unit}
\acroplural{ReLu}[ReLUs]{Rectified Linear Unit}
\acro{RGB}[RGB]{Red-Green-Blue}
\acro{RNN}[RNN]{Recurrent Neural Network}
\acroplural{RNN}[RNNs]{Recurrent Neural Networks}
\acro{ROC}[ROC]{Receiver Operating Characteristics}
\acro{ROI}[ROI]{Region of Interest}
\acro{ROOBI}[ROOBI]{Regions of Objects of Interest}
\acro{RPN}[RPN]{Region Proposal Network}
\acro{SDE}[SDE]{Scene Dependent Evaluation}
\acro{SIE}[SIE]{Scene Independent Evaluation}
\acro{SIFT}[SIFT]{Scale-Invariant Feature Transform}
\acro{TN}[TN]{True Negative}
\acroplural{TN}[TNs]{True Negatives}
\acro{TNR}[TNR]{True Negative Rate}
\acro{TPR}[TPR]{True Positive Rate}
\acro{TP}[TP]{True Positive}
\acroplural{TP}[TPs]{True Positives}
\acro{VGG}[VGG]{Video Geometry Group}
\acro{WAMI}[WAMI]{Wide Area Motion Imagery}
\acro{}[]{}
\end{acronym}

\addsec{Statement of Authorship}
\label{erklaerung}
\thispagestyle{plain}
I hereby declare that I am the sole author of this master thesis and that I have not used any sources other than those listed in the bibliography and identified as references. I further declare that I have not submitted this theses at any other institution in order to obtain a degree.\\

Oberkochen, \today \hspace{50pt} Markus Bosch: \hrulefill
\\[3.5cm]
\newpage
\thispagestyle{plain}

\addsec{Acknowledgements}

First I would like to thank Hensoldt Optronics, Prof. Roland Münzner and Prof. Anestis Terzis from Ulm University of Applied Sciences for the support and the opportunity of writing this thesis. Furthermore I would like to thank my supervisor Dr. Michael Teutsch for the possibility of writing this master's thesis and the valuable guidance and revealing discussions during the last six months. Finally I want to thank my family for the unfailing encouragement throughout my years of study.

\newpage

\newpage\null\thispagestyle{empty}\newpage
\pagenumbering{arabic}
\setcounter{page}{1}
\chapter*{Abstract}
\addcontentsline{toc}{chapter}{Abstract}
\label{abstract}

A key task in the field of computer vision applications for \mbox{(semi-)automated} visual surveillance exploitation is the detection and segmentation of moving objects in video sequences especially for non-static cameras. Besides assisting human operators in the task of visual surveillance, motion segmentation can be utilized as a pre-processing step for other computer vision tasks such as object tracking or human action recognition. Recently, \acp{DCNN} demonstrated their ability to represent and analyse image content, and therefore gained the interest of research since they surpassed the state-of-the-art in a wide field of challenges in computer vision. Motion segmentation approaches often struggle with frequent and continuous changes in the scene due to a panning, tilting or zooming camera. 
This work proposes a new end-to-end \ac{DCNN} based approach for motion segmentation, especially for video sequences captured with such non-static cameras, called MOSNET. While other approaches focus on spatial \cite{Lim_v2, LimS, Lim} or temporal \cite{Farneback, DISopticalflow} context only, the proposed approach uses 3D convolutions as a key technology to factor in, spatio-temporal features in cohesive video frames. This is done by capturing temporal information in features with a low and also with a high level of abstraction. The lean network architecture with about $21$k trainable parameters is mainly based on a pre-trained VGG-16 network. The MOSNET uses a new feature map fusion technique, which enables the network to focus on the appropriate level of abstraction, resolution, and the appropriate size of the receptive field regarding the input. Furthermore, the end-to-end deep learning based approach can be extended by feature based image alignment as a pre-processing step, which brings a gain in performance for some scenes. Evaluating the end-to-end deep learning based MOSNET network in a \ac{SIE} manner leads to an overall F-measure of $0.803$ on the CDNet2014 dataset. A small temporal window of five input frames, without the need of any initialization is used to obtain this result. Therefore the network is able to perform well on scenes captured with non-static cameras where the image content changes significantly during the scene.   
In order to get robust results in scenes captured with a moving camera, feature based image alignment can implemented as pre-processing step. The MOSNET combined with pre-processing leads to an F-measure of $0.685$ when cross-evaluating with a relabeled LASIESTA dataset, which underpins the capability generalise of the MOSNET.

\chapter{Introduction}
\label{Inroduction}

Due to the rapid growth of surveillance video data available nowadays, human operators can hardly keep track of all the contained information. The manual analysis and exploitation of such videos is an exhausting and error-prone work. Instead, the automatic detection and segmentation of motion in videos can be used to generate alarms that catch the operator's attention only in the few situations of noteworthy events. Motion segmentation can be used to detect motion in videos and thus can assist the operator in many tasks for \mbox{(semi-)automated} visual surveillance. In addition to static cameras, moving camera systems are increasingly used in the field of video surveillance. This is accompanied by new challenges for the detection and segmentation of moving objects, which need to become robust against camera movements resulting in image alignment issues and 3D parallax effects. Motion segmentation as an underlying technique for a large number of computer vision tasks in the field of intelligent video analysis has sparked a growing interest among researchers with 2,430 publications between 2008 and 2018 \cite{Kalsotra}. Its robust and pixel accurate functionality forms the basis for modern optronic camera systems and \mbox{(semi-)automated} visual surveillance.

\section{Motivation}
\label{IntroMotivation}

Motion segmentation aims to detect moving objects in video sequences and segment them pixel-precisely, while being independent of any assumptions about object categories or semantic labels in a class-agnostic manner. Nevertheless, motion segmentation is a challenging task since relevant motion coming from moving objects such as persons or vehicles needs to be distinguished from irrelevant motion originating from e.g. moving leaves or waves. The suppression of 3D parallax effects and image alignment errors for a moving camera is even more challenging.\\

State-of-the-art algorithms for motion segmentation are usually based on sparse optic flow, image alignment, subsequent frame differencing, and optional short-term object tracking. Such methods still suffer from weak contrast, low object velocity, or occlusions that make the relevant object motion disappear in the noise of irrelevant motion. Recently, it has been demonstrated in many computer vision tasks that the application of deep learning techniques, namely \acp{DCNN}, can be considered to surpass this state-of-the-art \cite{Lim_v2, Lim, LimS}. Many deep learning based approaches perform well on a specified subset of data but lack the capability to generalize the learned on unseen video sequences \cite{Braham, CascadeCNN}. In the course of this, the field of application of such methods is strictly limited to a static usage. In addition, some of the deep learning based approaches \cite{BSUV, 3DFR} require a host of input frames as a initialization before outputting a result, which brings in a delay and restricts the performance when the camera is panning, tilting or zooming.\\ 

The deep learning based approach, to be developed in this work aims to avoid those drawbacks by simultaneous reaching the state-of-the-art performance. 

\section{Contributions}
\label{Contribution}

This section points out the main contributions that have been achieved by this thesis:

\begin{itemize}
    \item End-to-end deep learning method
    \item Scene independent method
    \item Minimum temporal window
    \item Improvement of handling objects of different size
    \item LASIESTA dataset for motion segmentation
\end{itemize}

This work introduces a new approach for motion segmentation, which is end-to-end deep learning based, and therefore does not require any pre- or post-processing.
Furthermore, this approach aims to learn frame differencing, and therefore is not based on a previous learned background model. This enables the approach to be scene independent, which is demonstrated by performing a scene independent evaluation and a cross-evaluation. This distinctive ability to generalize makes the approach applicable to a broader field of motion segmentation applications than methods based on a background model. 
By processing only five input frames, the developed approach succeeds in adapting quickly to changing image contents without any need of further initialization frames, which is advantageous especially  for non-static cameras.
By combining several levels of abstraction, several resolutions, and considering different sized areas in image space, the developed network is able to find an appropriate trade-off of these parameters itself, which enhances the handling of objects of different size.
In order to cross-evaluate the proposed approach, the LASIESTA dataset got relabeled to represent the task of motion segmentation. 

\section{Thesis Organization}
\label{Organization}

In Chapter \ref{TechnicalFundation} a technical basis, and the nomenclature for this work is created by covering the basics, and some advanced methods of deep learning for computer vision. Chapter \ref{RelatedWork} points out the major challenges of motion segmentation tasks, and reveals the state-of-the-art leading into design constraints for the targeted approach. Thereafter, three datasets used in this thesis, namely CDNet2014, DAVIS 2016 and LASIESTA, get introduced and investigated. In Chapter \ref{Architecture} the deep learning network architecture developed in this thesis gets explained in detail, using the fundamentals of Chapter \ref{TechnicalFundation}. The developed process of training the network gets introduced in Chapter \ref{training}. In the next Section, experiments and results that have led to the final architecture of the deep learning network are presented. Further experiments underpin design decisions, and point out their gain in performance. Afterwards, in Chapter \ref{Eval} the developed approach gets brought in line with the state-of-the-art, surveyed in Chapter \ref{RelatedWork}. Chapter \ref{Conclusion} sums up the contributions and findings of this work by simultaneously exhibiting future work.  
\chapter{Technical Foundations}
\label{TechnicalFundation}

This chapter sets a technical basis and introduces the nomenclature used in this work.
Principles of deep learning get discussed in Section \ref{DeepLearning}, followed by neural network primitives (see Section \ref{DLPrimitives}) which are used to form the network architecture of the MOSNET in Section \ref{Architecture}.
Section \ref{DatasetBasics} sets the basis for Chapter \ref{Datasets}, by introducing principal foundations of datasets used for deep learning. In Section \ref{EvalMetrics}, common evaluation metrics are introduced, used to evaluate the performance of the MOSNET in Chapter \ref{Eval}.

\section{Deep Learning -- A Brief Overview}
\label{DeepLearning}

Deep Learning as a kind of machine learning is based on artificial neural networks. Whilst most of the machine learning methods are grounded on hand-crafted features, the benefit of deep learning methods is the network extracting its required features itself. This is a key attribute when it comes to computer vision tasks, since the feature engineering needed in most machine learning methods is challenging and error-prone but at the same time decisive for a good performance of the network. In addition, due to the deep network architecture, a deep level of abstraction between input and output data can be achieved, which enables the solution of complex computer vision problems.\\ 

The most common type of deep learning methods are the supervised methods, where the network calculates a objective function $\mathcal{L}$ using the predicted output of the neural network and the desired output, provided with the dataset (see Section \ref{Datasets}). This objective function represents an inaccuracy of the predicted output, called loss, and maps it on a multi-dimensional function of all weights and biases implemented in the network. During training, the aim of the neural network is to minimize the objective function. This is done by using numerical algorithms to calculate the gradient (\cite{Goodfellow}, pp. 290-307) and make use of the backpropagation algorithm to adjust the weights and biases of each neuron in the network using the chain rule (\cite{Goodfellow}, pp. 200-220). To validate the training process, an iterative validation step can be performed. Besides using disjoint data, the key point between training and validation is the disuse of backpropagation during validation.\\

Before training a deep learning network, a set of variables called hyperparameters must be defined to configure the network and the training process, respectively. Besides the learning rate, which represents the step size of the numerical optimization algorithm, the mini-batch size is a fundamental hyper-parameter in most deep learning approaches. To save computational cost, the data used for training gets sliced into multiple mini-batches. At each training step, the \ac{DCNN} gets fed with one mini-batch, until the network was trained on all mini-batches once. In the scope of machine learning, this is called an epoch.  

\subsection{Artificial Neurons}

A single artificial neuron can be defined with Equation \eqref{eq:neuron}, where $\hat{y}(n)$ is the output of the neuron at time-step $n$ and $x_i(n)$ is the i-th input of the neuron at time-step $n$. Each input gets multiplied with a weight $w_i(n)$ which is the i-th weight at time-step $n$. After summing up all weighted inputs (where $i$ ranges from 0 to $m$) and adding a bias $b$, the result gets mapped by an activation function $F$ in between $0$ to $1$ or $-1$ to $1$ respectively (see Section \ref{activationfunctions}). 

\begin{equation}
\hat{y}(n) =F \Biggl(\sum_{i=0}^m w_i(n) \cdot x_i(n) +b\Biggl)
\label{eq:neuron}
\end{equation}
\FloatBarrier

Neurons are combined in layers, which in turn are concatenated and combined in order to form a network structure. The weights and biases of a network represent its parameters. A feedforward structure consists of several layers, where the signal flow of all neurons is in the same direction as further discussed in Section \ref{DLPrimitives}.

\subsection{Activation Functions}
\label{activationfunctions}

The summed weighted inputs and biases $z = \sum_{i=0}^m w_i(n) \cdot x_i(n) +b$ can perform a linear tasks, but make it impossible for the network to solve non-linear transformations. In order to perform non-linear transformations, $z$ gets mapped by a non-linear activation function $F$ on a range between $0$ to $1$ or $-1$ to $1$, depending on the activation function (see Appendix \ref{App:Activation}).\\  
\pagebreak
Some common activation functions used in practice are (\cite{Goodfellow}, pp.177-193):

\begin{itemize}
    \item Sigmoid
    \item Hyperbolic Tangent
    \item Softmax
    \item Rectified Linear Unit
\end{itemize}{}
The output of the sigmoid function is often used for binary classification problems, since its derivative has its maximum at $z = 0$ which pushes the weights and biases to produce an output $\sigma(z)$ being either close to one or close to zero (\cite{Goodfellow}, pp.178-180). One drawback of the sigmoid function is its non-zero-centered characteristics, which gets fixed by the hyperbolic tangent (\cite{Goodfellow}, pp.191-192). The softmax function is used to tackle multi-class problems since its outputs can be interpreted as probability of mutually exclusive classes (\cite{Goodfellow}, pp.180-184). The most widely spread activation function used in practice is the \ac{ReLu} function. The \ac{ReLu} function sets values $z \leq 0$ to $0$ which results in not all neurons of a layer getting activated at the same time. It turned out that this improves the learning of a neural network and outperforms the previous mentioned activation functions (\cite{Fundamentals}, pp.13-15). The plotted sigmoid, hyperbolic tangent, and \ac{ReLu} activation function is appended in Appendix \ref{App:Activation}.\\
The choice of the activation function used for each neuron respectively layer is significant for the performance of the network. While the last activation function of a network is mainly specified by the task the network is addressed (e.g. whether a binary or multi-class problem is solved), the design of the intermediate activation functions is not fixed by the task of the network. \emph{\say{Deep convolutional neural networks with \acp{ReLu} train several times faster than their equivalents with \textit{tanh} units}\footnote{tanh is short for hyperbolic tangent}} \cite{Krizhevsky}. Following Krizhevsky et al. \cite{Krizhevsky}, all intermediate layers of the MOSNET use the \ac{ReLu} activation functions, which is further discussed in Section \ref{Architecture}.

\subsection{Regularization}
\label{regularization}

\begin{figure}[h!]
\centering
\includegraphics[width=0.55\textwidth]{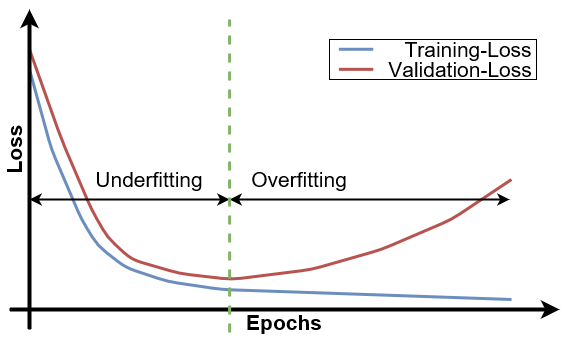}
\caption [Schematic illustration of over- and underfitting regarding the training- and validation-loss]{Schematic illustration of over- and underfitting regarding the training- and validation-loss.}
\label{fig:Overfitting}
\end{figure}

A central goal in machine learning is to achieve generalization ability. To be precise, the network must have the ability to abstract what has been learned and perform well on unseen data of the same domain. If the network is overfitted, it shows good performance on the previously observed training data but can not generalize well.\\ 
Figure \ref{fig:Overfitting} shows the training- and validation-loss \footnote{The training loss is calculated using the training data, while the validation loss gets calculated using the validation data} over the training epochs. The dotted green line marks the optimum time for stopping the training in order to achieve maximal generalization ability on unseen data, represented by the validation data. Stopping the training, while training- and validation-loss is still decreasing leads to an underfitted model. If the validation-loss increases, the model overfits and therefore loses its ability to perform on previously unseen data. 

A detailed explanation of over- and underfitting and ways to overcome both is given in \cite{Goodfellow}, pp. 108-114, pp. 224-270. One common way of tackling overfitting is called regularization. \emph{\say{Regularization is any modification we make to a learning algorithm that is intended to reduce its generalization error but not its training error}} (\cite{Goodfellow} p.117). The regularization methods \textit{Dropout} and \textit{Early Stopping} are discussed in Section \ref{dropout}, and Section \ref{monitoring}, respectively.\\

Another regularization method is the parameter norm penalty (\cite{Goodfellow} p. 226), where specific weights get a norm penalty term added (e.g. the $\ell_2$-norm), in order to drive their values closer to the origin \cite{Zeeshan}. This regularizes the effect of the specific weight not getting to high. The $\ell_2$-norm is used to regularize some kernel weights of the MOSNET, further discussed in Section \ref{Decoder}.

\pagebreak
\section{Neural Network Primitives}
\label{DLPrimitives}

The architecture of neural networks is usually formed by combining a selection of neural network layers, called neural network primitives. The ones used in this work are introduced in the following sections.  

\subsection{Convolutional Layers}
\label{convolutionallayer}

\acp{CNN} are neural networks that contain at least one convolutional layer, while this can be a one-, two-, or multidimensional discrete convolution. Besides the ability of being able to handle inputs of variable size, convolutional layers have their advantages in their sparse interactions, the parameter sharing and the equivariant representations, which is explained in detail in \cite{Goodfellow}, pp. 329-335.\\

Essential parameters of convolutional layers are the (filter-) kernel size and the number of (filter-) kernels also called filters. The kernel size describes the shape of the convolution matrix, which is slided over the input, while the number of filters describes how many convolutional matrices are used, and therefore how many convolutions are performed. Each filter has its exclusive weights and generates a so called feature map consisting of spatially local\footnote{In the case of a 2D convolution, spatial features get detected. The use of 3D convolutions enables the detection of spatio-temporal features, further discussed in Section \ref{3DConvGrundlagen}}, semantically related features found by each filter (\cite{Fundamentals}, pp.90-95). Each filter produces a different feature map including the features learned during training. All feature maps, get stacked along the depth dimension to form an output volume of the convolution layer. In example applying a convolution with $64$ filters on an image with three channels (e.g. an image using the \ac{RGB} color space), the output of the convolution layer has a depth of $64$. Another parameter of convolution layers is the stride, which represents the step size when sliding the evaluation matrix over the input.\\

\begin{figure}[h!]
\centering
\includegraphics[width=\textwidth]{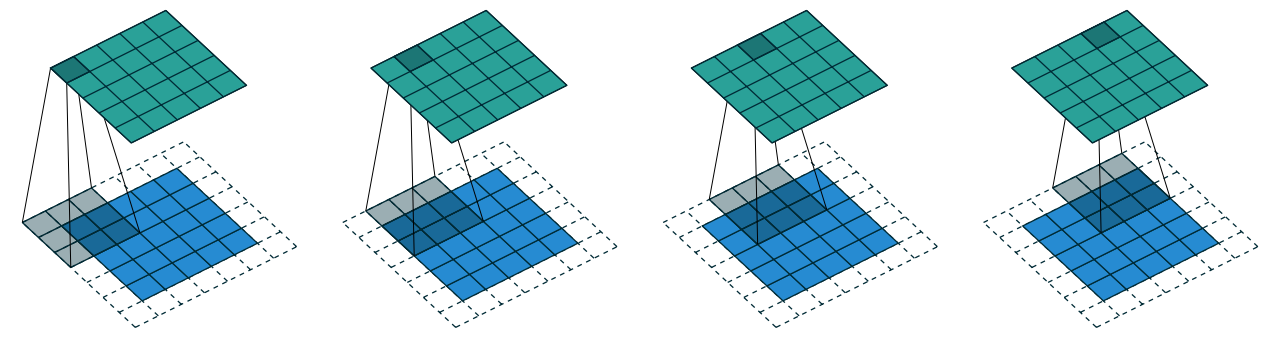}
\caption [Schematic illustration of the convolution operation with padding]{Schematic illustration of the convolution operation with padding, a kernel size of $3 \times 3$ and a stride of one. \cite{ConvExplained}}
\label{fig:ConvExplained}
\end{figure}

Figure \ref{fig:ConvExplained} illustrates four steps of the convolution operation, while the dotted fields are called padding. These fields are often set to zero and have the effect of keeping the dimensions of input (blue) and output (green) the same. The kernel size in Figure \ref{fig:ConvExplained} is $3 \times 3$ and the stride is one.\\   

\begin{figure}[h!]
\centering
\includegraphics[width=0.5\textwidth]{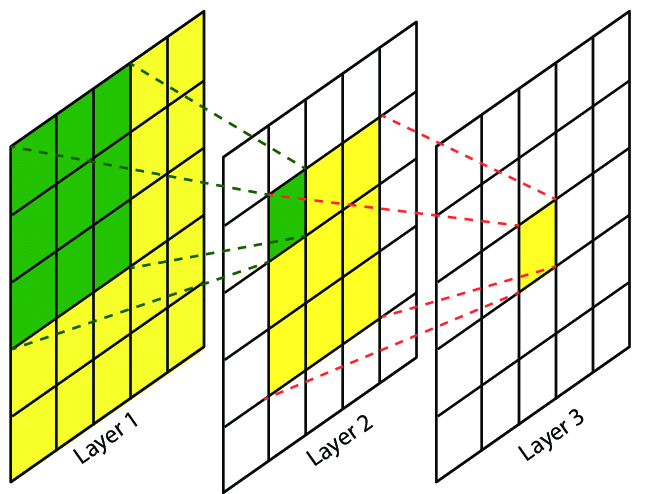}
\caption [Illustration of the evolution of the receptive field]{Illustration of the evolution of the receptive field in an three-layer 2D convolutional network. The yellow pixels mark the receptive field of Layer 3, while the receptive field of Layer 2 is marked in green \cite{receptiveimage}.}
\label{fig:receptiveimage}
\end{figure}

While in fully-connected layers (see Section \ref{fullyllayer}) every neuron interacts with every input, convolutional layers can sparse interactions by implementing a kernel size smaller than the input size. This enables the detection of spatially local, semantically related structures in the input data, called local features, e.g. corners, edges or texture in an image. These sparse interactions between neurons lead to a limited receptive field of each neuron. A receptive field indicates, which inputs of previous layers, or of the input data are able to affect the value of the neuron. Figure \ref{fig:receptiveimage} shows the evolution of a three-layer network of 2D convolutions with a kernel size of $3 \times 3$ and a stride of one in both dimensions. The yellow pixels in Figure \ref{fig:receptiveimage} mark the receptive field of Layer 3, while the receptive field of Layer 2 is marked in green. A closer look on receptive fields of convolution layers and fully-connected layers (see Section \ref{fullyllayer}) is given in in \cite{Goodfellow} pp. 331-332. The receptive field is an important characteristic of a neural network layer, which is further discussed in Section \ref{EC2}.

\FloatBarrier

Convolution layers, performing a 2D convolution get denoted with Conv2D in further sections.

\subsubsection{3D Convolution}
\label{3DConvGrundlagen}

The convolution matrices of a 2D convolution get slid along two dimensions of the input data, while the convolution matrices of a 3D convolution get slid along three dimensions of the input data. This is often used to tackle computer vision tasks in volumetric data such as \ac{CT} scans or multi-channel data such as color video data (\cite{Goodfellow}, p.355). Previous deep learning approaches \cite{LaLonde, Turki, 3DFR} used the 3D convolution to capture spatial and temporal components in video sequences. In this case, the convolution matrices get slid along two spatial dimensions of the image space and on the temporal dimension in a sequence of images. This is a method of choice when it comes to capturing spatio-temporal features in video sequences, and further discussed in Section \ref{LLConv3D} and Section \ref{HLConv3D}. Convolution layers, performing a 3D convolution get denoted with Conv3D in further sections. 

\subsection{Transposed Convolution Layers}
\label{ConvT}
\emph{\say{The need for transposed convolutions generally arises from the desire to use a transformation going in the opposite direction of a normal convolution, i.e., from something that has the shape of the output of some convolution to something that has the shape of its input while maintaining a connectivity pattern that is compatible with said convolution}} \cite{ConvExplained}. Convolution and pooling layers (see Section \ref{poolinglayer}) are used in the encoder parts (see Section \ref{EC1} and Section \ref{EC2}) of the MOSNET to generate a more abstract representation of the input by simultaneously reducing the dimensions of the input. Since the outputted motion masks of the MOSNET shall have the same dimensions as the input frame to evaluate in a pixel-precise manner, the abstract representation must be up-scaled in some way. Transposed convolutions perform a inverse convolution and are one way to up-scale feature maps. A closer look on the arithmetic of transposed convolutions is given in \cite{ConvExplained}. A set of 2D transposed convolutions is performed in the decoder part of the MOSNET in Section \ref{Decoder}. Convolution layers, performing a transposed 2D convolution get denoted with Conv2DT in further sections. 

\subsection{Fully-Connected Layers}
\label{fullyllayer}

In fully-connected layers, every input is directly connected with every output, which leads to every input being able to affect every output. Therefore fully-connected layers have significantly more weights and biases than convolutional layers, but are not dependent on local structures since their receptive field is the same size as the input of the layer. Besides in fully-connected neural networks, fully-connected layers are often used in \acp{CNN}, e.g., for combining local features produced by previous convolutional layers to make a decision about an image in its entirety. The VGG-16 network, introduced in Section \ref{VGG-16} uses fully-connected layers in the last three layers of the network in order to tackle an image labeling problem.   

\subsection{Pooling Layers}
\label{poolinglayer}

\begin{minipage}{0.5\textwidth}
Pooling layers combine nearby inputs in order to be invariant to small translations and to reduce the dimensionality of feature maps. \emph{\say{Invariance to local translation can be a useful property if we care more about whether some feature is present than exactly where it is}} (\cite{Goodfellow}, p. 336). The most common used pooling strategy is the \textit{Max Pooling}, where the maximum value of the receptive field of the pooling layer (defined by the kernel size) is carried over. Figure \ref{fig:Maxpooling} illustrates a Max Pooling operation with a kernel size of $2 \times 2$ and a stride of two in both dimensions, whereby the maximum value is taken over. The Max Pooling, and other pooling strategies are explained in-depth in \cite{Goodfellow}, pp.335-341. Max pooling is used as part of the MOSNET in order to reduce the dimensions of feature maps. Layers, performing the Max Pooling operation get denoted with MaxPooling in further sections. 
\end{minipage}
\hfill
\begin{minipage}{0.4\textwidth}
\begin{center}
\includegraphics[width=0.5\textwidth]{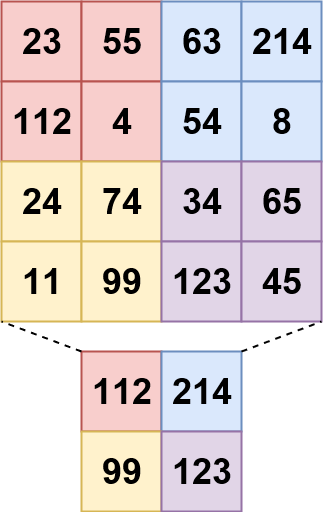}
\captionof{figure}[Schematic illustration of Max Pooling]{Schematic illustration of Max Pooling with a kernel size of $2 \times 2$ and a stride of two in both dimensions.}
\label{fig:Maxpooling}
\end{center}
\end{minipage}
\FloatBarrier

\subsection{Dropout}
\label{dropout}

\begin{minipage}{0.5\textwidth}
A way of regularizing the network and preventing overfitting is the use of \textit{Dropout}. 
\emph{\say{The  key  idea  is  to  randomly  drop  units  (along  with  their  connections)  from  the  neural network during training.  This prevents units from co-adapting too much}} \cite{Dropout}. Figure \ref{fig:Dropout} illustrates the Dropout operation, with a fully-connected neural network on the left, and the network after applying Dropout on the right. In a Dropout layer, a random set of neurons are not activated during one training step. The percentage of neurons, not getting activated is referred to as Dropout rate.
The use of Dropout in the architecture of the MOSNET in order to tackle overfitting is discussed in Section \ref{EC2}. 
\end{minipage}
\hfill
\begin{minipage}{0.4\textwidth}
\includegraphics[width=\textwidth]{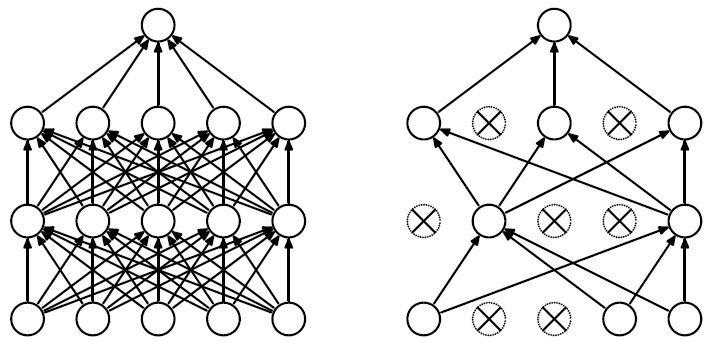}
\captionof{figure}[Schematic illustration of Dropout]{Schematic illustration of Dropout. Fully-connected neural network before (left) and after (right) applying Dropout \cite{Dropout}.}
\label{fig:Dropout}
\end{minipage}
\FloatBarrier

\subsection{Batch Normalization}
\label{BatchNorm}

In 2015, Ioffe and Szegedy \cite{batchnorm} introduced a novel technique to accelerate the learning speed of deep neural networks, called \textit{Batch Normalization}. It normalizes the inputs of the layers by subtracting the mean of each mini-batch and divides by its standard deviation in order to reduce internal covariate shift \cite{batchnorm}. Ioffe and Szegedy also denoted a regularization effect of adding Batch Normalization to a deep neural network, which is taken into account when implementing the Dropout rate in Section \ref{EC2}. Layers, performing the Batch Normalization operation get denoted with BatchNorm in further sections.

\section{VGG-16}
\label{VGG-16}

In 2014, the \ac{VGG} at Oxford University introduced the VGG-16 network \cite{Simonyan}, trained on the large-scale dataset ImageNet \cite{ImageNet}. Besides beating all state-of-the-art methods on the Large Scale Visual Recognition Challenge 2014, the network got popular since its trained weights and biases were made publicly available. This results in many deep learning methods in the field of computer vision using the network as backbone for their approaches in order to decrease the time used for training significantly.\\

\begin{figure}[h!]
\centering
\includegraphics[width=0.9\textwidth]{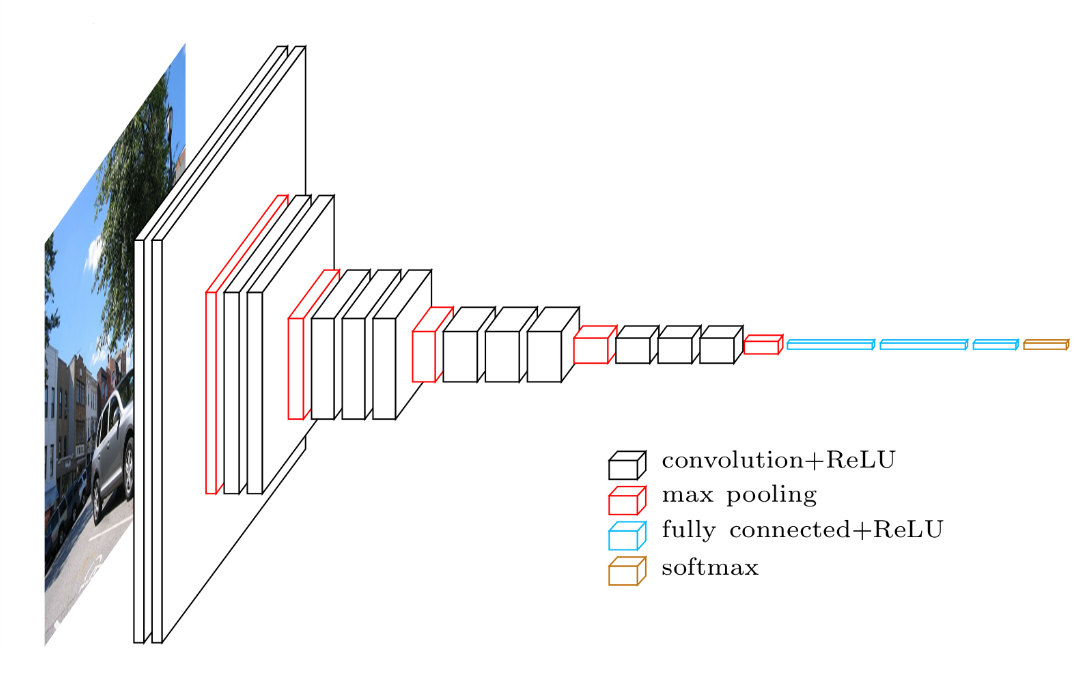}
\caption[Architecture of the VGG-16 network]{Architecture of the VGG-16 network consisting of 13 convolutional layers, three fully connected layers and five Max Pooling layers (adapted from \cite{VGG16Image}).}
\label{fig:VGG16Image}
\end{figure}

Some of the basic layers used in feedforward networks like convoluational layers, fully-connected layers and pooling layers introduced in previous sections are combined in order to build the VGG-16 network architecture.
Figure \ref{fig:VGG16Image} shows the principal architecture of the VGG-16 network consisting of 13 convolutional layers with a kernel size of $3 \times 3$ and a number of filters ranging from $64$ to $512$. Also five Max Pooling layers with a kernel size of $2 \times 2$ and three fully-connected layers followed by a softmax activation are implemented. As mentioned in Section \ref{activationfunctions}, due to their gain in performance all intermediate layers use \ac{ReLu} as an activation function. A closer look into the architecture is given in Section \ref{EC1} and Section \ref{EC2} since the VGG-16 is partly used in the encoder parts of the MOSNET.  

\FloatBarrier

\section{Dataset Basics}
\label{DatasetBasics}

\subsection{Dataset Split}
\label{Datasetsplit}

Datasets used for supervised learning consist of input data, e.g., video frames and the corresponding desired output data, e.g., motion masks, called \ac{GT}. The three major phases of developing a neural network are training, validation and evaluation, where each of the three phases requires a disjoint set of data, representative for the task it addresses. The data $D_T$ used for training the network, the data $D_V$ used for validating the training process and the data $D_E$ for evaluating the trained network with $D_T \cap D_V \cap D_E = \emptyset$ can originate of the same dataset\footnote{An exception is the cross-evaluation, where $D_T$ and $D_E$ originate of different datasets, which is elaborated in Section \ref{ExpCrossEval}.}.\\ 

All three splits $D_T$, $D_V$ and $D_E$ of the dataset should be representative for the whole dataset, in order to prevent a corruption due to overfitting in one of the three phases. The choice of the dataset and its split has a strong impact on the performance of the approach. That's why this work distinguishes between \ac{SDE}, where $D_T$ and $D_E$ originate of the same video sequence, and \ac{SIE}, where $D_T$ and $D_E$ do not originate of the same video sequence. More details about the evaluation methods \ac{SDE} and \ac{SIE} are given in Chapter \ref{Eval}.\\

In order to monitor the training process and prevent the network from overfitting, a loss can be calculated on the training data $D_T$ and the validation data $D_V$. During the training, with the training data $D_T$, the weights and biases get adjusted, while during validation with the validation data $D_V$ the weights and biases of the network stay the same. Therefore, multiple validations can be performed using the same data $D_V$ without corrupting the training process. After developing the deep learning based approach, the evaluation data $D_E$ is used to predict outputs and then evaluate them.

\subsection{Annotation}
\label{Annotation}

Figure \ref{img:GroundTruthExample} shows an example \ac{GT} of the CDNet2014 dataset, which is discussed in Section \ref{CDNet2014}. Each pixel is annotated to be either static, hard shadow, outside \ac{ROI}, unknown or motion. \\

\begin{minipage}{0.4\textwidth}
\includegraphics[width=\textwidth]{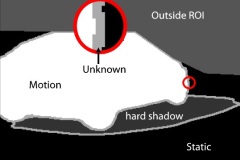}
\captionof{figure}[Groundtruth example of the CDNet2014-dataset]{Annotation example of a \ac{GT} image, showing the different classes, which are labeled in the CDNet2014 dataset \cite{Wang}.}
\label{img:GroundTruthExample}
\end{minipage}
\hfill
\begin{minipage}{0.55\textwidth}
Unknown pixels usually occur around moving objects, since there can be no classification into static or motion due to semi-transparency and motion blur. Pixels labeled as being outside \ac{ROI} mark pixels in the input frame, which are not of interest and therefore should not be considered when training, validating or evaluating the motion detection method.\\
Additionally to the labels discussed, multi object video segmentation datasets like LASIESTA introduced in Section \ref{LASIESTA} distinguish between the occurring moving object instances of the same class in one scene by using additional labels. The handling of these labels in order to use the dataset for motion segmentation is performed in Section \ref{LASIESTA}.    
\end{minipage}

\section{Evaluation Metrics}
\label{EvalMetrics}

The evaluation metrics used in this work are adopted by the CDNet2014 challenge, which in turn enables the MOSNET to be comparable to state-of-the-art methods.

\subsection{Confusion Matrix}
\label{sec:ConfusionMatrix}

The confusion matrix is used for performance measurement of discrete binary classification tasks.
The $2\times2$ matrix entails all four comparison results of a binary classification with the predicted output (i.e. produced by a neural network) and the actual output (i.e. provided with a dataset).\\\\
The evaluation metrics introduced in Section \ref{sec:F1score} are based on the following primitives of the confusion matrix:\\

\begin{minipage}{0.55\textwidth}
\begin{itemize}
    \item \textbf{TP} - True Positive
    \item \textbf{TN} - True Negative
    \item \textbf{FP} - False Positive
    \item \textbf{FN} - False Negative
\end{itemize}

Regarding the problem adressed in this work, \acp{TP} are predicted moving pixels, which are actual moving pixels in the \ac{GT} of the dataset and \acp{TN} are predicted static pixels, which are actual static pixels in the \ac{GT} of the dataset. Analogously to this, \acp{FP} are predicted moving pixels, which are actual static pixels and  \acp{FN} are predicted static pixels, which are actual moving pixels. Figure \ref{fig:ConfusionMatrix} shows a tabular view of the confusion matrix.
\end{minipage}
\hfill
\begin{minipage}{0.35\textwidth}
\includegraphics[width=\textwidth]{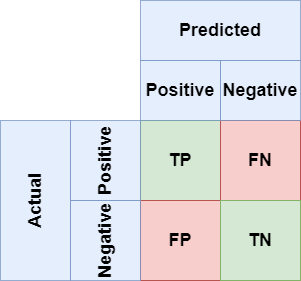}
\captionof{figure}[Confusion Matrix]{Tabular view of the confusion matrix.}
\label{fig:ConfusionMatrix}
\end{minipage}
\newline
\newline

To obtain relative values, the evaluation measures \ac{TPR}, \ac{TNR}, \ac{FPR} and \ac{FNR} are introduced in Equation \eqref{eq:TPR} to Equation \eqref{eq:FNR}:

\begin{equation}
TPR = \frac{TP}{TP + FN}
\label{eq:TPR}
\end{equation}

\begin{equation}
TNR = \frac{TN}{TN + FP}
\label{eq:TNR}
\end{equation}

\begin{equation}
FPR = \frac{FP}{TN + FP}
\label{eq:FPR}
\end{equation}

\begin{equation}
FNR = \frac{FN}{TP + FN}
\label{eq:FNR}
\end{equation}

\subsection{Derived Evaluation Metrics}
\label{sec:F1score}

For a better measurability of the performance of different approaches, the following metrics are introduced, based on the primitives of the confusion matrix:

\subsubsection{Precision}

The precision (Equation \eqref{eq:PRE}) indicates, how many of the predicted moving pixels are actual moving. More precisely, it is the fraction of the \acp{TP} among all predicted pixels:  

\begin{equation}
PRE = \frac{TP}{TP + FP}
\label{eq:PRE}
\end{equation}

\subsubsection{Recall}

The recall indicates how many of the actual moving pixels are predicted as being in motion. It is the fraction of the \acp{TP} among all actual moving pixels. Therefore, it is identical with the \ac{TPR} introduced in Equation \eqref{eq:TPR}.  

\subsubsection{Percentage of Wrong Classification}

The \ac{PWC}, as calculated in Equation \eqref{eq:PWC}, indicates the percentage of wrong classified pixels among all pixels.

\begin{equation}
PWC = \frac{100\cdot(FP + FN)}{TP + TN + FP + FN}
\label{eq:PWC}
\end{equation}

\subsubsection{F-Measure}

The most frequently used and commonly accepted evaluation metric for motion segmentation, background subtraction and change detection is the F-measure as shown in Equation \eqref{eq:F1}. It is the harmonic mean of precision and recall. Since the F-measure focuses on the \acp{TP}  and does not take the \acp{TN} into consideration, it is preferable in an imbalanced binary classification task, where there are only a few positive (i.e. moving) pixels and an overwhelming majority of negative (i.e. static) pixels. The F-measure ranges from zero to one, while it has its best value at one and worst at zero.  

\begin{equation}
\text{F-measure} = \frac{2}{TPR^{-1} + PRE^{-1}} = \frac{2 \cdot TPR \cdot PRE}{TPR + PRE}
\label{eq:F1}
\end{equation}

{}

\subsubsection{Precision-Recall Curve}

The precision-recall curve is the method of choice when it comes to performance measurement of binary classification tasks, when there is a strong imbalance in the observation of each class \cite{Saito}. In the case of a motion segmentation, there are usually many static pixels, but just a few moving ones. In the precision-recall curve, only the minority class of moving pixels are taken into consideration. Like the F-measure (Equation \eqref{eq:F1}), the precision-recall curve sums up the precision (Equation \eqref{eq:PRE}) and the recall (Equation \eqref{eq:TPR}) to form one evaluation metric.

\begin{minipage}{0.35\textwidth}
\includegraphics[width=\textwidth]{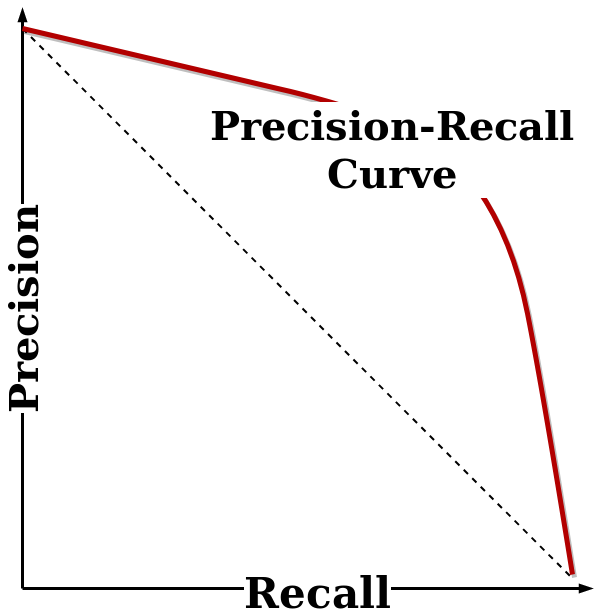}
\captionof{figure}{Schematic illustration of the precision-recall curve.}
\label{img:ROC}
\end{minipage}
\hfill
\begin{minipage}{0.55\textwidth}
Figure \ref{img:ROC} shows an example of a precision-recall curve. To get the precision-recall curve, several probability thresholds are plotted on the precision against the recall. The \ac{AUC} indicates the capability of distinguishing between two classes.  The best reachable \ac{AUC} is 1.0, since the maximum value of the precision and the recall is 1.0. The dotted line marks the precision-recall curve of random classification with an \ac{AUC} of $0.5$. 
\end{minipage}
\FloatBarrier

\chapter{Related Work}
\label{RelatedWork}

Foreground-background segmentation and background subtraction, which can be equated in many cases, attempt to classify pixels belonging either to the background or foreground of an image. The principle aim of these approaches is the creation of a stable background image in order to subtract the current frame from the modeled background. Performing a background subtraction, in which moving objects are considered as foreground pixels and static objects are considered as background pixels, leads to a segmentation of a scene into static and moving pixels. Therefore background subtraction can be an effective method of choice when it comes to segmentation of motion in video sequences.\\

Whilst early approaches (see Section \ref{earlyapproaches}) tackled the task of motion detection using classical algorithms like frame differencing or statistical methods, recent methods using neural networks (see Section \ref{DeepLearningbasedApp}) often outperform classic algorithmic approaches.\\
In this chapter, typical challenges of motion segmentation tasks, and state-of-the-art approaches to tackle those, are introduced. By pointing out their drawbacks, design goals for this work get derived.

\section{Challenges of Motion Segmentation Tasks}
\label{ChallengesofComputerVision}

There are several challenges in indoor and outdoor video sequences, which make motion segmentation error-prone. Some of the most defiant challenges are listed below:

\paragraph{Illumination Changes}
Sudden and gradual illumination changes due to changing environmental influences such as clouds passing by the sun bring a strong change in pixel values, which can generate an erroneous classification of a pixels.

\paragraph{Irrelevant Motion}
Motion segmentation tasks often struggle with irrelevant motion, which should be ignored or more specifically be not segmented as motion. In the case of the CDNet2014 dataset (see Section \ref{CDNet2014}), moving people, moving animals and moving man-made objects (e.g. cars, boats, trains) are considered as being relevant motion, while waves, waving trees, snowfall, fire or shadows and light reflections of moving objects should not be mistakenly perceived as relevant motion\footnote{The LASIESTA dataset (see Section \ref{LASIESTA}) uses the same definition of relevant and irrelevant motion.}.   

\paragraph{Moving Camera}
Motion due to camera jitter, panning, tilting, zooming or translational moving cameras is hard to separate from motion that originates from moving objects.


\section{State-of-the-Art}

\begin{figure}[h!]
\centering
\includegraphics[width=\textwidth]{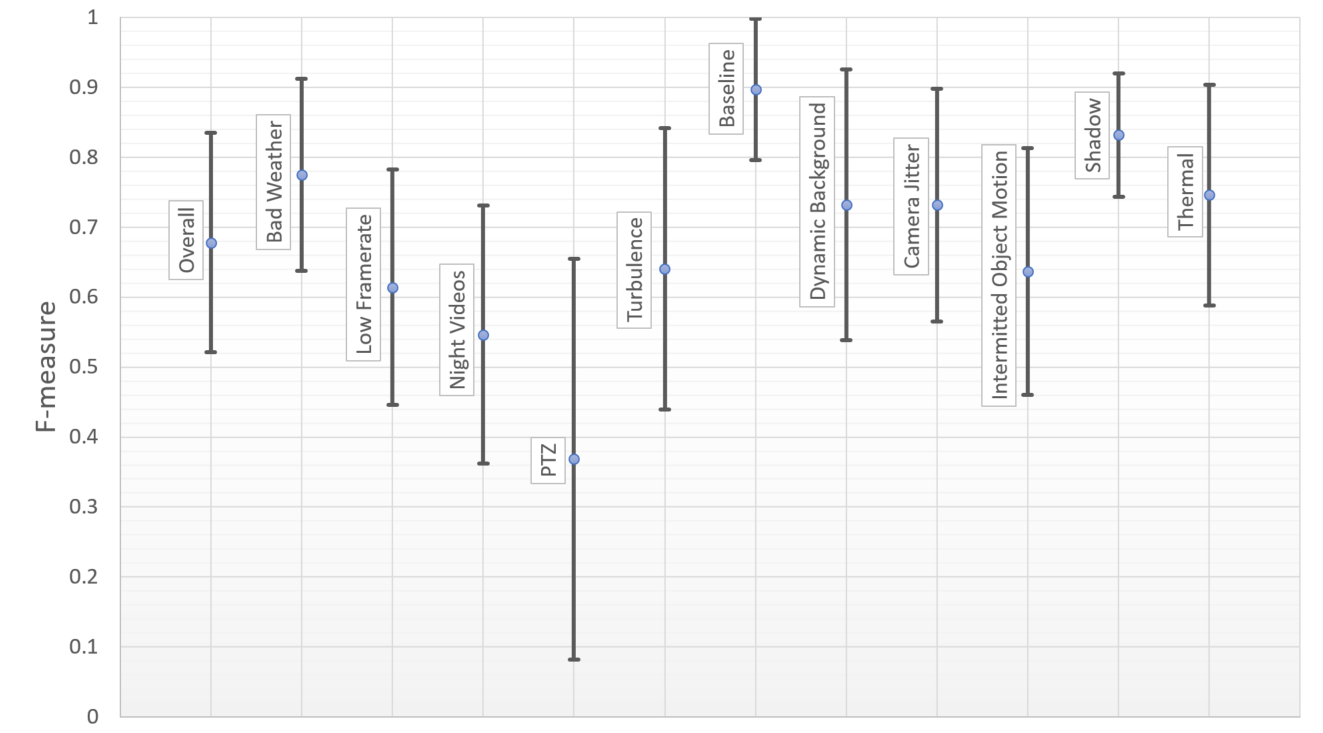}
\caption {Mean and standard deviation of F-measures of the 40 best ranked algorithms across all categories of the CDNet2014-dataset.}
\label{fig:FmeasuresOverview}
\end{figure}

Even tough not all methods mentioned in Section \ref{earlyapproaches} and Section \ref{DeepLearningbasedApp} are listed in the \mbox{CDNet2014} challenge\footnote{Methods \cite{Farneback, DISopticalflow, BSUV, 3DFR}, which are introduced in the following sections, are not listed in the CDNet2014 ranking} the published ranking on the CDNet2014 dataset \cite{CDNet2014} is the method of choice when stating the state-of-the-art. 
Figure \ref{fig:FmeasuresOverview} illustrates the the mean F-measure and its standard deviation over all categories of the dataset CDNet2014 (see Section \ref{CDNet2014}), where best 40 submitted methods are taken into consideration\footnote{The figure is based on the F-measure results, published on the CDNet2014 website \cite{CDNet2014} (Accessed: \today).}.\\

It should be noted that no uniform evaluation protocol is defined in the ranking of the CDNet2014 dataset, which makes comparison of methods within the ranking more difficult. Since there is no pre-defined evaluation strategy for the CDNet2014 ranking, there is a high variance of the resulting F-measures obtained by each method, what can be observed in the high standard deviations up to $0.305$ in Figure \ref{fig:FmeasuresOverview}\footnote{$0.305$ is the standard deviation in the \textit{\ac{PTZ}} category}. Nevertheless, the figure shows, which categories are more demanding. Whilst most methods show good performance in the standard \textit{Baseline} category, there is still weak performance when it comes to challenging categories like \textit{\ac{PTZ}}.

\FloatBarrier

Table \ref{tab:OverallF1_Realted} shows the ten best performing methods of the CDNet2014 ranking, regarding their overall F-measure, which indicates the state-of-the-art. A closer look on the methods \cite{Lim_v2, LimS, Lim}, of Lim et al. are given in Section \ref{recentmethods}, while in Chapter \ref{Eval} the MOSNET gets brought in line with the state-of-the-art.

{\rowcolors{2}{gray!3!}{gray!30!}
\begin{table}[h!]
\begin{center}
\caption[Ten methods with the highest overall F-measure on the CDNet2014-dataset]{Ten methods with the highest overall F-measure on the CDNet2014-dataset \cite{CDNet2014} (Accessed: \today)}
\begin{tabular}{cc}
\toprule 
\rule[-1ex]{0pt}{4ex}\textbf{Method} & \textbf{F-measure}\\ 
\midrule
\rule[-1ex]{0pt}{4ex}FgSegNet\_v2 \cite{Lim_v2} & 0.985 \\
\rule[-1ex]{0pt}{4ex}FgSegNet\_S \cite{LimS} & 0.980\\ 
\rule[-1ex]{0pt}{4ex}FgSegNet\_M \cite{Lim}& 0.977\\ 
\rule[-1ex]{0pt}{4ex}BSPVGAN \cite{BSPVGAN} & 0.950\\ 
\rule[-1ex]{0pt}{4ex}BSGAN \cite{BSGAN}& 0.934\\ 
\rule[-1ex]{0pt}{4ex}Cascade CNN \cite{CascadeCNN} & 0.921\\ 
\rule[-1ex]{0pt}{4ex}SemanticBGS \cite{Braham2} & 0.789\\ 
\rule[-1ex]{0pt}{4ex}IUTIS-5 \cite{Bianco} & 0.772\\ 
\rule[-1ex]{0pt}{4ex}SWCD \cite{SWCD} & 0.758\\ 
\rule[-1ex]{0pt}{4ex}IUTIS-3 \cite{Bianco} & 0.755\\ 
\bottomrule
\end{tabular}
\label{tab:OverallF1_Realted}
\end{center}
\end{table}
}
\FloatBarrier

\section{Early Approaches}
\label{earlyapproaches}

\emph{\say{Optic flow describes the displacement field in an image sequence.}}\cite{opticalflow} This can be used to separate moving objects from a stationary background in a video sequence. 
The open source computer vision library OpenCV \cite{opencv} provides algorithms \cite{Farneback, DISopticalflow}, which can be used for \ac{MOD}, based on optic flow. Their ease of use due to their public availability is contrasted by poor performance, which is presented in Section \ref{Comparisson}.\\ 

In 1999, Stauffer and Grimson \cite{StaufferGMM} proposed a method called Gaussian Mixture Model, which models the background using mixture of Gaussians as probability distribution function for each pixel belonging either to the background or foreground.\\

Another approach to tackle background subtraction is the use of frame differencing. Here, the differential image of two or more consecutive frames gets calculated and filtered in order to detect moving objects \cite{Singla}. This method struggles with alignment errors, especially in sequences captured with a moving camera.\\

Barnich et al. \cite{Barnich} proposed a method called Visual Background Extractor, that builds a background model by aggregating previously observed values for each pixels location \cite{Bouwmans2}.

\section{Deep Learning Based Approaches}
\label{DeepLearningbasedApp}

Since the utilization of machine learning in the field of computer vision, traditional image processing approaches have been overhauled in many computer vision tasks. The top ranked algorithms on the large-scale public dataset CDNet2014, which is widely used for background subtraction and change detection, are seven deep learning methods (\cite{CDNet2014}, Accessed: \today), namely FgSegNet\_v2 \cite{Lim_v2}, FgSefNet\_S \cite{LimS}, FgSegNet\_M \cite{Lim}, BSPVGAN \cite{BSPVGAN}, BSGAN \cite{BSGAN} and Cascaded \ac{CNN} \cite{CascadeCNN}, to name just a few of them. The ranking according their F-measure is shown in detail in Section \ref{Comparisson}.\\

The network architecture itself, the training strategy, as well as the split of the dataset used during training and evaluation have a great impact on the performance  of the deep learning based approach and its capability to generalize. Therefore, the deep learning based approaches are parted in scene-specific and scene-independent methods. In scene-specific approaches, a neural network is optimized for a specific scene, and therefore the data used to train and evaluate the network may originate from the same scene. In contrast to this, scene-independent methods aim to train and evaluate on data originating of different scenes, called \ac{SIE}.   

\subsection{Scene-Specific Deep Learning Methods}
\label{recentmethods}

In 2016, Braham and Van Droogenbroek \cite{Braham} introduced the first convolutional neural network to tackle the problem of background subtraction. Even though they have exceeded most of the conventional methods, there are still weaknesses in their approach. The network is not generazing well, since it has to be trained on the scene its evaluated on in order to generate a model of the background. This may be suitable for scenes captured with a static camera, but leads to a weak performance if the camera is moving, panning, tilting or zooming. Aside from that the network contains fully-connected layers, which limits the network in handling images of various size.\\


Wang et al. \cite{CascadeCNN} introduced a method, called Cascaded CNN, which is based on a multi-scale convolution network. One input frame runs through three networks with a different scale, in order to combine the results and perform a prediction. The approach of Wang et al. is scene-specific, since it generates a background model during training, what entails the data used during training and evaluation must originate from the same scene.

Lim et al. published three approaches, namely FgSegNet\_S \cite{LimS}, FgSegNet\_M \cite{Lim} and FgSegNet\_v2 \cite{Lim_v2}, which are currently leading the CDNet2014 challenge. The FgSegNet\_S and FgSegNet\_M are based on the approach of Wang et al. \cite{CascadeCNN}, which entails the drawback of being scene-specific.\\ 

The FgSegNet\_v2, which is currently ranked as number one on the CDNet2014 dataset with an overall F-measure of $0.9847$ \cite{Lim_v2}, is trained using a similar training strategy but a new network architecture.  
Lim et al. are using an encoder-decoder structure for learning multi-scale features for foreground segmentation. A pre-trained VGG-16 network (see Section \ref{VGG-16}), based on the work from Simonyan et al. \cite{Simonyan}, is used as encoder to generate feature maps. The decoder network extracts multi-scale features, fuses them and generates a binary foreground mask. Lim et al. trained their models on a set of randomly chosen images of each video sequence of the CDNet2014 dataset separately, in order to get a separate neural network for each video sequence. Evaluating these models with unseen frames of the same video sequence leads to the outstanding F-measure obtained on each scene. Due to the random selection of training images in one scene, it can be assumed that the neural network learns the whole background occurring in the scene during training. With the learned background model, the network can perform a background subtraction and generate foreground masks, respectively motion masks.\\

\begin{figure}[h!]%
  \centering
  \subfloat[][Input]{\includegraphics[width=0.3\textwidth]{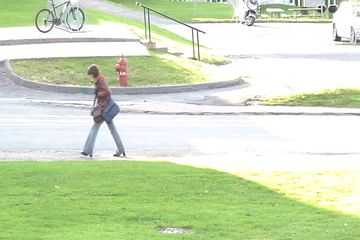}}%
  \qquad
  \subfloat[][\ac{GT}]{\includegraphics[width=0.3\textwidth]{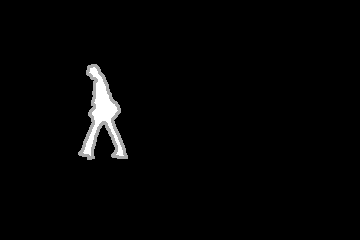}}%
  \qquad
  \subfloat[][Prediction]{\includegraphics[width=0.3\textwidth]{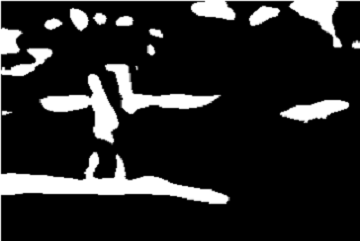}}%
  \caption[A motion mask of the scene \textit{pedestrians} predicted by the FgSegNet\_v2]{Input image (a) and \ac{GT} (b) of the scene \textit{pedestrians} and the motion mask (c), predicted by the FgSegNet\_v2 trained on the CDNet2014 dataset except the scene \textit{pedestrians}.}%
  \label{FgSegNet_Model}
\end{figure}

In order to evaluate the capability to generalize and point out the drawbacks of the FgSegNet\_v2, the training and evaluation strategy is changed to perform an \ac{SIE}. To do so, the split of the CDNet2014 dataset is changed, such that in each case one scene is used for evaluation and the rest for training. All further training settings and hyperparameters are retained from Lim et al. The published code and trained models of the FgSegNet\_v2 can be accessed at \cite{FuckNetv2}.
Unlike the training strategy of Lim et al., a single network is trained on all scenes of the CDNet2014 dataset, except the scene \textit{pedestrians} of the \textit{Baseline} category in order to evaluate the performance of the network on unseen scenes.  
Figure \ref{FgSegNet_Model} (c) shows the predicted motion mask for the corresponding input image shown in Figure \ref{FgSegNet_Model} (a). A comparison with the \ac{GT} in Figure \ref{FgSegNet_Model} (b) shows the weak performance of the network on unseen scenes. This is because the network could not create a background model of the \textit{pedestrians} scene during training.       

\FloatBarrier

\begin{figure}[h!]%
  \centering
  \subfloat[][Input]{\includegraphics[width=0.3\textwidth]{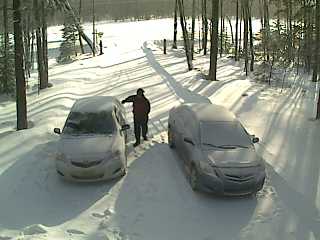}}%
  \qquad
  \subfloat[][\ac{GT}]{\includegraphics[width=0.3\textwidth]{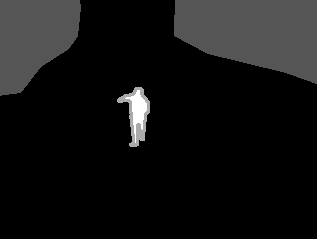}}%
  \qquad
  \subfloat[][Prediction]{\includegraphics[width=0.3\textwidth]{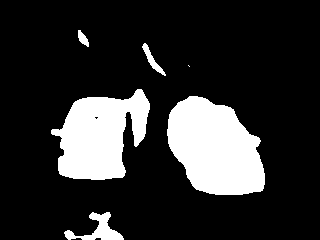}}%
  \caption[A motion mask of the scene \textit{WinterDriveway} predicted by the FgSegNet\_v2]{Input image (a) and \ac{GT} (b) of the scene \textit{WinterDriveway} and the motion mask (c), predicted by the FgSegNet\_v2 trained on the CDNet2014 dataset except the scene \textit{WinterDriveway}.}%
  \label{FgSegNet_Semantic}
\end{figure}

Another weak point of the approach of Lim et al. can be shown when training a single network on all scenes of the CDNet2014 dataset, except the scene \textit{WinterDriveway} of the \textit{IntermittentObjectMotion} category. The \ac{GT} in Figure \ref{FgSegNet_Semantic} (b) shows the pixels belonging to the person in the input image Figure \ref{FgSegNet_Semantic} (a), are labeled as motion. The predicted output of the FgSegNet\_v2 in Figure \ref{FgSegNet_Semantic} (c) shows the two parked cars are wrongly predicted as motion. It can be deduced that the network mainly learns semantic context instead of motion itself.    

\FloatBarrier

\subsection{Scene-Independent Deep Learning Methods}
\label{recentmethods_2}

While scene-specific approaches lead the CDNet2014 challenge, their practical use is strongly limited due to their poor capability to generalize, i.e. the area of application of the approach must already be known during training.
Two supervised scene-independent methods try to tackle this weakness and outperform most of the scene-specific methods in a \ac{SIE}. The approach of Mandal et al. \cite{3DFR}, namely 3DFR, uses \ac{SIE} to point out the performance of their network on unseen scenes of each category of the CDNet2014 dataset, with an overall F-measure of 0.86 \cite{3DFR}. Mandel et al. feed their network with $50$ frames and perform a set of 3D convolutions, which lead to the great amount of 126.45k trainable parameters. A drawback of the method of Mandal et al. is the exclusion of the \textit{PTZ} category for developing and evaluating their approach.\\

In 2020, Ozan et al. \cite{BSUV} introduced a fully-convolutional neural network for background subtraction of unseen videos, reaching an overall F-measure of 0.787 on the CDNet2014 dataset. They performed a \ac{SIE} with a set of 18 different dataset split combinations, to demonstrate their networks capability to generalize. Besides the current frame, and a previous frame, the network must be fed with a reference frame, which does not contain any motion. The method of Ozan et al. needs 100 motion-free frames to calculate the reference frame, which limits the method for an application on scenes captured with a moving camera.

\section{Summary and Findings}
\label{sec:Motivation}

The disadvantage of methods, which are based on the generation of a background model is the poor ability to abstract the learned on differing scenes, which make them scene-specific. Furthermore, it is essential for such approaches that they can obtain a model of the background occurring in the scene during the training. Therefore, these approaches perform well when the data used for training and evaluation originates from the same scene but show poor performance when the training data and evaluation data originate of different scenes. This restricts these approaches to videos captured by static cameras, which exact application purpose is known during development.\\

The few scene-independent approaches published lack in their ability to handle scenes captured with non-static cameras, which restricts their use  applications with \ac{PTZ}-cameras.\\

The general idea of motion segmentation in video sequences is to represent a moving object as a set of pixels in an image of a video sequence having a coherent motion over time and a semantic similarity over image space \cite{Yazdi}.
Therefore, the neural network shall be able to recognize movement by means of a temporal and spatial component in cohesive image frames.
The aim is to develop an approach that is not based on a background model of the scenery, but rather is able to pixel-precise detect relevant motion by exploiting a temporal and spatial component in cohesive image frames. This leads to a scene-independent deep learning based approach, which is able to handle video sequences captured with a non-static camera.

\chapter{Datasets}
\label{Datasets}

Since the aimed approach is based on deep learning, the dataset used to train, validate and evaluate the network has a great impact on the performance of the approach. As mentioned in Section \ref{Datasetsplit}, especially the split of the dataset is one of the main influencing factors when inferring the performance of the approach. The following chapter introduces two datasets used in this work.

\section{Requirements on the Dataset and its Split}
\label{requirementsDataset}

To fit the task addressed in this work, the data used for evaluation $D_E$ must fulfil the following preconditions:

\begin{itemize}
    \item pixel-wise annotations
    \item all positive annotated pixels belong to moving objects
    \item all pixels belonging to relevant moving objects are annotated positive
    \item contains sequences captured with a moving camera
\end{itemize}{}

In addition to the necessary requirements for the data $D_E$, there are soft requirements, which strongly affect the generalization ability of the deep learning based approach. The data should be as diverse as possible, in order to be representative for a wide range of practical applications. Furthermore, it should have diversity in size and semantic of the moving object and diversity in the semantic of the whole scene.\\

The data $D_T$ used to train the neural network should largely meet the same requirements in order to get good performance on the task the network is addressed for represented by the evaluation data $D_E$.\\

The training progress can be validated at defined intervals during the training using the validation data $D_V$. Validation reveals how well the network is performing on unseen data using predefined metrics like the validation loss or the validation F-measure. Since the network does not adjust its weights and biases during validation, the same validation data $D_V$ can be used multiple times.\\

In order to perform a \ac{SIE}, the data used to train the network $D_T$ and the data used to evaluate the approach $D_E$ do not originate of the same scene.

\section{CDNet2014}
\label{CDNet2014}

In 2019, Kalsotra et al. \cite{Kalsotra} published a comprehensive survey of video datasets for background subtraction. In this paper, comparing 41 datasets for background subtraction, Kalsotra et al. mentioned the CDNet2014 dataset as \emph{\say{[..] the most prolific dataset in the list of background subtraction datasets [...]}} (\cite{Kalsotra}, p.17). Based on a citation frequency of 959 citation between 2011 and 2018 Kalostra et al. mentioned the CDNet2014 dataset as the most common accepted dataset to evaluate background subtraction methods. Since there is only a small domain gap between background subtraction and motion segmentation\footnote{Performing a background subtraction, where moving objects are considered as foreground and static objects are considered as background, leads to a segmentation into pixels being in motion and pixels being static.}, the CDNet2014 dataset is used to train, validate and evaluate the MOSNET.\\

The CDNet2014 dataset consists of a wide range of camera-captured videos packed in the categories shown in Table \ref{tab:CDNet2014Categories}. Most categories of the CDNet2014 dataset aim at a specific challenge in computer vision. For example the categories \textit{Dynamic Background}, \textit{Shadows}, \textit{Bad Weather}, \textit{\ac{IOM}} and \textit{Turbulence} focus the challenge of distinguishing relevant and irrelevant motion while the categories \textit{Camera Jitter} and \textit{\ac{PTZ}} focus the challenge of camera motion.  

{\rowcolors{3}{gray!3!}{gray!30!}
\begin{table}[h!]
\begin{center}
\caption[CDNet2014 categories]{CDNet2014 categories \cite{Wang}.}
\begin{tabular}{ccccl}
\toprule
\multicolumn{5}{c}{\textbf{CDNet2014}}              \\ \midrule
\textbf{Category}          & \textbf{Scenes} & \textbf{Indoor} & \textbf{Outdoor} & \textbf{Description} \\ \midrule
Baseline                  & 4      & 2      & 2  &  simple scenes   \\ 
\begin{tabular}[c]{@{}l@{}}Dynamic\\ Background\end{tabular}       & 6      & -      & 6  & irrelevant object motion    \\
Camera Jitter             & 4      & 1      & 3  &  heavy camera jitter  \\ 
Shadows                   & 6      & 2      & 4  &   \begin{tabular}[c]{@{}l@{}}prevalent hard \\ and soft shadow \\ and intermittent shades\end{tabular}  \\
\begin{tabular}[c]{@{}l@{}}Intermittent\\ Object Motion\end{tabular} & 6      & 1      & 5  &  \begin{tabular}[c]{@{}l@{}}background objects \\ moving away\end{tabular}    \\
Thermal                   & 5      & 2      & 3   &  thermal camera videos  \\
Bad Weather               & 4      & -      & 4   & \begin{tabular}[c]{@{}l@{}}strong rain,\\ snowfall and wind\end{tabular}  \\
Low Framerate             & 4      & -      & 4   &  \begin{tabular}[c]{@{}l@{}}videos with\\ low framerate\end{tabular}  \\ 
Night Videos                     & 6      & -      & 6   & \begin{tabular}[c]{@{}l@{}}videos shot\\ at night\end{tabular}      \\ 
PTZ                       & 4      & -      & 4   &   \begin{tabular}[c]{@{}l@{}}panning, tilting\\ and zooming cameras\end{tabular} \\ 
Turbulence                & 4      & -      & 4   &   containing turbulence \\ \bottomrule
\end{tabular}
\label{tab:CDNet2014Categories}
\end{center}
\end{table}
}
\FloatBarrier

\begin{figure}[h!]
\centering
\includegraphics[scale=0.5]{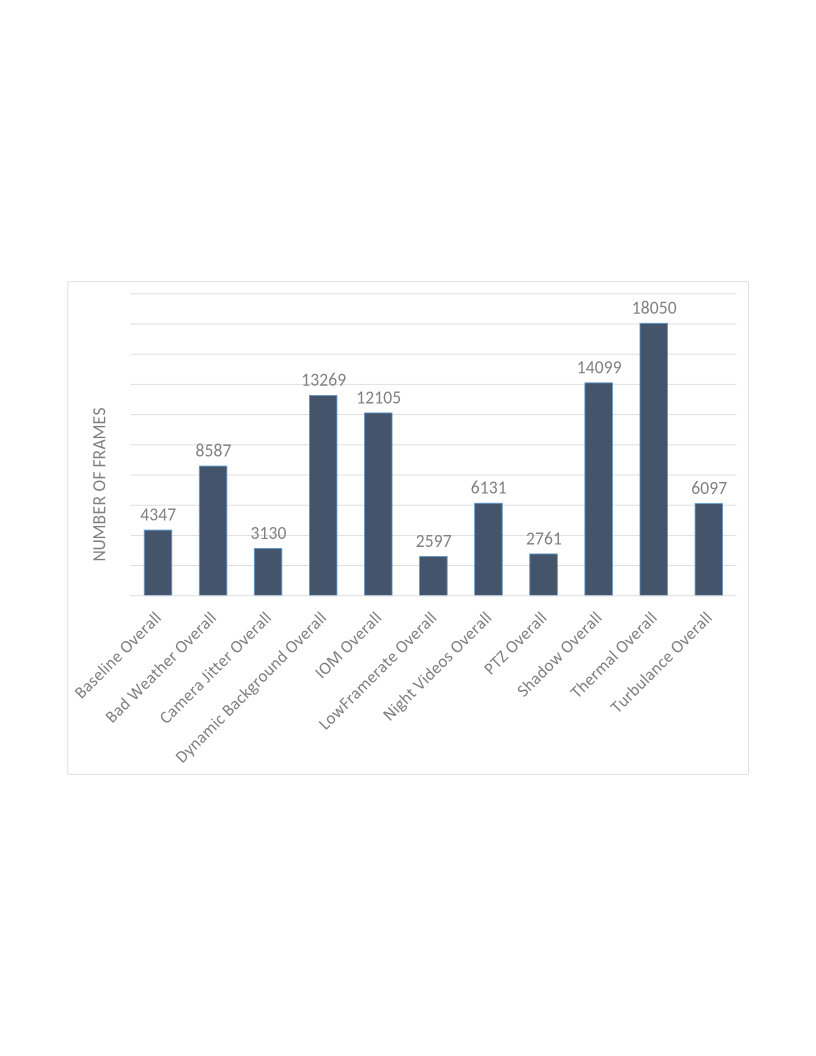}
\caption {Number of not fully non-\ac{ROI} annotated frames of the CDNet2014-dataset listed per category.}
\label{fig:CDNet2014_Anal_chart}
\end{figure}

The $53$ scenes, consists of $\approx\!160,000$ annotated frames \cite{Wang}, while the first few hundred frames of each scene are annotated as completely being non-\ac{ROI}. As mentioned in Section \ref{datahandling}, pixels labeled as non-\ac{ROI} are not considered when training, validating and evaluating the MOSNET, only the 88,412 frames which are not fully annotated as being non-\ac{ROI} get considered in Figure \ref{fig:CDNet2014_Anal_chart} and Table \ref{tab:Splitshit}.\\ 
Figure \ref{fig:CDNet2014_Anal_chart} shows the number of not fully non-\ac{ROI} annotated frames listed per category. It can be seen that there is a strong imbalance in the number of annotated images per category. Especially the categories \textit{Camera Jitter} and \textit{\ac{PTZ}} make up only a small part of the dataset. Therefore, a additional dataset, consisting of further scenes captured with a moving camera, namely LASIESTA (see Section \ref{LASIESTA}) gets used for cross-evaluation in Section \ref{ExpCrossEval}. A cross-evaluation, in which the MOSNET gets trained on a split of the CDNet2014 dataset (i.e. the data $D_T$) and gets evaluated on a different dataset (i.e. the LASIESTA dataset), demonstrates also the MOSNETs capability to generalize.

\FloatBarrier

\paragraph{Split into Training-, Validation- and Evaluation-Data}
\label{CDNet2014split}
While the training data and validation data are used during the development process, the evaluation data is used to analyze the deep learning approach. A common practice of splitting the data into data used during development and used during evaluation is a ratio of about $(80:20)$.  
Furthermore, Guyon recommends using about $10\% - 25\%$ of the developing data for validation \cite{Guyon}. To split the dataset into data used for training, validation, and evaluation the following split criteria are defined:  

\begin{itemize}
    \item each category is present during the development
    \item each category is present during evaluation
    \item data used during development and evaluation do not originate from the same scene\footnote{This aims at performing a \ac{SIE} in Chapter \ref{Eval}.}
    \item $\approx 80\%$ of the dataset is used during the development process
    \item $\approx 20\%$ of the dataset is used during evaluation
    \item $10\% - 25\%$ of the development data is used for validation
\end{itemize}

The CDNet2014 dataset is split into three parts: while two of them ($D_T$ and $D_V$) are used during the developing process, the third is used for evaluating the MOSNET ($D_E$). The Table \ref{tab:Splitshit} names the split of the dataset into the three parts. The data used during the development process is split into training data $D_T$, which make up $80\%$ and validation data $D_V$, which make up $20\%$. The split is implemented after shuffling the data used during development ($D_T$ and $D_V$), further explained in Section \ref{datahandling}.\\

\begin{longtable}{llcc}
\caption[Split of the CDNet2014 dataset into]{Split of the CDNet2014 dataset into data used for development,i.e. $D_T$ and $D_V$ and data used for evaluation $D_E$. The numbers indicate how many not fully non-\ac{ROI} annotated frames per scene are allocated for development and evaluation, respectively.} \label{tab:Splitshit} \\

\toprule \multicolumn{1}{l}{\textbf{Category}}& \multicolumn{1}{l}{\textbf{Scene}} & \multicolumn{1}{c}{\textbf{$D_T$/$D_V$}} & \multicolumn{1}{c}{\textbf{$D_E$}} \\ \hline 
\endfirsthead

\multicolumn{2}{c}%
{{\bfseries \tablename\ \thetable{} -- continued from previous page}} \\
\toprule \multicolumn{1}{l}{\textbf{Category}}& \multicolumn{1}{l}{\textbf{Scene}} & \multicolumn{1}{c}{\textbf{$D_T$ / $D_V$}} & \multicolumn{1}{c}{\textbf{$D_E$}} \\ \hline 
\endhead

\multicolumn{4}{r}{{\textbf{Continued on next page --}}} \\
\endfoot

\bottomrule
\endlastfoot

\multirow{4}{*}{Baseline}& \cellcolor{gray!3!}Office                & \cellcolor{gray!3!}1466                  &\cellcolor{gray!3!}            \\
&\cellcolor{gray!30!}Highway               &\cellcolor{gray!30!} 1230                       &\cellcolor{gray!30!}        \\
&\cellcolor{gray!3!}Pedestrians           &\cellcolor{gray!3!}               &\cellcolor{gray!3!}     751            \\
&\cellcolor{gray!30!}PETS2006              &\cellcolor{gray!30!} 900                   &\cellcolor{gray!30!}            \\ \midrule
\multirow{4}{*}{Bad Weather}&\cellcolor{gray!3!}Blizzard              &\cellcolor{gray!3!} 2689                  &\cellcolor{gray!3!}            \\
&\cellcolor{gray!30!}Skating               &\cellcolor{gray!30!}                       &\cellcolor{gray!30!} 1550       \\
&\cellcolor{gray!3!}Snowfall              &\cellcolor{gray!3!} 2849                  &\cellcolor{gray!3!}            \\
&\cellcolor{gray!30!}Wet Snow              &\cellcolor{gray!30!} 1499                  &\cellcolor{gray!30!}            \\ \midrule
\multirow{4}{*}{Camera Jitter}&\cellcolor{gray!3!}Badminton             &\cellcolor{gray!3!} 350                   & \cellcolor{gray!3!}           \\
&\cellcolor{gray!30!}Boulevard             &\cellcolor{gray!30!}                   &\cellcolor{gray!30!}     1710       \\
&\cellcolor{gray!3!}Sidewalk              &\cellcolor{gray!3!} 400                   &\cellcolor{gray!3!}            \\
&\cellcolor{gray!30!}Traffic               &\cellcolor{gray!30!}    670                   &\cellcolor{gray!30!}         \\ \midrule
&\cellcolor{gray!3!}Boats                 &\cellcolor{gray!3!} 6099                  &\cellcolor{gray!3!}            \\ 
\multirow{6}{*}{Dynamic Background}&\cellcolor{gray!30!}Canoe                 &\cellcolor{gray!30!}                       &\cellcolor{gray!30!} 389        \\
&\cellcolor{gray!3!}Fall                  &\cellcolor{gray!3!} 299                   &\cellcolor{gray!3!}            \\
&\cellcolor{gray!30!}Fountain01            &\cellcolor{gray!30!} 784                   &\cellcolor{gray!30!}            \\
&\cellcolor{gray!3!}Fountain02            &\cellcolor{gray!3!} 999                   &\cellcolor{gray!3!}            \\
&\cellcolor{gray!30!}Overpass              &\cellcolor{gray!30!} 1999                  &\cellcolor{gray!30!}            \\ \midrule
\multirow{6}{*}{\begin{tabular}[c]{@{}l@{}}Intermittent\\ Object Motion\end{tabular}}&\cellcolor{gray!3!}AbandonedBox          &\cellcolor{gray!3!} 2050                  &\cellcolor{gray!3!}            \\
&\cellcolor{gray!30!}Parking               &\cellcolor{gray!30!} 1400                  &\cellcolor{gray!30!}            \\
&\cellcolor{gray!3!}Sofa                  & \cellcolor{gray!3!}2250                  & \cellcolor{gray!3!}           \\
&\cellcolor{gray!30!}StreetLight           &\cellcolor{gray!30!} 3025                  &\cellcolor{gray!30!}            \\
&\cellcolor{gray!3!}Tramstop              &\cellcolor{gray!3!} 1880                  & \cellcolor{gray!3!}           \\
&\cellcolor{gray!30!}WinterDriveway        &\cellcolor{gray!30!}                       &\cellcolor{gray!30!} 1500       \\ \midrule
\multirow{4}{*}{Low Framerate}&\cellcolor{gray!3!}Port\_0\_17fps        &\cellcolor{gray!3!} 999                   &\cellcolor{gray!3!}            \\
&\cellcolor{gray!30!}Tram\_Crossroad\_1fps &\cellcolor{gray!30!}                       &\cellcolor{gray!30!} 249        \\
&\cellcolor{gray!3!}TunnelExit\_0\_35fps  &\cellcolor{gray!3!} 1000                  &\cellcolor{gray!3!}            \\
&\cellcolor{gray!30!}Turnpike\_0\_5fps     &\cellcolor{gray!30!} 349                   &\cellcolor{gray!30!}            \\ \midrule
\multirow{6}{*}{Night Videos}&\cellcolor{gray!3!}BridgeEntry           &\cellcolor{gray!3!} 749                   &\cellcolor{gray!3!}            \\
&\cellcolor{gray!30!}BusyBoulevard         &\cellcolor{gray!30!} 1014                  &\cellcolor{gray!30!}            \\
&\cellcolor{gray!3!}FluidHighway          &\cellcolor{gray!3!} 481                   &\cellcolor{gray!3!}            \\
&\cellcolor{gray!30!}StreetCorneratNight   &\cellcolor{gray!30!} 2199                  &\cellcolor{gray!30!}            \\
&\cellcolor{gray!3!}TramStation           &\cellcolor{gray!3!} 1249                  &\cellcolor{gray!3!}            \\
&\cellcolor{gray!30!}WinterStreet          &\cellcolor{gray!30!}                       &\cellcolor{gray!30!} 439        \\ \midrule
\multirow{4}{*}{\ac{PTZ}} &\cellcolor{gray!3!}ContiuousPan          &\cellcolor{gray!3!} 549                   &\cellcolor{gray!3!}            \\
&\cellcolor{gray!30!}intermittentPan       &\cellcolor{gray!30!}1149                  &\cellcolor{gray!30!}            \\
&\cellcolor{gray!3!}twoPositionPTZCam     &\cellcolor{gray!3!}                       &\cellcolor{gray!3!} 749        \\
&\cellcolor{gray!30!}ZoomInZoomOut         &\cellcolor{gray!30!} 314                   &\cellcolor{gray!30!}            \\ \midrule
\multirow{6}{*}{Shadow}&\cellcolor{gray!3!}Backdoor              &\cellcolor{gray!3!} 1600                  &\cellcolor{gray!3!}            \\
&\cellcolor{gray!30!}Bungalows             &\cellcolor{gray!30!} 1400                  &\cellcolor{gray!30!}            \\
&\cellcolor{gray!3!}BusStation            &\cellcolor{gray!3!} 950                   &\cellcolor{gray!3!}            \\
&\cellcolor{gray!30!}CopyMachine           &\cellcolor{gray!30!} 2900                  &\cellcolor{gray!30!}            \\
&\cellcolor{gray!3!}Cubicle               &\cellcolor{gray!3!} 6300                  &\cellcolor{gray!3!}            \\
&\cellcolor{gray!30!}PeopleinShade         &\cellcolor{gray!30!}                       &\cellcolor{gray!30!} 949        \\ \midrule
\multirow{5}{*}{Thermal}&\cellcolor{gray!3!}Corridor              &\cellcolor{gray!3!} 4900                  &\cellcolor{gray!3!}            \\
&\cellcolor{gray!30!}DiningRoom            &\cellcolor{gray!30!} 3000                  &\cellcolor{gray!30!}            \\
&\cellcolor{gray!3!}Lakeside              &\cellcolor{gray!3!} 5500                  &\cellcolor{gray!3!}            \\
&\cellcolor{gray!30!}Library               &\cellcolor{gray!30!} 4300                  &\cellcolor{gray!30!}            \\
&\cellcolor{gray!3!}Park                  & \cellcolor{gray!3!}                      & \cellcolor{gray!3!}350        \\ \midrule
\multirow{4}{*}{Turbulence }&\cellcolor{gray!30!}Turbulance0           &\cellcolor{gray!30!} 1999                  &\cellcolor{gray!30!}            \\
&\cellcolor{gray!3!}Turbulance1           &\cellcolor{gray!3!} 1399                  &\cellcolor{gray!3!}            \\
&\cellcolor{gray!30!}Turbulance2           &\cellcolor{gray!30!} 2000                  &\cellcolor{gray!30!}            \\
&\cellcolor{gray!3!}Turbulance3           &\cellcolor{gray!3!}                       &\cellcolor{gray!3!} 699        \\ \midrule
\multirow{1}{*}{\textbf{Sum}} &                      & \textbf{79818}                 &\textbf{8594}      
\end{longtable}

In order to evaluate the networks ability to generalize, the data used for training $D_T$ and the data used for evaluation $D_E$ are not from the same scene. One scene of each category is used for evaluating the network. The data used for evaluating makes up $11.2\%$ of the whole dataset. Since an additional dataset (see Section \ref{LASIESTA}) is used for cross-evaluating, the evaluation split of the CDNet2014 is smaller than $20\%$.

\section{LASIESTA}
\label{LASIESTA}

The LASIESTA dataset \cite{LASIESTA} is a pixel-wise annotated dataset for multi object segmentation in video sequences. The dataset consists of 13 categories listed in Table \ref{tab:LASIESTA} with 18,425 frames in total. Since 9,850 frames of the dataset belong to scenes captured with a moving camera, the dataset is suitable for evaluate the networks capability to handle scenes captured with a moving camera. Using the LASIESTA dataset for cross-evaluation is further discussed in Section \ref{ExpCrossEval}.\\
 
\begin{figure}[h!]%
  \centering
  \subfloat[][Input]{\includegraphics[width=0.3\textwidth]{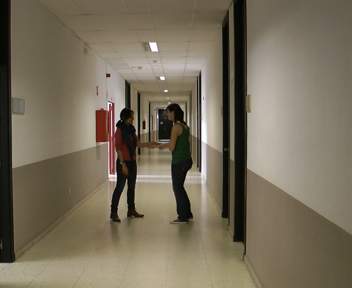}}%
  \qquad
  \subfloat[][\ac{GT}]{\includegraphics[width=0.3\textwidth]{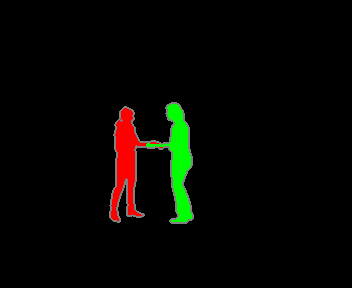}}%
   \qquad
  \subfloat[][\ac{GT} after relabeling]{\includegraphics[width=0.3\textwidth]{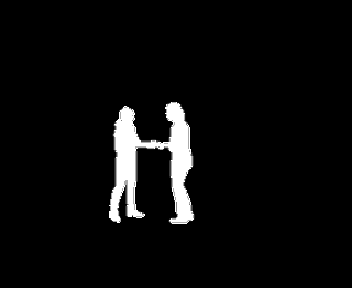}}%
  \caption[Example input frame and corresponding \ac{GT} of the LASIESTA dataset]{Example input frame (a) and corresponding \ac{GT} (b) of the LASIESTA dataset, respectively the resulting \ac{GT} after relabeling (c).}%
  \label{LASIESTA_example}
\end{figure}

Figure \ref{LASIESTA_example} shows an example frame (a) of the dataset with its corresponding \ac{GT} (b). Since this dataset is used to tackle multi object video segmentation, the annotation distinguishes between a maximum of three object instances, occurring in the scene by different labels. In order to use the dataset for the task of motion segmentation, the \ac{GT} has to be relabeled into moving objects and background. To do so, the different labels for different object instances get relabeled to a single label, representing a moving object. An example of the relabeled \ac{GT} can be seen in Figure \ref{LASIESTA_example} (c.).

{\rowcolors{3}{gray!3!}{gray!30!} 
\begin{table}[h!]
\caption[LASIESTA dataset overview]{LASIESTA dataset overview \cite{LASIESTA}}
\label{tab:LASIESTA}
\begin{tabular}{ccccc}
\toprule
\multicolumn{5}{c}{\textbf{LASIESTA}}                                                                                                                                                                                                                                                                                                         \\ \midrule
\textbf{Category}         & \textbf{\begin{tabular}[c]{@{}c@{}}Indoor/\\  Outdoor\end{tabular}} & \textbf{\begin{tabular}[c]{@{}c@{}}Scenes\end{tabular}} & \textbf{\begin{tabular}[c]{@{}c@{}}Frames\end{tabular}} & \textbf{Description}                                                                        \\ \midrule
Simple Sequences (SI)     & Indoor                                                              & 2                                                                       & 600                                                                   & simple scenes                                                                               \\ 
Camouflage (CA)           & Indoor                                                              & 2                                                                       & 875                                                                   & \begin{tabular}[c]{@{}c@{}}moving object with similar\\  color like the background\end{tabular} \\
Occlusion (OC)            & Indoor                                                              & 2                                                                       & 500                                                                   & \begin{tabular}[c]{@{}c@{}}totally or partly occluded\\  objects\end{tabular}               \\ 
Illumination Changes (IL) & Indoor                                                              & 2                                                                       & 825                                                                   & global illumination changes                                                                 \\ 
Modified Background (MB)  & Indoor                                                              & 2                                                                       & 800                                                                   & suddenly changing background                                                                \\ 
Bootstrap (BS)            & Indoor                                                              & 2                                                                       & 550                                                                   & \begin{tabular}[c]{@{}c@{}}moving objects from\\  the first frame\end{tabular}              \\ 
Moving Camera (MC)        & Indoor                                                              & 2                                                                       & 550                                                                   & \begin{tabular}[c]{@{}c@{}}pan and tilt motion or\\  camera jitter\end{tabular}             \\ 
Simulated Motion (SM)     & Indoor                                                              & 12                                                                      & 3,600                                                                  & \begin{tabular}[c]{@{}c@{}}different types and intensities\\  of camera motion\end{tabular} \\ 
Cloudy Conditions (CL)    & Outdoor                                                             & 2                                                                       & 650                                                                   & scene have cloudy conditions                                                                \\ 
Rainy Conditions (RA)     & Outdoor                                                             & 2                                                                       & 1,775                                                                  & \begin{tabular}[c]{@{}c@{}}scenes with strong irrelevant\\  motion (rain)\end{tabular}      \\ 
Snowy Conditions (SN)     & Outdoor                                                             & 2                                                                       & 1,350                                                                  & \begin{tabular}[c]{@{}c@{}}scenes with strong irrelevant\\  motion (snow)\end{tabular}      \\ 
Sunny Conditions (SU)     & Outdoor                                                             & 2                                                                       & 650                                                                   & \begin{tabular}[c]{@{}c@{}}sunny scenes that show\\  hard shadows\end{tabular}              \\ 
Moving Camera (MC)        & Outdoor                                                             & 2                                                                       & 600                                                                   & \begin{tabular}[c]{@{}c@{}}pan and tilt motion or\\  camera jitter\end{tabular}             \\ 
Simulated Motion (SM)     & Outdoor                                                             & 12                                                                      & 5,100                                                                  & \begin{tabular}[c]{@{}c@{}}different types and intensities\\  of camera motion\end{tabular} \\ \bottomrule
\end{tabular}
\end{table}
}

\FloatBarrier

\chapter{Network Architecture}
\label{Architecture}

This chapter introduces the network architecture developed in this work.
Since a moving object can be represented as a set of connected pixels in an image of a video sequence having a coherent motion over time and a semantic similarity over image space \cite{Yazdi}, the network architecture aims at capturing spatial and temporal information in subsequent frames.

\paragraph{Problem Formulation}

A video sequence of a dataset used for supervised learning with $N$ frames consists of $N$ input images $x_t$ with $x_t \in \mathbb{R}^{h \times w \times 3}$ and $t \in {0..N-1}$ and $N$ groundtruth images $y_t$ with $y_t \in \mathbb{R}^{h \times w \times 1}$ and $t \in {0..N-1}$, while $t$ denotes the time step and $h$ and $w$ the height and width of the image. The MOSNET aims at finding a function $f$ wich leads to a minimum inaccuracy between $y_t$ and $\hat{y}_t$, calculated by the objective function. The MOSNET targets to output a predicted motion mask $\hat{y}_t$ for $x_t$ by leveraging not only $x_t$ but also frames at previous time steps and following time steps.

Therefore, the network gets fed with an odd number $n \ll N$ of input frames, in order to predict a motion mask $\hat{y}_t$, noted in Equation \eqref{eq:MOSNET}:    .

\begin{equation}
\hat{y}_t  = f(x_{t-\frac{n-1}{2}}, ... ,x_{t-1},x_t, x_{t+1}, ..., x_{t+\frac{n-1}{2}})
\label{eq:MOSNET}
\end{equation}
\FloatBarrier

\section{MOSNET -- A Brief Overview}
\label{MOSNETOverview}

The basic network architecture of the MOSNET is an encoder-decoder structure, where the networks \textit{EC1}\footnotemark and \textit{EC2}\footnotemark[\value{footnote}], introduced in Section \ref{EC1} and Section \ref{EC2} form the encoder and the network \textit{DC}\footnotemark[\value{footnote}] represents the decoder. The encoder-decoder structure is a common architecture in the field of computer vision e.g., \cite{Lim, Lim_v2, NingXu, Akilan}, since its input and output have the same dimension, which enables a pixel wise calculation of the objective function.
\footnotetext{These are designations, which are introduced in the context of this work in order to describe certain parts of the MOSNET.}

\subsection{Capturing Spatio-Temporal Features by using 3D Convolutions}
\label{spatiotempfeatures}

Quite a few approaches \cite{Artem, Moez, Ji, TwoStream, 3DFR, LaLonde, Turki} in the field of background subtraction, motion segmentation, and human action recognition demonstrated the benefit of combining spatial and temporal context for tackling computer vision tasks. Simonyan and Zisserman used a \ac{DCNN} to capture spatial information and a \ac{DCNN} to capture temporal information in order to incorporate both results into the final classification. This approach requires optic flow calculation as a pre-processing step and does not capture temporal context in the network itself.\\

3D convolutions gained attention in research since they are able to capture information in two dimension of the image space and in the time space dimension, represented by a number of subsequent images. LaLonde et al. used 3D convolutions to capture motion on the ground with an airborne camera while Turki et al. insert 3D convolutions in their \ac{RNN} architecture to perform background subtraction. Mandal et al. use a set of 25 3D convolutions in their scene independent network for change detection.\\

The MOSNET uses 3D convolutions in the \textit{Low-Level Conv3D}\footnotemark~network and the \textit{High-Level Conv3D}\footnotemark[\value{footnote}] network, which is discussed in detail in the following sections.
\footnotetext{These are designations, which are introduced in the context of this work in order to describe certain parts of the MOSNET.}

\subsection{Architecture Overview}
\label{ArchitectureOverview}

\begin{figure}[h!]
\centering
\includegraphics[width=0.8\textwidth]{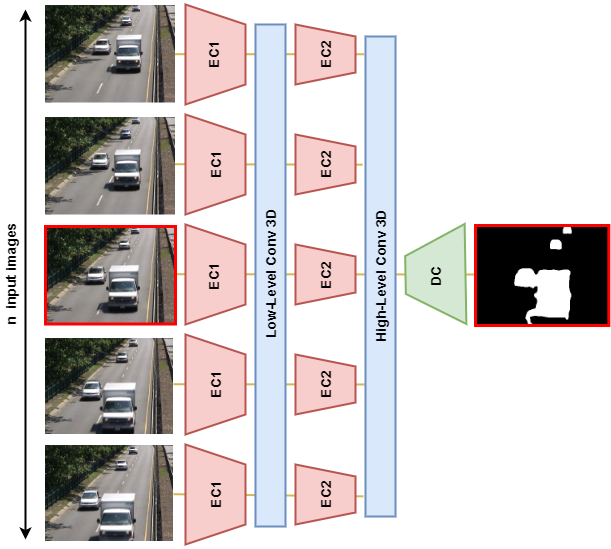}
\caption {Network architecture of the MOSNET including the encoder networks EC1 and EC2, the decoder network DC and the Low- and High-Level Conv3D networks.}
\label{fig:MOSNET_structure}
\end{figure}

Figure \ref{fig:MOSNET_structure} shows the network structure of the MOSNET, including the encoder networks EC1 and EC2, the decoder network DC and the Low- and High-Level Conv3D network layers. A hyperparameter of the MOSNET is $n$, which represents the number of cohesive input images used to predict one motion mask. In addition, the hyperparameter $n$ influences the number of EC1 and EC2 networks, since each input image gets processed by separate EC1 and EC2 networks.
The $n$ input images get encoded by $n$ EC1 networks in order to generate features that have a low level of abstraction but roughly keep the spatial resolutions. These features are further called low-level features.
The $n$ EC2 networks get low-level feature maps and increase the level of abstraction by simultaneously losing spatial resolution. These features with a high abstraction level are further called high-level features.\\

The Low-Level Conv3D and High-Level Conv3D networks perform 3D convolutions, whereby the convolution is performed over image space to capture spatial features and over $n$ cohesive feature maps in order to capture a temporal component. Performing a set of convolution, and pooling operations in the encoder parts of the MOSNET leads to features with a high abstraction level that capture more spatial context than features with a lower level of abstraction. Therefore, the High-Level Conv3D network captures those spatio-temporal features with a high level of spatial abstraction.
However, by convolution and pooling operations, information about the exact position of the features in the image space is lost. To preserve this information, the Low-Level Conv3D network brings in spatio-temporal features with a low level of abstraction and therefore a smaller spatial context but a high accuracy regarding the exact location of the features in image space.   

\FloatBarrier

All layers except for the last layer of the DC network use the \ac{ReLu} activation function. In order to output a binary probability motion mask, the last layer of the MOSNET uses a sigmoid activation function (see Section \ref{activationfunctions}).  Furthermore, there are several Batch Normalizations inserted in order to accelerate the convergence during training of the network (see Section \ref{ExpUsingBatchNorm}).

\subsection{Using Weights and Biases of the VGG-16 Network}
\label{UsingFuck}

Zeiler et al. \cite{Zeiler} researched the evolution of features during the training of a convolutional network. They showed that the time, required for the weights and biases of a layer to converge during training increases the earlier the layer is located in a feed-forward network.\\

In order to overcome this issue, the encoder networks EC1 and EC2 are based on layers of the VGG-16 network, introduced in Section \ref{VGG-16}. In this way, the networks can partly be initialized with pre-trained weights and biases. The weights and biases used in this work to initialize the VGG-16 network based layers originate from \cite{VGGweights}.\\

Since these weights are used in early layers of the network, which capture basic low-level features, the domain gap between image classification and motion segmentation is expected to be sufficiently small. On the contrary, the weights and biases of these layers are not adjusted during backpropagation to maintain the generalization capability of the low-level features generated by training with the large scale dataset ImageNet. In addition, the reduction from about $21$ Mio. to about $19.2$ Mio. trainable parameters safes computational cost\footnote{The number of parameters applies to the network architecture configuration described in Chapter \ref{training}, where $n$ is set to five.}.\\

\section{Encoder EC1}
\label{EC1}

\begin{minipage}{0.40\textwidth}
\includegraphics[width=\textwidth]{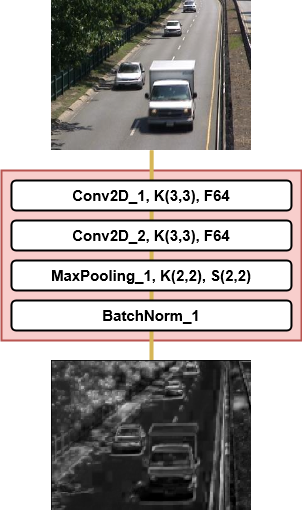}
\captionof{figure}[Layer structure of the EC1 network]{Layer structure of the EC1 network, where each weighted layer is noted with it's kernel size K, followed by it's filters F and stride S (default stride S(1,1) is not noted).\footnotemark}
\label{img:EC1}
\end{minipage}
\hfill
\begin{minipage}{0.55\textwidth}
As discussed in the previous section, the EC1 network is adopted from the first three layers of the VGG-16 network in order to initialize them with the weights and biases of the VGG-16 network. When training the MOSNET, the weights and biases of the EC1 network are not getting adjusted by the backpropagation algorithm. Figure \ref{EC1} shows the layer structure of the EC1 network, consisting of two 2D convolution layers with a kernel size of $3 \times 3$ and 64 filters. In order to reduce the dimensions to a size of $\frac{h}{2} \times \frac{w}{2} \times 64$, Max Pooling is implemented. The output of the EC1 network gets normalized using Batch Normalization in the BatchNorm\_1 layer. The output dimensions of each layer of the EC1 network can be seen in Table \ref{dimEC1}.   
\end{minipage}
\footnotetext{Only one output feature map is shown in order to get a visual impression.}
\FloatBarrier

{\rowcolors{2}{gray!3!}{gray!30!}
\begin{table}[h!]
\centering
\caption{Output dimensions of the layers in the EC1 network.}
\label{dimEC1}
\begin{tabular}{cc}
\toprule
\textbf{Layer} & \textbf{\begin{tabular}[c]{@{}c@{}}Output\\ Dimension\end{tabular}} \\ \midrule
Conv2D\_1     & $h \times w \times 64 $ \\ [0.8ex] 
Conv2D\_2     & $h \times w \times 64 $ \\ [0.8ex] 
MaxPooling\_1 & $\frac{h}{2} \times \frac{w}{2} \times 64 $ \\ [0.8ex] 
BatchNorm\_1 & $\frac{h}{2} \times \frac{w}{2} \times 64 $ \\ [0.8ex] \bottomrule
\end{tabular}
\end{table}
}

\section{Low-Level Conv3D}
\label{LLConv3D}

\begin{minipage}{0.4\textwidth}
\includegraphics[width=\textwidth]{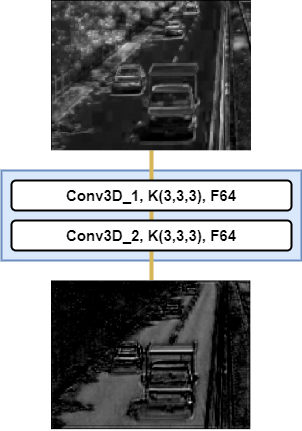}
\captionof{figure}[Layer structure of the Low-Level Conv3D network]{Layer structure of the Low-Level Conv3D network, where each weighted layer is noted with its kernel size K, followed by its filters F and stride S (default stride S(1,1) is not noted).\footnotemark}
\label{img:EC2}
\end{minipage}
\hfill
\begin{minipage}{0.55\textwidth}
The Low-Level Conv3D network gets the output of $n$ EC1 networks, with the dimensions $\frac{h}{2} \times \frac{w}{2} \times 64 $ as an input. Figure \ref{LLConv3D} shows the layer structure of the Low-Level Cov3D network with two layers performing one 3D convolution each. Each of these 3D convolutions share their weights and biases not only in the spatial dimensions, but also across all $n$ input feature maps. In this way, the capturing of spatial and temporal information by performing a convolution over image space with a kernel size of $3 \times 3$ and over time domain with a kernel size of $3$ represented by feature maps of $n$ cohesive input frames can be done. The added value in performance, which is achieved by inserting the Low-Level Conv3D network, is shown in Section \ref{ExpLLConv3D}.
\end{minipage}
\footnotetext{Only one input and output feature map is shown in order to get a visual impression.}
\FloatBarrier

\section{Encoder EC2}
\label{EC2}

The EC2 network gets fed with low-level spatio-temporal features generated by the EC1 and the Low-Level Conv3D network. In order to improve the level of spatial abstraction, further 2D convolution layers are implemented. The layer structure of the EC2 network can be seen in Figure \ref{EC2}. Each convolution layer is denoted with its kernel size K, its filters F and its stride S. The layers $Conv2D\_3$, $Conv2D\_4$, $MaxPooling\_2$, $Conv2D\_5$, $Conv2D\_6$, and $Conv2D\_7$ are adopted from the VGG-16 network, which enables the usage of pre-trained weights and biases from \cite{VGGweights}. The layers, adopted from the VGG-16 network, are not considered during backpropagation in order to decrease computational effort and preserve the generalization ability of the basic features generated in these layers.

\begin{minipage}{0.46\textwidth}
\includegraphics[width=\textwidth]{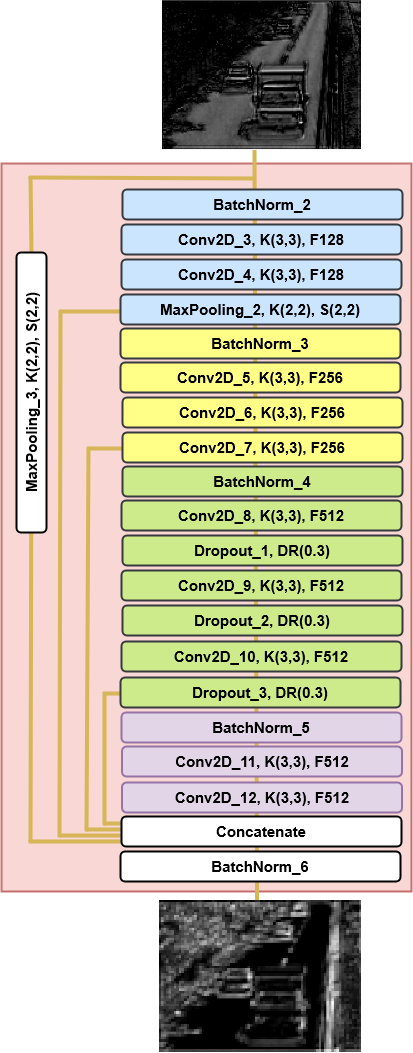}
\captionof{figure}[Layer structure of the EC2 network]{Layer structure of the EC2 network, where each weighted layer is noted with its kernel size K followed by its filters F and stride S (default stride S(1,1) is not denoted).\footnotemark The dropout rate is noted with DR.}
\label{EC2}
\end{minipage}
\hfill
\begin{minipage}{0.49\textwidth}
Since the VGG-16 network was trained to recognize image content in its entirety, the last three layers of the VGG-16 network are fully-connected, which leads to one-dimensional output vector. In contrast to this, the encoder part in an encoder-decoder architecture is designed to output two-dimensional feature maps, which can be decoded by the decoder. Lim et al. used an amended VGG-16 network as an encoder network in their FgSegNet\_v2 \cite{Lim_v2}, FgSegNet\_M \cite{Lim} and FgSegNet\_S \cite{LimS} that produces two-dimensional feature maps. The layers Conv2D\_8, Dropout\_1, Conv2D\_9, Dropout\_2, Conv2D\_10, and Dropout\_3 are based on the approaches of Lim et al.. The layers Conv2D\_11 and Conv2D\_12, MaxPooling\_3 and Concatenate are part of the multi feature map fusion technique, discussed in Section \ref{EC2fusing}.\\

The EC2 network can be divided into four blocks that handle feature maps of different dimensions marked with different colors in Figure \ref{EC2}. At the beginning of the first three blocks (blue, yellow and green), Batch Normalization is implemented to normalize the input of each block. Besides accelerating the convergence during training, Batch Normalization has the side effect of regularizing the network, as mentioned in Section \ref{BatchNorm}. The dropout rate is adjusted from network of Lim et al. and set to a value of $0.3$.\\
\end{minipage}
\footnotetext{Only one input and output feature map is shown in order to get a visual impression.}
\FloatBarrier

Table \ref{dimEC2} shows the output dimensions of each layer in the EC2 network. The network outputs $1,472$ feature maps of size $\frac{h}{4} \times \frac{w}{4}$, which are fed into the High-Level Conv3D network. 

{\rowcolors{2}{gray!3!}{gray!30!}
\begin{table}[h!]
\centering
\caption{Output dimensions of the layers in the EC2 network.}
\label{dimEC2}
\begin{tabular}{cc}
\toprule
\textbf{Layer} & \textbf{\begin{tabular}[c]{@{}c@{}}Output\\ Dimension\end{tabular}} \\ \midrule
BatchNorm\_2   & $\frac{h}{2} \times \frac{w}{2} \times 64 $ \\ [0.8ex] 
Conv2D\_3     & $\frac{h}{2} \times \frac{w}{2} \times 128 $ \\ [0.8ex] 
Conv2D\_4     & $\frac{h}{2} \times \frac{w}{2} \times 128 $ \\ [0.8ex] 
MaxPooling\_2 & $\frac{h}{4} \times \frac{w}{4} \times 128 $ \\ [0.8ex] 
BatchNorm\_3 & $\frac{h}{4} \times \frac{w}{4} \times 128 $ \\ [0.8ex] 
Conv2D\_5     & $\frac{h}{4} \times \frac{w}{4} \times 256 $ \\ [0.8ex] 
Conv2D\_6     & $\frac{h}{4} \times \frac{w}{4} \times 256 $ \\ [0.8ex] 
Conv2D\_7     & $\frac{h}{4} \times \frac{w}{4} \times 256 $ \\ [0.8ex] 
BatchNorm\_4  & $\frac{h}{4} \times \frac{w}{4} \times 256 $ \\ [0.8ex] 
Conv2D\_8     & $\frac{h}{4} \times \frac{w}{4} \times 512 $ \\ [0.8ex] 
Dropout\_1    & $\frac{h}{4} \times \frac{w}{4} \times 512 $ \\ [0.8ex] 
Conv2D\_9     & $\frac{h}{4} \times \frac{w}{4} \times 512 $ \\ [0.8ex] 
Dropout\_2    & $\frac{h}{4} \times \frac{w}{4} \times 512 $ \\ [0.8ex] 
Conv2D\_10    & $\frac{h}{4} \times \frac{w}{4} \times 512 $ \\ [0.8ex] 
Dropout\_3    & $\frac{h}{4} \times \frac{w}{4} \times 512 $ \\ [0.8ex]
BatchNorm\_5   & $\frac{h}{4} \times \frac{w}{4} \times 512 $ \\
[0.8ex]
Conv2D\_11    & $\frac{h}{4} \times \frac{w}{4} \times 512 $ \\ [0.8ex]
Conv2D\_12    & $\frac{h}{4} \times \frac{w}{4} \times 512 $ \\ [0.8ex]
MaxPooling\_3 & $\frac{h}{4} \times \frac{w}{4} \times 64  $ \\ [0.8ex] 
Concatenate   & $\frac{h}{4} \times \frac{w}{4} \times 1,472 $ \\ [0.8ex] 
BatchNorm\_6   & $\frac{h}{4} \times \frac{w}{4} \times 1,472 $ \\ [0.8ex] \bottomrule
\end{tabular}
\end{table}
}
\FloatBarrier

\section{Multiple Feature Map Fusion}
\label{EC2fusing}

In order to improve the handling of objects of different size in the image, multiple feature maps of different abstraction levels and different receptive fields regarding the input frame get concatenated depth-wise in the Concatenate layer.

\begin{minipage}{0.45\textwidth}
To do so, the input feature maps of the EC2 network get pooled with a Max Pooling layer (MaxPooling\_3), in order to decrease the dimension of the input feature maps of the EC2 network to a size of $\frac{h}{4} \times \frac{w}{4} \times 64$. This enables the depth-wise concatenation of feature maps generated by the Conv3D\_2, MaxPooling\_2, Conv2D\_7, Dropout\_3 and Conv2D\_12 layer in order to fuse feature maps with different levels of abstraction and different receptive fields regarding the input of the network. Figure \ref{fig:Concatiation} illustrates the depth-wise concatenation of $1,472$ feature maps in the  Concatenate layer of the EC2 network. Concatenating these feature maps prepares them to be fused in the following High-Level Conv3D network. This aims to ensure that the network is able to find an appropriate compromise of these parameters during the training itself.
\end{minipage}
\hfill
\begin{minipage}{0.5\textwidth}
\begin{center}
\includegraphics[width=\textwidth]{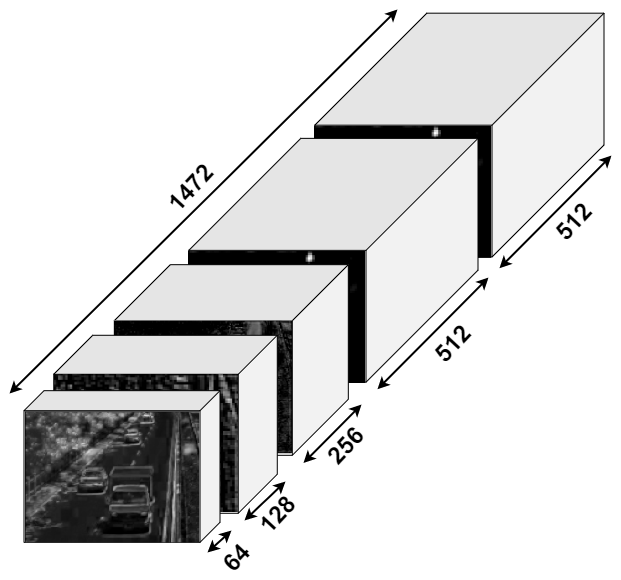}
\captionof{figure}[Illustration of concatenating feature maps depth-wise in the Concatenation layer]{Illustration of concatenating feature maps depth-wise in the Concatenation layer.}
\label{fig:Concatiation}
\end{center}
\end{minipage}
\FloatBarrier

\subsection{Fusing Different Abstraction Levels}
\begin{figure}[h!]%
  \centering
  \subfloat[][]{\includegraphics[width=0.15\textwidth]{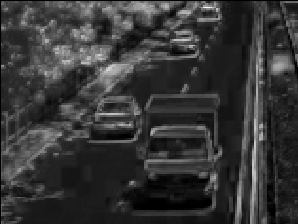}}%
  \qquad
  \subfloat[][]{\includegraphics[width=0.15\textwidth]{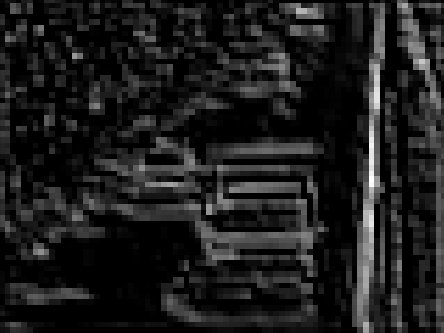}}%
  \qquad
  \subfloat[][]{\includegraphics[width=0.15\textwidth]{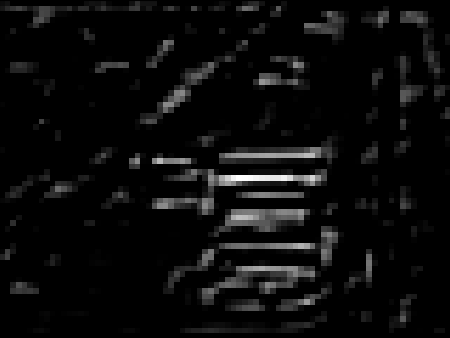}}%
  \qquad
  \subfloat[][]{\includegraphics[width=0.15\textwidth]{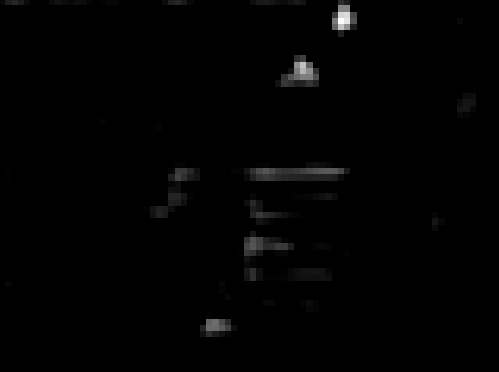}}%
  \qquad
  \subfloat[][]{\includegraphics[width=0.15\textwidth]{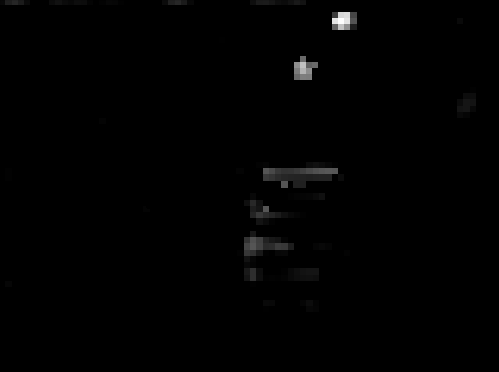}}%
  \qquad
  \caption[Illustration of the different abstraction levels concatenated depth-wise in the last layer of EC2 network]{Illustration of the different abstraction levels, concatenated depth-wise in the last layer of EC2 network. The 64 feature maps, outputted by the MaxPooling\_3 layer (a), the 128 feature maps outputted by the MaxPooling\_2 layer (b), the 256 feature maps, generated by the Conv2D\_7 (c), the 512 feature maps generated by the Dropout\_3 layer (d), and the 512 feature maps generated by the Conv2D\_12 layer (e) are concatenated before getting fed into the High-Level Conv3D network.\footnotemark} %
  \label{Scales}
\end{figure}

Figure \ref{Scales} gives an impression of the abstraction level's evolution in the EC2 network. It can be seen that the level of abstraction increases from (a) to (e), while the resolution decreases. In order to get a visual impression, only one of the 64 (resp. 128, resp. 256, resp. 512) feature maps is illustrated. 

\FloatBarrier
\footnotetext{Only one feature map of each abstraction level is shown in order to get a visual impression.}

\subsection{Fusing Different Receptive Fields}
The effective receptive field of \acp{CNN} is significantly smaller than the theoretical receptive field \cite{Zhou2}. To be more precise, \emph{\say{[...] the distribution of impact within the receptive field is asymptotically Gaussian, and the effective receptive field only takes up a fraction of the full theoretical receptive field.}} \cite{Luo} Feature maps generated by different convolutional layers of the MOSNET have different receptive fields regarding the input of the network. Fusing them, enables the network to learn the appropriate receptive field by itself by combing some of them. 

Table \ref{tab:receptive} lists the size of the receptive field of the layers, which are considered in the feature map fusion. A table showing the evolution of the receptive field in the encoder part of the MOSNET is appended in Appendix \ref{Appevolution}. The size of the receptive field is calculated as introduced in \cite{receptivefield}. 

{\rowcolors{2}{gray!3!}{gray!30!}
\begin{table}[h!]
\centering
\caption{Size  of the receptive field, considered during feature map fusion.}
\label{tab:receptive}
\begin{tabular}{cc}
\toprule
\textbf{Layer} & \textbf{Size of receptive field} \\ \midrule
\textcolor{red}{MaxPooling\_3}  & $16 \times 16$                                          \\ 
\textcolor{OliveGreen}{MaxPooling\_2}  & $24 \times 24$                                          \\ 
\textcolor{Dandelion}{Conv2D\_7}      & $48 \times 48$                                         \\ 
\textcolor{purple}{Dropout\_3}     & $72 \times 72$                                       \\ 
\textcolor{blue}{Conv2D\_12}     & $88 \times 88$                                       \\ \bottomrule
\end{tabular}
\end{table}
}

\begin{minipage}{0.5\textwidth}
Figure \ref{fig:highwayRF} illustrates the scale of receptive fields listed in Table \ref{tab:receptive} on an input frame of the CDNet2014 dataset. The MOSNET can learn the amount of spatial context needed by weighting the feature maps with differing receptive fields. 
A related approach was first introduced by Zhao et al. \cite{ZhaoPyramid} to tackle the task of semantic segmentation by fusing several feature maps to capture local and global context information in an image. A comparison of the performance of the MOSNET with and without fusing feature maps with different abstraction level and receptive field regarding the input of the network is performed in Section \ref{ExpFeaturemapfusing}.\\
\end{minipage}
\hfill
\begin{minipage}{0.4\textwidth}
\begin{center}
\includegraphics[width=\textwidth]{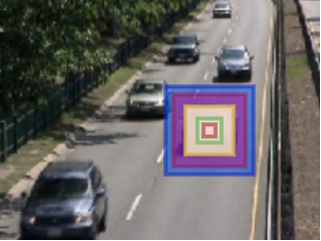}
\captionof{figure}[Illustration of the receptive fields]{Illustration of the receptive fields, listed in \mbox{Table \ref{tab:receptive}} on a frame of the CDNet2014 dataset.}
\label{fig:highwayRF}
\end{center}
\end{minipage}
\FloatBarrier

\section{High-Level Conv3D}
\label{HLConv3D}

\begin{minipage}{0.40\textwidth}
\includegraphics[width=\textwidth]{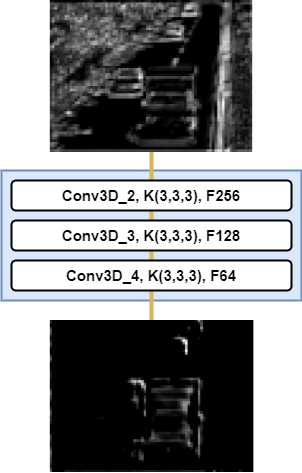}
\captionof{figure}[Layer structure of the High-Level Conv3D network]{Layer structure of the High-Level Conv3D network, where each weighted layer is noted with its kernel size K followed by its filters F and stride S (default stride S(1,1) is not denoted). The dropout rate is noted with DR \footnotemark.}
\label{img:EC2}
\end{minipage}
\hfill
\begin{minipage}{0.55\textwidth}
The High-Level Conv3D network  consists of three convolutional layers, while each of them performing a 3D convolution. Sharing all weights and biases in the image space and temporal domain allows the capturing of spatio-temporal features with a high level of abstraction. The number of filters $F$ is decreasing by a factor of two from $256$ to $128$ to $64$, in order to perform a smooth transition between the $1,472$ feature maps of the EC2 networks and the $64$ feature maps of the DC network (see Section \ref{Decoder}).
\end{minipage}
\footnotetext{Only one input and output feature map is shown in order to get a visual impression.}
\FloatBarrier

\section{Decoder DC}
\label{Decoder}

The decoder network, called DC, performs a set of transposed convolutions (see Section \ref{ConvT}) aiming at generating feature maps with the same dimensions as the input frame. The decoder reduces the level of abstraction, while reconstructing the spatial resolution of the spatio-temporal features generated by the previous networks.        

\begin{minipage}{0.4\textwidth}
\includegraphics[width=0.9\textwidth]{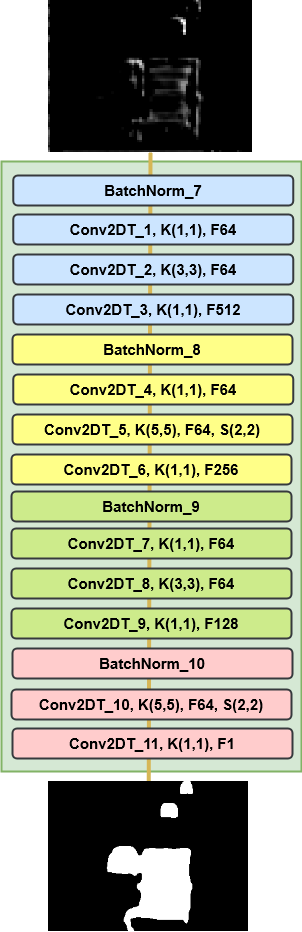}
\captionof{figure}[Layer structure of the DC network]{Layer structure of the DC network, where each weighted layer is noted with its kernel size K followed by its filters F and stride S (default stride S(1,1) is not denoted).\footnotemark}
\label{img:DC}
\end{minipage}
\hfill
\begin{minipage}{0.55\textwidth}
The DC network consists of eleven convolutional layers which are adopted from the FgSegNet\_M \cite{Lim}, while each of these layers perform a transposed 2D convolution. Applying a stride of two in both spatial dimensions in the layers Conv2DT\_5 and Conv2DT\_10, enlarges the feature maps by the factor of two in each dimension. This leads to a feature map size of $h \times w$ after the Conv2DT\_10 layer. Finally, the Conv2DT\_11 layer performs a convolution with one filter and a kernel size of $1 \times 1$, which fuses the 64 feature maps into one feature map, further called motion mask. Additionally, the Conv2DT\_11 layer is the only layer in the MOSNET which does not have a \ac{ReLu} activation function. Since \ac{MOD} is a binary classification task, the sigmoid activation function (see Section \ref{activationfunctions}) is implemented in order to generate a probability of each pixel belonging to a moving object. 
The DC network can be divided in four blocks marked with the colors blue, yellow, green and red in Figure \ref{img:DC}. At the beginning of each block, the inputs get normalized in order to accelerate the training. The first convolutional layer of the first three blocks (Conv2DT\_1, Conv2DT\_4, Conv2DT\_7) use a kernel size of $1 \times 1$ in order to increase the non-linearity of the decision function and project the high dimensional feature map depth into a lower dimension of 64 feature maps \cite{Lim}. 
\end{minipage}
\footnotetext{Only one input feature map is shown in order to get a visual impression.}
\FloatBarrier

To alleviate overfitting due to the strong change in the number of feature maps, Lim et al. regularized the kernel weights of the layers Conv2DT\_4 and Conv2DT\_7 with an $\ell_2$-norm (see Section \ref{regularization}). The second and third convolutional layer of each block performs a transposed convolution with a kernel size of $3 \times 3$ or $5 \times 5$ and enlarge the number of feature maps along the depth axis afterwards \cite{Lim}. The four blocks of the DC network slowly reduce the number of feature maps from 512 down to 1, which then represents the resulting motion mask of the MOSNET. The layers Conv2DT\_5 and Conv2DT\_10 use a stride of two in both spatial dimensions to increase the size of the feature maps.  Table \ref{dimDC} shows the output dimensions of each layer in the DC network.

{\rowcolors{2}{gray!3!}{gray!30!}
\begin{table}[h!]
\centering
\caption{Output dimensions of the layers of the DC network}
\label{dimDC}
\begin{tabular}{cc}
\toprule
\textbf{Layer} & \textbf{\begin{tabular}[c]{@{}c@{}}Output\\ Dimension\end{tabular}} \\ \midrule
BatchNorm\_7 & $\frac{h}{4} \times \frac{w}{4} \times 64 $  \\ [0.8ex] 
Conv2DT\_1 & $\frac{h}{4} \times \frac{w}{4} \times 64 $  \\ [0.8ex] 
Conv2DT\_2 & $\frac{h}{4} \times \frac{w}{4} \times 64 $  \\ [0.8ex] 
Conv2DT\_3 & $\frac{h}{4} \times \frac{w}{4} \times 512 $ \\ [0.8ex] 
BatchNorm\_8 & $\frac{h}{4} \times \frac{w}{4} \times 512 $ \\ [0.8ex]
Conv2DT\_4 & $\frac{h}{4} \times \frac{w}{4} \times 64 $  \\ [0.8ex] 
Conv2DT\_5 & $\frac{h}{2} \times \frac{w}{2} \times 64 $  \\ [0.8ex] 
Conv2DT\_6 & $\frac{h}{2} \times \frac{w}{2} \times 256 $ \\ [0.8ex] 
Conv2DT\_7 & $\frac{h}{2} \times \frac{w}{2} \times 64 $  \\ [0.8ex] 
BatchNorm\_9 & $\frac{h}{2} \times \frac{w}{2} \times 64 $ \\ [0.8ex]
Conv3DT\_8 & $\frac{h}{2} \times \frac{w}{2} \times 64 $ \\ [0.8ex] 
Conv2DT\_9 & $\frac{h}{2} \times \frac{w}{2} \times 128 $ \\ [0.8ex]
Conv2DT\_10    & $h \times w \times 64 $ \\ [0.8ex]
BatchNorm\_10    & $h \times w \times 64 $ \\ [0.8ex]
Conv2DT\_11    & $h \times w \times 1 $ \\ [0.8ex] \bottomrule
\end{tabular}
\end{table}
}
\FloatBarrier

\section{Number of Parameters}
\label{AnzParams}

This section shows the distribution of the MOSNETs parameters on the individual layers.\\ 

In a convolution layer the weights of the convolution matrix get shared over the whole input data in order to generate a feature map. Therefore, the number of parameters to generate one feature map is the number of elements in the convolution matrix multiplied with the depth of the input. In example, looking at the first layer of the VGG-16 network, which generates a feature map with kernel size of $3 \times 3$ on an \ac{RGB} image leads to $3 \cdot 3 \cdot 3 = 27$ parameters in order to obtain one feature map. Since the first layer of the VGG-16 network generates $64$ feature maps, the number of parameters are multiplied by the number of filters (i.e. $27 \cdot 64 = 1,728$) in order to get the number of trainable weights of the layer. As every filter having a bias parameter (see Equation \ref{eq:neuron}), the number of filters are added (i.e. $1,728 + 64 = 1,792$) to calculate the number of parameters of the convolution layer.\\
As mentioned in \cite{batchnorm}, Batch Normalization calculates four parameters for each input feature map. Therefore the number of parameters can be calculated by multiplying the number of input feature maps by four.
Layers performing Dropout or Max Pooling do not have parameters, since they are not based on additional neurons.\\

Table \ref{Parapara} shows the parameters, i.e. all weights and biases the layers in the MOSNET. As mentioned in Section \ref{UsingFuck} some weights and biases are not adjusted during training and therefor are not trainable. It can be observed that more than half of the overall parameters pertain to the Conv3D\_3 layer. Implementing the multi feature map fusion technique, increases the number of parameters of the Conv3D\_3 layer from $3,539,200$ to $10,174,720$ parameters\footnote{$3 \cdot 3 \cdot 3 \cdot 512 \cdot 256 + 256 = 3,539,200$ and $3 \cdot 3 \cdot 3 \cdot 1,472\cdot 256 + 256 = 10,174,720$}. When performing experiments on the network architecture in Section \ref{ExpMOSNET}, the number of parameters of the network, and therefore its capacity may change, which must be considered when interpreting the results. This is discussed in detail in Chapter \ref{ExpandRes}.   

{\rowcolors{2}{gray!3!}{gray!30!}
\begin{table}[h!]
\centering
\caption{Parameters listed per layer.}
\label{Parapara}
\begin{tabular}{lrc||lrc}
\toprule
Layer        & Parameter & Trainable & Layer         & Parameter  & Trainable \\ \midrule
Conv2D\_1    & 1,792     &   -        & BatchNorm\_6  & 5,888      &    \checkmark       \\
Conv2D\_2    & 36,928    &   -        & Conv3D\_3     & 10,174,720 &    \checkmark       \\
BatchNorm\_1 & 256       &   \checkmark        & Conv3D\_4     & 884,864    &   \checkmark        \\
Conv3D\_1    & 110,656   &  \checkmark         & Conv3D\_5     & 221,248    &  \checkmark         \\
Conv3D\_2    & 110,656   &  \checkmark         & BatchNorm\_7  & 256        &  \checkmark         \\
BatchNorm\_2 & 256       &  \checkmark         & Conv2DT\_1    & 4,160      &  \checkmark         \\
Conv2D\_3    & 73,856    &  -         & Conv2D\_2     & 36,928     &   \checkmark        \\
Conv2D\_4    & 147,584   &  -         & Conv2DT\_3    & 33,280     &   \checkmark        \\
BatchNorm\_3 & 512       &  \checkmark         & BatchNorm\_8  & 2,048      &   \checkmark        \\
Conv2D\_5    & 295,168   &   -        & Conv2DT\_4    & 32,832      &   \checkmark        \\
Conv2D\_6    & 590,080   &      -     & Conv2DT\_5    & 102,464     &   \checkmark        \\
Conv2D\_7    & 590,080   &  -         & Conv2DT\_6    & 16,640      &   \checkmark        \\
BatchNorm\_4 & 1024      &    \checkmark       & BatchNorm\_9  & 1,024       &   \checkmark        \\
Conv2D\_8    & 118,0160  &    \checkmark       & Conv2DT\_7    & 16,448      &   \checkmark        \\
Conv2D\_9    & 2,359,808 & \checkmark          & Conv2DT\_8    & 36,928      &   \checkmark        \\
Conv2D\_10   & 2,359,808 & \checkmark          & Conv2DT\_9    & 8,320       &   \checkmark        \\
BatchNorm\_5 & 2,048     & \checkmark          & BatchNorm\_10 & 512        &   \checkmark        \\
Conv2D\_11   & 2,359,808 & \checkmark          & Conv2DT\_10   & 204,864     &   \checkmark        \\
Conv2D\_12   & 2,359,808 & \checkmark          & Conv2DT\_11   & 65         &   \checkmark\\
\bottomrule
\end{tabular}
\end{table}
}

\chapter{Training}
\label{training}

The process of training the MOSNET including the hyperparameter setting, the training strategy, and the objective function is introduced in this chapter. Additionally, an optional pre-processing step is introduced, which aims at achieving a better performance at scenes captured with a moving camera.

\section{Hyperparameters}
\label{hyperparameters}

Besides the network architecture itself, hyperparameters like the learning rate used during training have a major effect on the performance of the approach. Since these hyperparameters are determined empirically and partly influence each other, it is difficult to find the global optimum, which leads to the best performance. Therefore, a limited selection of hyperparameters is empirically optimized (see Section \ref{ExpMOSNET}) in this work. The majority of the hyperparameters are implemented with standard values from the literature and comparable approaches like the FgSegNet\_M \cite{Lim} such as the regularization factor $l$ of the $\ell_2$-norm.\\

Table \ref{tab:hyperparameter} shows the hyperparameter setting used in this work. One mini-batch is used to calculate the stochastic objective function for one training step. The size of the mini-batch is a trade-off between hardware constraints and the variance of the stochastic gradient (see also \cite{Goodfellow}, pp. 274-276). The mini-batch size is set to two, which is the maximum number of mini-batches to fit in the \ac{GPU} memory\footnote{The hardware used in this work is described in Section \ref{ExpToolingHardware}.}. Therefore, the network is fed with two blocks of $n$ input frames for one training step. The optimizer used to calculate the gradient of the stochastic objective function is a state-of-the-art method for optimization called \ac{Adam} \cite{Adam}. The survey on gradient descent optimization algorithms of Ruder \cite{Ruder} mentioned the Adam optimizer as the method of choice since it uses adaptive learning rate methods, which makes the use of a learning rate schedule\footnote{A learning rate schedule tends to adjust the learning rate during training to obtain a trade-off between learning progress and overshooting the optimum of the objective function. } needless during training. The  hyperparameters of the optimizer are initialized with the values recommended in the corresponding paper \cite{Adam}. Section \ref{ExpNLookback} shows the empirical determination of the hyperparameter $n$ with $n=5$ showing the best results. 

{\rowcolors{2}{gray!3!}{gray!30!}
\begin{table}[h!]
\begin{center}
\caption{Hyperparameters}
\label{tab:hyperparameter}
\begin{tabular}{ccc}
\toprule
\textbf{Description}        & \textbf{Hyperparameter} & \textbf{Value}             \\ \midrule
Size of mini-batch           & $\text{mini-batch size}$              & 2                          \\
Number of input frames      & $n$                       & 5                          \\
Initial learning rate       & $lr$                      & 0.0001                           \\
Regularization factor of the $\ell_2$-norm       & $l$                      & 0.0005                           \\ \bottomrule
\end{tabular}
\end{center}{}
\end{table}
}

\section{Training Strategy}
\label{trainingstrategy}

The network is trained for a maximum of $30$ epochs using the training data $D_T$. Before starting an epoch, the training data is shuffled to avoid biasing of the network during training due to a specific order of the data in the mini-batches, which is explained in detail in Section \ref{datahandling}.\\

After each epoch of training, the trained model gets validated using the validation data $D_V$ in order to calculate the validation F-measure and the validation loss. As the validation loss indicates the performance of the network on unseen data, it can be used to observe if the network is overfitting. When there is no further decrease of the validation loss the training stops, which explained in Section \ref{monitoring}. 

\subsection{Data Handling}
\label{datahandling}

The input frames (\ac{GT} frames, respectively) of a scene of the data used during \mbox{development} $(D_T$ and $D_V)$ are loaded in sequential order and then divided into blocks of $n$ cohesive input frames each. The scene is now represented by a 5D tensor containing the input frames of the scene, and a 5D tensor of corresponding \ac{GT} frames with the dimensions, while $c$ denotes the frame channels

\begin{equation}
[number\:of\:blocks\;,\;n\;,\;h\;,\;w\;,\;c],
\notag
\label{eq:datahandeling}
\end{equation}
\FloatBarrier

with

\begin{equation}
number \: of \: blocks = \Bigl\lfloor\frac{number \:of\: frames}{n}\Bigr\rfloor.
\notag
\label{eq:datahandeling}
\end{equation}
\FloatBarrier

In the case of a 5D tensor of \ac{RGB} input frames there are three frame channels, while a 5D tensor of \ac{GT} frames has one frame channel. 
The 5D tensor of input frames of every scene is concatenated along the first dimension in order to get a 5D tensor of input frames including all scenes. At the beginning of each epoch, the resulting 5D tensors containing all input frames, and all corresponding \ac{GT} frames, respectively, get shuffled at the first dimension without losing their assignment. Before starting with the first epoch, the data used during development is split into $80\,\%$ training data $D_T$ and $20\,\%$ validation data $D_V$.\\

\begin{figure}[h!]
\centering
\includegraphics[width=0.6\textwidth]{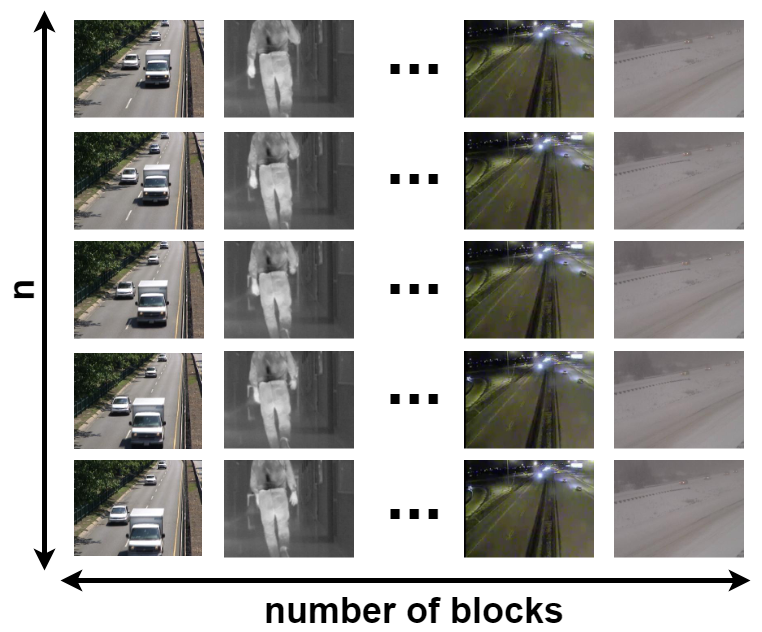}
\caption [Illustration of a 5D tensor of input frames]{Illustration of the 5D tensor of input frames before training.}
\label{fig:5DTensor}
\end{figure}

Since the architecture of the MOSNET is fully convolutional, the network is able to handle frames of various size. For a better utilization of the \ac{GPU} memory, the input and \ac{GT} frames are resized to $240 \times 320$. Figure \ref{fig:5DTensor} illustrates the two dimensions $number\:of\: blocks$ and $n$ of the 5D tensor of input frames, while each frame consists of the dimensions $h$, $w$ and $c$. The 5D tensor of input frames gets sliced along the first dimension in order to form mini-batches.

\FloatBarrier

\paragraph{Handling Pixels Labeled as Outside \ac{ROI} or  Hard Shadow}
The pixels in the \ac{GT} of the \mbox{CDNet2014} dataset are annotated as mentioned in Section \ref{Annotation}. Besides the labels \textit{motion} and \textit{static}, the annotation labels pixels as being \textit{unknown}, \textit{outside \ac{ROI}} and \textit{hard shadow}. Also the relabeled LASIESTA dataset, introduced in Section \ref{LASIESTA}, distinguishes between pixels being motion, pixels being static and unknown pixels.\\

In order to handle pixels labeled as outside \ac{ROI} or  hard shadow the \ac{GT} data used during development gets relabeled before training, and validating, respectively. A label called \textit{ignore label} gets used to relabel the \ac{GT} as follows:

\begin{itemize}
    \item[] $Hard \;shadow \rightarrow Static$
    \item[] $Unknown \rightarrow Ignore\; label$
    \item[] $Outside\; \ac{ROI} \rightarrow Ignore\; label$
\end{itemize}

The ignore label is a self defined label, which labels all pixels, which should not be taken into consideration when calculating the training and validation loss. The handling of the ignore label before calculating the objective function is described in detail in Section \ref{lossfunction}.  

\subsection{Training Monitoring}
\label{monitoring}

As mentioned in Section \ref{regularization}, the best performance on unseen data can be achieved when the network is neither underfitted nor overfitted. During the validation after each epoch, a validation loss on the validation data $D_V$ is calculated to evaluate the networks performance on unseen data\footnote{In this context, unseen data is data which was not previously used to adjust the weights and biases.}. The training process is continued until the validation loss starts to increase, or more precisely if there is no improvement in the validation loss over the previous $3$ epochs\footnote{This time window before stopping the training process was defined empirically}. This regularization method, called \textit{Early Stopping}, prevents the automatized training process from overfitting the network.

\section{Pre-Processing}
\label{Preprocessing}

As motivated in Section \ref{sec:Motivation}, central challenges of this work are the handling of video sequences captured by a camera that pans, tilts, zooms or undergoes a translatory movement.
In an attempt to overcome these challenges, feature based image alignment is implemented as a pre-processing step. 

A block of input frames consists of $n$ frames ($x_{t-\frac{n-1}{2}}, \dots, x_t, \dots ,x_{t+\frac{n-1}{2}}$) while all frames of a block get aligned on the target input frame $x_{t}$. 

To do so, a set of $5000$ keypoints that occur in the target input frame $x_t$ and the aligning frame (e.g. $x_{t-1}$) get detected by using the \ac{ORB} feature detector, introduced by Rublee et al. \cite{ORB}. $10\%$ of the keypoints with the highest similarity regarding their hamming distance are selected to calculate a robust homography matrix $H$. After calculating $H$, the input frames $x_{t-1}$ and  $x_t$ get aligned by multiplying each pixel of $x_{t-1}$ with $H$.\\

\begin{figure}[h!]
\centering
\includegraphics[width=\textwidth]{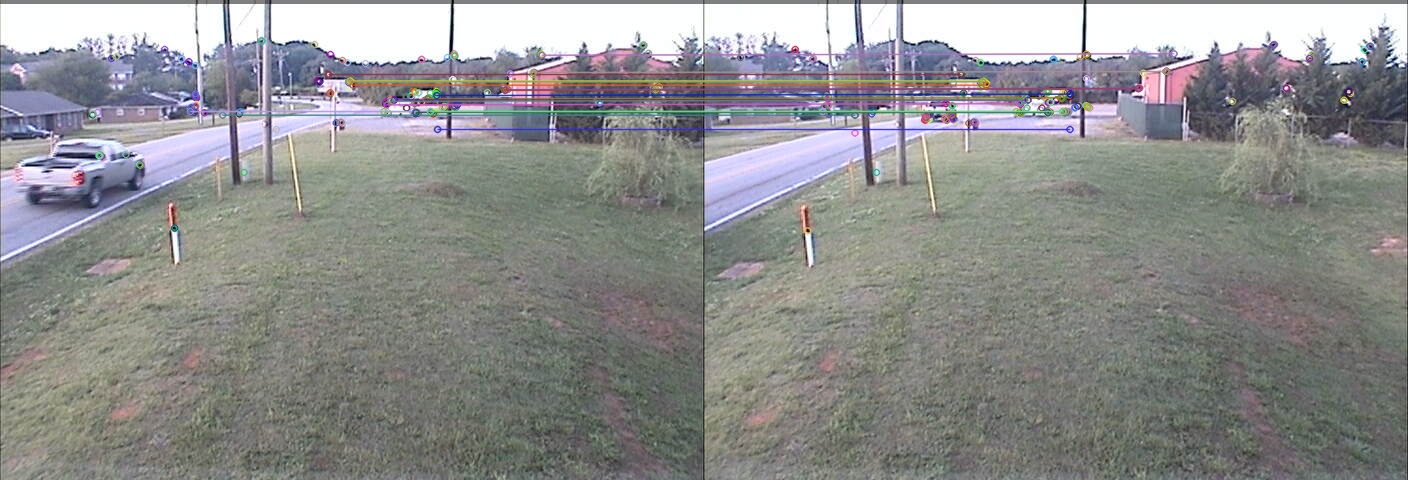}
\caption {Illustration of keypoint detection and matching in a panned scene as being part of the feature based image alignment pre-processing step.}
\label{fig:matches}
\end{figure}

Figure \ref{fig:matches} shows two frames taken from the scene $continuousPan$ of the CDNet2014 dataset's $PTZ$ category between which the camera had been panned to the right. The ORB feature detector recognizes keypoints, in both images and rates their similarity. The best $10\%$ are shown in Figure \ref{fig:matches}. Using these keypoints, the homography matrix $H$ is calculated and each pixel on the left image is multiplied with $H$ in order to result in an aligned image, shown in Figure \ref{fig:aligned}.

\FloatBarrier

\begin{minipage}{0.4\textwidth}
Figure \ref{fig:aligned}, shows the left image of Figure \ref{fig:matches}, aligned on the right image of Figure \ref{fig:matches}. The panning camera mainly brings a horizontal translation in image space, which results in missing content information on the right side of the aligned image. Calculating the homography matrix, defined in equation \eqref{eq:H}, for the example in Figure \ref{fig:matches}, leads to the matrix shown in Equation \eqref{eq:Hexample}.
\end{minipage}
\hfill
\begin{minipage}{0.5\textwidth}
\includegraphics[width=\textwidth]{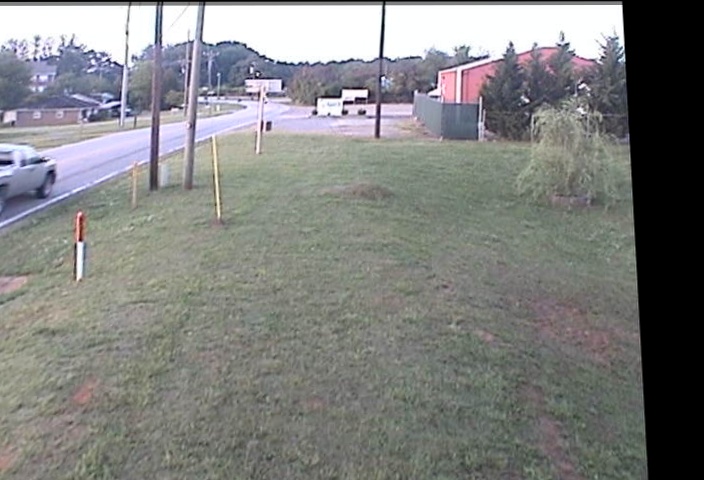}
\captionof{figure}{Aligned image based on the features shown in Figure \ref{fig:matches}.}
\label{fig:aligned}
\end{minipage}
\FloatBarrier

\begin{equation}
H= \left[\begin{matrix}h_{00}&h_{01}&h_{02}\\h_{10}&h_{11}&h_{12}\\h_{20}&h_{21}&h_{22}\end{matrix}\right]
\label{eq:H}
\end{equation}

\begin{equation}
H = \left[\begin{matrix}1.00&-0.15&-70.32\\0.11&1.00&2.00\\0.00&0.00&1.00\end{matrix}\right]
\label{eq:Hexample}
\end{equation}

Having a homography matrix, where $h_{00} = h_{11} = h_{22}=1$, the translation along the horizontal axis and the translation along the vertical axis can be directly read off $h_{02}$, and $h_{12}$, respectively (\cite{Hubert} pp. 57-62). The rotation and scaling that the image undergoes can be derived from $h_{00}$, $h_{01}$, $h_{10}$, and $h_{11}$ \cite{Collins}.
Due to the translation in horizontal and vertical direction, a rotation, as well as a change in scale by the homography transformation, artifacts on the borders of the image may arise, illustrated in black color in Figure \ref{fig:aligned}. These areas get labeled with the ignore label, introduced in Section \ref{datahandling} and therefore are not considered when calculating the objective function during neural network training and validation.  
\\    

The precondition of aligning images using an affine transformation is both images having the same planar surface in image space \cite{Collins}. Even though the images that are aligned do not originate from planar surfaces in the scene space, the implementation of image alignment enhances the performance of the MOSNET in some scenes as further discussed in Section \ref{ExpPreprocessing}.

\section{Objective Function}
\label{lossfunction}

The objective function, also called loss function $\mathcal{L}$, maps the performance inaccuracy of the network to a function with respect to all weights and biases by comparing the predicted output $\hat{y_{t}}$ and the \ac{GT} $y_{t}$ at timestep $t$. The aim is to minimize the objective function by using an optimization algorithm and improve the performance of the network by adjusting the weights an biases. In this section, two loss functions namely binary cross-entropy loss (Equation \eqref{eq:crossentropy}) and Focal Loss (Equation \eqref{eq:focalloss}) get introduced. Additionally, the handling of pixels labeled with the ignore label before calculating the objective function is explained in this section. \\

As $y_p \in \{0,1\}$, specifying the class of a single pixel in the \ac{GT} and $p_1 \in [0,1]$ specifying the probability of a predicted pixel being class $y_p = 1$, $p_t$ can be denoted as followed:

\[ p_t =
  \begin{cases}
    p_1       & \quad \text{if } y_p = 1\\
    1-p_1     & \quad \text{otherwise}
  \end{cases}
\]

\paragraph{Binary Cross Entropy}
The binary cross-entropy loss is commonly used in classification tasks where the output exclusively belongs to one of two classes. In the case of motion detection, a predicted pixel can either belong to a moving object or being static.\\

Using $p_t$, the binary cross-entropy can be noted as follows:
\begin{equation}
\mathcal{L}_\text{Binary CE}(p_t) = - \log_2(p_t)
\label{eq:crossentropy}
\end{equation}

\paragraph{Focal Loss}
In 2018, Lin et al. \cite{Yi} introduced the Focal Loss (equation \eqref{eq:focalloss}), which is a modification of the binary cross-entropy loss, in order to consider the imbalance in some classification tasks. The \emph{\say{[...] Focal Loss focuses training on a sparse set of hard examples and prevents the vast number of easy negatives from overwhelming [...] during training}} \cite{Yi}. Lin et al. introduced the focusing parameter $\gamma \geq 0$ and a parameter $\alpha \in [0,1]$ to improve numerical stability. They empirically determined that setting $\gamma = 2$ and $\alpha = 0.25$ leading to the best results \cite{Yi}. 

\begin{equation}
\mathcal{L}_\text{Focal Loss}(p_t) = -\alpha (1-p_t)^\gamma \log_2(p_t)
\label{eq:focalloss}
\end{equation}

A comparison of the performance of both objective functions when training the MOSNET is made in Section \ref{ExpFocalLoss}.\\

\paragraph{Handling Ignore Labels}
In Section \ref{datahandling}, the relabeling when loading the \ac{GT} frames of the CDNet2014 and the LASIESTA dataset gets discussed. To disregard pixels labeled with the ignore label when calculating the objective function, the following algorithm gets implemented before calculating the objective function: 

\begin{algorithm}
    \DontPrintSemicolon
    \SetKwInOut{KwIn}{Input}
    \SetKwInOut{KwOut}{Output}
    \tcc{Preparing $\hat{y}_t$ and $y_t$ in order to ignore pixels labeled as unknown or outside \ac{ROI} in Section \ref{datahandling}}
    \BlankLine
    motion = 1\;
    static = 0\;
    \BlankLine
    \KwIn{$\hat{y}_t = [0, 1]$, $y_t = \{ static,\: motion,\: ignore \;label \}$}\;
    \KwOut{$\hat{y}_t = [0, 1]$, $y_t = \{ static,\: motion \}$}\;
    \For{pixel $\leftarrow$ $y_t$}{
        \If{$pixel = ignore\; label$}{
            i = get\_index\_of\_pixel\;
            delete pixel of $y_t$ at index i\;
            delete pixel of $\hat{y}_t$ at index i\;
         }
    }
    \KwRet{$\hat{y}_t$, $y_t$}
    \caption{Prepare $\hat{y}_t$ and $y_t$ before calculating the objective function}
    \label{algoROI_2}
\end{algorithm}

Algorithm \ref{algoROI_2} gets the predicted motion mask $\hat{y}_t$ including pixels with a probability of being either static or motion and the \ac{GT} frame $y_t$ at time step $t$ including pixels labeled as static, motion or with the ignore label. Pixels labeled with the ignore label in $y_t$ and all corresponding pixels in the predicted motion mask $\hat{y}_t$ are deleted.

\chapter{Experiments and Results}
\label{ExpandRes}

This chapter points  out the enhancement in performance of capturing spatio-temporal features in the Low-Level Conv3D network, the fusion of feature maps in the EC2 network, the usage of batch normalization and  adjusting the hyperparameter $n$ in order to state reasons for the MOSNET's architecture as introduced in Chapter \ref{Architecture}.
Further sections demonstrate the improvement of using a weighted objective function and the implementation of a pre-processing step.\\

Since it can be assumed that there is no strong interdependence between the individual experiments described in this section, each experiment is compared with the MOSNET setup described in Section \ref{ExpSetup} without evaluating any combination and therefore any influence of the experiments on each other. Furthermore, the experiments are not mono-causal\footnote{For example adjusting the network architecture to perform no feature map fusion, changes the number of trainable parameters and therefore capacity of the network significantly as mentioned in Section \ref{Parapara}.}, which prevents a trace back to a single cause. Nevertheless, they can be used to underpin design decisions, made in Chapter \ref{Architecture} and Chapter \ref{training}.

\section{Experimental Setup}
\label{ExpSetup}

Unless stated otherwise, the experiments in Section \ref{ExpMOSNET} are conducted with the MOSNET architecture introduced in Chapter \ref{Architecture} with $n$ set to five. The MOSNET gets trained using the hyperparameters mentioned in Section \ref{hyperparameters} and the training strategy expound in Section \ref{trainingstrategy},with the focal loss objective function and \textbf{no} pre-processing. The global threshold $T$, introduced in Section \ref{ExpThresholding}, is set to $0.4$. The dataset used for training, validation and evaluation is the CDNet2014 dataset with the dataset split described in Section \ref{Datasetsplit}. While the data $D_T$ and $D_V$ is used to develop (i.e. train end validate) the MOSNET, the data $D_E$ is used to perform the experiments and evaluate the approach.\\

Since the MOSNET uses an odd number $n$ of input frames $x_{t-\frac{n-1}{2}}, \dots, x_t, \dots ,x_{t+\frac{n-1}{2}}$ in order to predict a motion mask $\hat{y}_t$ for the input frame $x_t$, the first and last $\frac{n-1}{2}$ frames of a scene do not get considered when evaluating a scene.

\subsection{Tooling and Hardware}
\label{ExpToolingHardware}
The MOSNET is implemented in Python 3.6 using the deep learning framework Keras 2.3.0 with TensorFlow 1.14 as a backend. The training and evaluation of the model is done with an Nvidia Titan V \ac{GPU} using CUDA 10.1. 

\subsection{Thresholding}
\label{ExpThresholding}

The trained MOSNET model outputs a 2D probability motion mask, where the probability of each pixel belonging to a moving object is mapped. Since the last layer of the MOSNET uses sigmoid as an activation function, the values of the probability motion mask ranges from zero to one. In order to generate a binary motion mask, where each pixel is labeled belonging either to a moving object or remaining static, a threshold is implemented.\\

Before implementing a threshold, the probability motion mask is smoothed using a Gaussian filter with a kernel size of $3 \times 3$ in order to perform an outlier reduction. 

Using a global threshold $T \in [0,1]$ maps a pixel of the probability motion mask $pr \in [0,1]$ on a thresholded pixel of the motion mask $t \in \{0,1\}$. 

\[ t =
  \begin{cases}
    1   & \quad \text{if } pr \geq T\\
    0     & \quad \text{otherwise}
  \end{cases}
\]

\begin{minipage}{0.55\textwidth}
\includegraphics[width=\textwidth]{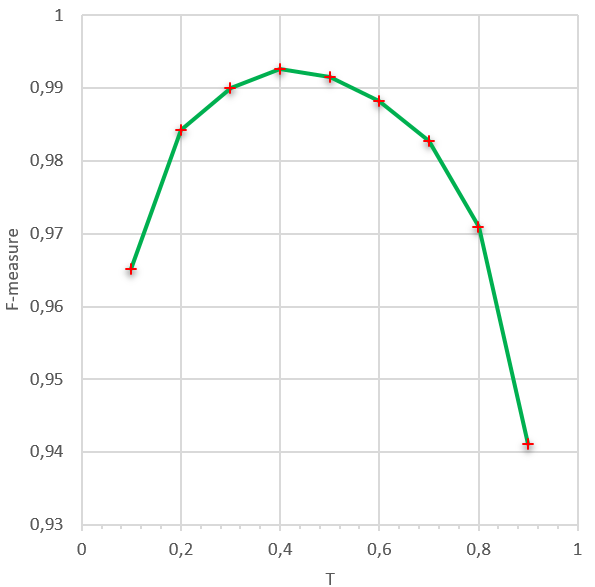}
\captionof{figure}[Plotted F-measure for different values of $T$]{Plotted F-measure for different values of $T$, evaluated on the training data $D_T$ of the CDNet2014 dataset.}
\label{img:threshold}
\end{minipage}
\hfill
\begin{minipage}{0.40\textwidth}
The optimum global threshold $T$ is determined by applying multiple thresholds on predicted motion masks and evaluate the resulting F-measure. Figure \ref{img:threshold} shows the resulting F-measure for different values of $T$, evaluated on the data $D_T$ and $D_V$. The peak value of the concave function between $T = 0.1$ and $T = 0.9$ plotted in Figure \ref{img:threshold} can be found at $T = 0.4$. This global threshold is implemented in order to transform probability motion masks into binary motion masks, which can be evaluated with metrics based on the confusion matrix introduced in Section \ref{EvalMetrics}.
\end{minipage}
\FloatBarrier

\section{Ablation Study on the MOSNET's Architecture}
\label{ExpMOSNET}

This section states design decisions which led to the architecture introduced in Chapter \ref{Architecture}. Unless mentioned otherwise, the experimental setup, described in Section \ref{ExpSetup} is used.

\subsection{Capturing Spatio-Temporal Features in Early Layers}
\label{ExpLLConv3D}

The aim of the Low-Level Conv3D network discussed in Section \ref{LLConv3D} is to generate spatio-temporal features with a low level of abstraction but a high resolution. This is intended to get a higher spatial accuracy of the contours of the moving objects, and simultaneously capturing more temporal content. Figure \ref{fig:VergleichLLConv3D} shows a spider chart of the F-measure obtained in each category of the CDNet2014 dataset with and without implementing the Low-Level Conv3D network. The architecture including the Low-Level Conv3D network is introduced in Chapter \ref{Architecture}, while the network architecture without the Low-Level Conv3D network is appended in Appendix \ref{AppohneLLConv3D}. The overall F-measure increases from 0.323 to 0.803 when implementing the MOSNET with the Low-Level Conv3D network. \\  

\begin{figure}[h!]
\centering
\includegraphics[width=0.73\textwidth]{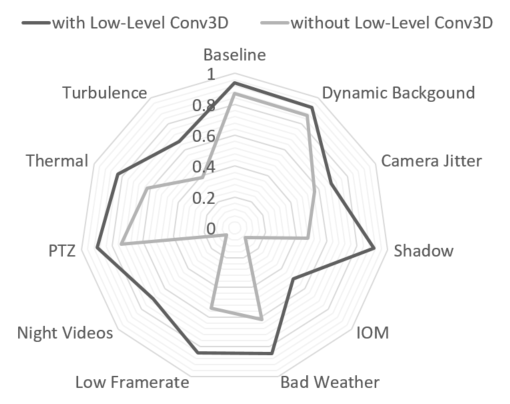}
\caption [F-measure of the MOSNET with and without the Low-Level Con3D network]{F-measure of the MOSNET with and without the Low-Level Con3D network.}
\label{fig:VergleichLLConv3D}
\end{figure}
\FloatBarrier

\subsection{Feature Map Fusion Technique}
\label{ExpFeaturemapfusing}

\begin{figure}[h!]
\centering
\includegraphics[width=0.85\textwidth]{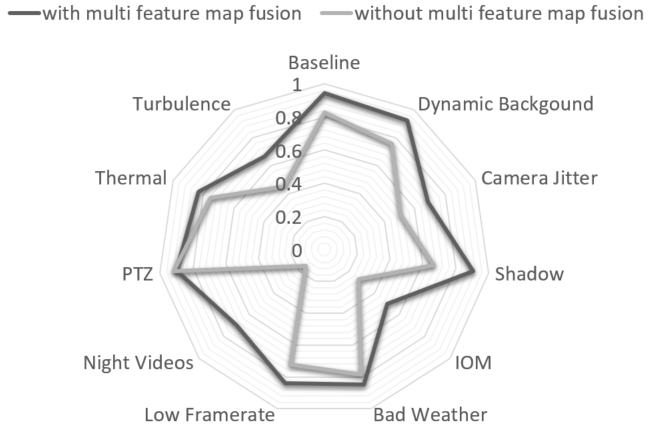}
\caption [F-measure of the MOSNET with and without feature map fusion]{F-measure of the MOSNET with and without feature map fusion evaluated on each category of the CDNet2014 dataset.}
\label{fig:VergleichMulti}
\end{figure}

The EC2 network concatenates feature maps generated by the layers MaxPooling\_2, MaxPooling\_3, Conv2D\_7, Dropout\_3, and Conv2D\_12. Each of the five layers generates feature maps with a different level of abstraction, a different resolution and a different receptive field regarding the input of the network. Fusing them aims at not finally defining these characteristics of the network by the architecture, but at enabling the network to learn an appropriate trade-off by itself. Figure \ref{fig:VergleichMulti} shows a comparison of the MOSNET with and without implementing the multi feature map fusion using the F-measure for each category of the CDNet2014 dataset. While the MOSNET with multi feature map fusion is introduced in Chapter \ref{Architecture}, the MOSNET without multi feature map fusion has a modified EC2 network, whose architecture is appended in Appendix \ref{AppE2ohnemulti}. The overall F-measure increases by $53.5\%$ from 0.523 to 0.803 when implementing a multi feature map fusion.\\
\FloatBarrier

\subsection{Using Batch Normalization}
\label{ExpUsingBatchNorm}

\begin{figure}[h!]
\centering
\includegraphics[width=\textwidth]{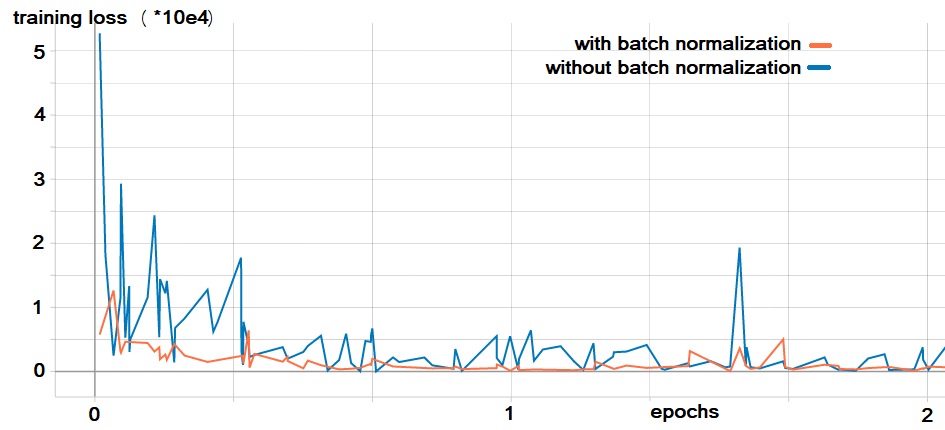}
\caption [Effect of batch normalization on the training loss]{Effect of batch normalization on the training loss.}
\label{fig:train_loss_batchnorm}
\end{figure}

Batch normalization as introduced in Section \ref{BatchNorm} aims at accelerating training. Besides this, batch normalization has a regularization effect \cite{batchnorm}. The impact on the training loss\footnote{The loss represents an inaccuracy calculated by the objective function, and thus is unit-less.} of using batch normalization is shown in Figure \ref{fig:train_loss_batchnorm}. Applying batch normalization leads to a smoother training loss, which is on average smaller than the training loss without batch normalization.  

\FloatBarrier

This leads to a faster convergence of the validation loss and therefore to an earlier stopping of the training process due to the Early Stopping technique (see Section \ref{monitoring}).
Table \ref{tab:batchnorm} shows the effect of applying batch normalization on the number of epochs and the resulting F-measure.

{\rowcolors{2}{gray!3!}{gray!30!}
\begin{table}[h!]
\centering
\caption{Using batch normalization to accelerate training.}
\label{tab:batchnorm}
\begin{tabular}{ccc} \toprule
                            & \textbf{number of epochs} & \textbf{F-measure} \\ \midrule
without batch normalization &       23                    &  0.798                   \\
with batch normalization    &       12                    &  0.803                \\ \bottomrule 
\end{tabular}
\end{table}
}
\FloatBarrier

\subsection{Optimizing the Hyperparameter n}
\label{ExpNLookback}

The fundamental hyperparameter of the MOSNET is the number of input frames $n$ fed to the network (see Section \ref{MOSNETOverview}). Increasing $n$ leads to more temporal context but raises the number of trainable parameters significantly due to the increase of the number of EC1 and EC2 networks needed. Furthermore, an increase of $n$ limits the real-time capability of the approach, since the delay of the approach is $(\frac{n-1}{2})$ frames \footnote{The calculation time of the network is neglected.} (see also Equation \eqref{eq:MOSNET}).\\

A trade-off has to be made, while $n = 1$ does not provide any temporal context, and a hyperparameter $n \geq 9$ is not considered for the reasons stated above. Table \ref{tab:nparameter} shows the impact of the hyperparameter $n$ on the trainable parameters and the resulting F-measure of the MOSNET. Setting $n$ to five leads to the best trade-off between temporal context and capacity of the network with an overall F-measure of $0.803$. 

{\rowcolors{2}{gray!3!}{gray!30!}
\begin{table}[h!]
\centering
\caption{Impact of $n$ on the number of trainable parameters and the resulting F-measure.}
\label{tab:nparameter}
\begin{tabular}{ccc}
\toprule
\textbf{n} & \textbf{trainable parameters} & \textbf{F-measure}\\ \midrule
1          & 9.0 Mio.     &    -  \\ 
3          & 14.2 Mio.    &    0.770   \\ 
5          & 19.2 Mio.    &    0.803 \\ 
7          & 24.5 Mio.    &    0.678   \\ 
9          & 29.8 Mio.    &    -   \\ \bottomrule
\end{tabular}
\end{table}
}
\FloatBarrier

\begin{figure}[h!]
\centering
\includegraphics[width=\textwidth]{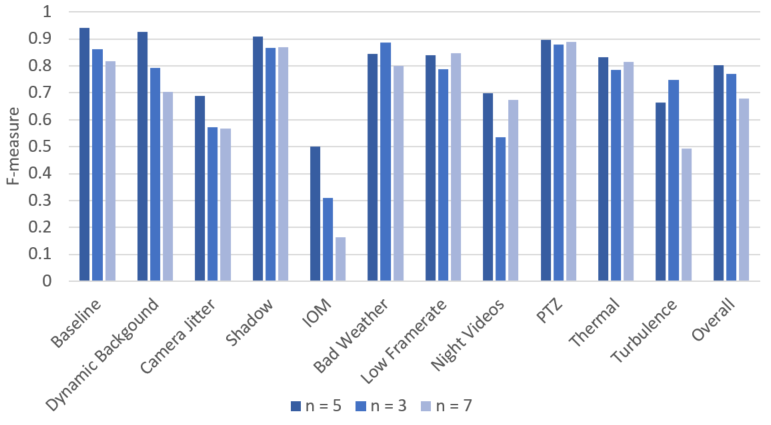}
\caption [Comparison of using three different values for the hyperparameter $n$]{Comparison of using three different values for the hyperparameter $n$.}
\label{fig:VergleichnInputs}
\end{figure}

Figure \ref{fig:VergleichnInputs} compares the F-measure of the MOSNET on each category of the CDNet2014 dataset, when setting the hyperparameter $n$ to three, five and seven, respectively.

\FloatBarrier

\begin{figure}[h!]%
  \centering
  \subfloat[][Input]{\includegraphics[width=0.2\textwidth]{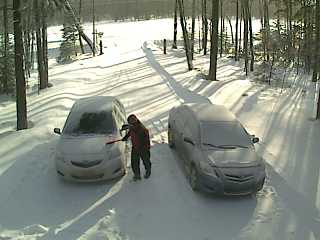}}%
  \qquad
  \subfloat[][$n = 5$]{\includegraphics[width=0.2\textwidth]{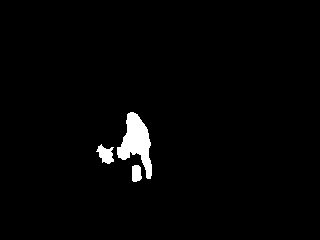}}%
  \qquad
  \subfloat[][$n = 3$]{\includegraphics[width=0.2\textwidth]{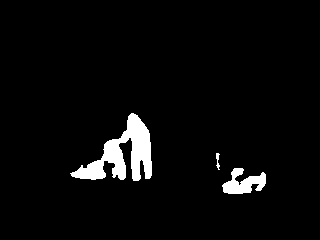}}%
  \qquad
  \subfloat[][$n = 7$]{\includegraphics[width=0.2\textwidth]{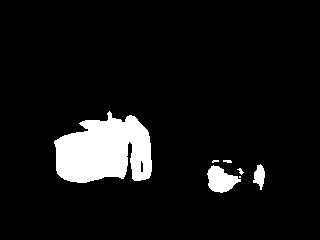}}%
  \caption[Example input frame of the \textit{WinterDriveWay} scene and predictions generated with different hyperparameter settings for $n$]{Example input frame of the \textit{WinterDriveWay} scene (a). Corresponding predictions of the MOSNET with the hyperparameter $n$ set to five (b), three (c), and seven (d).}%
  \label{Expn7shit}
\end{figure}

Figure \ref{Expn7shit} shows an input frame (a) of the \textit{WinterDriveWay} scene with two parked cars and a moving person representing the \ac{IOM} category of the $D_E$ data. Additionally three corresponding predictions of the MOSNET with five (b), three (c), and seven (d) input frames are shown. For $n = 3$ and $n = 7$, the resulting network capacity\footnote{Represented by the number of trainable parameters.} and the different temporal context leads to a worse prediction, where the network presumably starts learning saliency detection \cite{Saliency} and semantic segmentation \cite{SemanticSurvey} rather than motion segmentation.

\FloatBarrier

\section{Different Training Strategies}
\label{Exptraining}

In this section, different training strategies and their difference in performance get shown. If not stated otherwise, the experimental setup as described in Section \ref{ExpSetup} is used.

\subsection{Weighted Objective Function}
\label{ExpFocalLoss}

A common objective function for pixel-wise binary classification tasks, e.g. \cite{Lim, Lim_v2, LimS, Akilan, XuWen} is the binary cross-entropy discussed in Section \ref{lossfunction}. Since the ratio of pixels belonging to moving objects and pixels being static is approximately $1\!:\!38$ in the training data $D_T$, the weighted objective function called focal loss is used to train the MOSNET. The F-measure increases from $0.519$ when using binary cross-entropy to $0.803$ when using focal loss.

\begin{figure}[h!]
\centering
\includegraphics[width=0.73\textwidth]{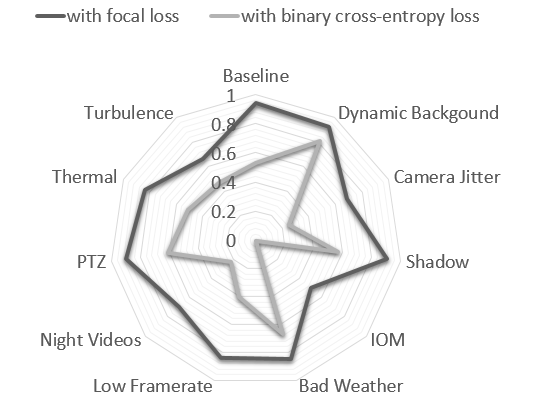}
\caption [F-measure of the MOSNET, trained with focal loss and with binary cross-entropy objective function]{F-measure of the MOSNET, trained with focal loss and with binary cross-entropy objective function.}
\label{fig:VergleichLLConv3D}
\end{figure}
\FloatBarrier

\subsection{Performance Enhancement due to Pre-Processing}
\label{ExpPreprocessing}

\begin{figure}[h!]
\centering
\includegraphics[width=0.7\textwidth]{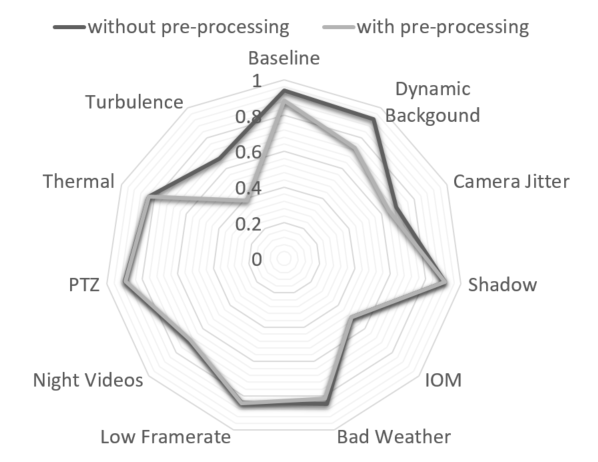}
\caption [Evaluation of the MOSNET with and without using pre-processing]{MOSNET with and without adding pre-processing, evaluated on each category of the CDNet2014 dataset using the F-measure.}
\label{fig:VergleichPrePro}
\end{figure}

Section \ref{Preprocessing} introduced feature based image alignment as a possible pre-processing method, which aims at boosting the performance of the approach especially in video sequences captured by a moving camera. Therefore, the F-measure for the categories captured with static camera should not change significantly while the F-measure for the scenes \textit{\ac{PTZ}} and \textit{Camera Jitter} should improve. Figure \ref{fig:VergleichPrePro} shows a spider chart comparing the F-measure of the MOSNET with and without additional pre-processing. While there is no significant change in F-measure for most of the categories, especially the categories \textit{Turbulence} and \textit{Dynamic Background} have a significant decrease in the F-measure of $41.81\%$ and $20.74\%$, respectively. The overall F-measure decreases from $0.803$ to $0.715$, when adding feature based image alignment as pre-processing.\\  

\FloatBarrier

\begin{figure}[h!]%
  \centering
  \subfloat[][Dynamic Background]{\includegraphics[width=0.3\textwidth]{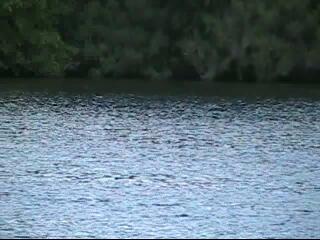}}%
  \qquad
  \subfloat[][Turbulence]{\includegraphics[width=0.3\textwidth]{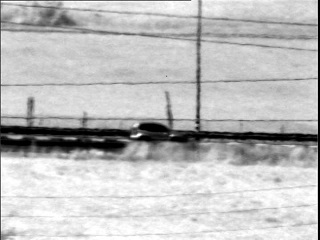}}%
  \caption[Example input frame of the \textit{Dynamic Background} and \textit{Turbulence} category]{Example input frame of the \textit{Dynamic Background} (a) and \textit{Turbulence} category (b) of the data $D_E$.}%
  \label{Exppreproinput}
\end{figure}

Figure \ref{Exppreproinput} shows an input frame of the \textit{Dynamic Background} and \textit{Turbulence} category. Since in the data $D_E$ both categories are represented by scenes, in which a large part of the input frames contain irrelevant coherent motion originated from waves and heat turbulences, the feature based image alignment does not work properly. The feature detector as a part of the pre-processing finds many strong features with a coherent motion in cohesive input frames. Using these features for image alignment leads to alignment errors. Since the scene representing the \textit{\ac{PTZ}} category does not involve continuous camera movement, but a scene where the camera is static most of the time and only performs quick pans in order to change the field of view, the F-measure of the \textit{\ac{PTZ}} category does not change significantly. Since this scenes covers just a small field of challenges occuring in videos captured by \ac{PTZ}-cameras, the effect of adding pre-processing is further evaluated on the LASIESTA dataset in Section \ref{ExpCrossEval}.\\

\FloatBarrier

Since the overall F-measure decreases by $11\%$, feature based image alignment pre-processing is not used when evaluating the MOSNET with the CDNet2014 dataset. To show that pre-processing on scenes where the camera experiences continuous movement brings improvement, a cross-evaluation with the LASIESTA dataset is performed in Chapter \ref{ExpCrossEval}. 

\chapter{Evaluation}
\label{Eval}

If not stated otherwise, the evaluation setup is equivalent to the experimental setup introduced in Section \ref{ExpSetup}. Therefore, the MOSNET evaluated in Section \ref{Comparisson} is an end-to-end deep learning method, without any pre-processing. The tooling and hardware used for evaluation is the same as introduced in Section \ref{ExpToolingHardware}.\\

Figure \ref{PRcurve} shows the precision-recall curve of the MOSNET for thresholds ranging from $0.0$ to $0.9$. A threshold of $T=1.0$ is excluded, since there are no pixels with a probability of $1.0$ in the predicted motion masks. Figure \ref{EvalFMeasure} shows the corresponding F-measures, whereby the data-point generated by a threshold of $0.4$ is marked in red color in both figures. The precision-recall curve is asymmetrical to the first main diagonal and the optimum threshold is $T \neq 0.5$. This indicates an imbalance in precision and recall. The precision describes how many of the predicted pixels are actually moving, and the recall indicates how many of the actually moving pixels are predicted as being in motion The MOSNET does not predict a lot of static pixels as being in motion but tends to miss moving pixels. This is also apparent in the imbalance of the \ac{FPR} and the \ac{FNR} of $0.003$ and $0.269$, respectively. Therefore, the approach is suited for applications, which aims at minimizing false segmentations rather than aiming at minimizing missed segmentations.

\begin{minipage}{0.48\textwidth}
\includegraphics[width=\textwidth]{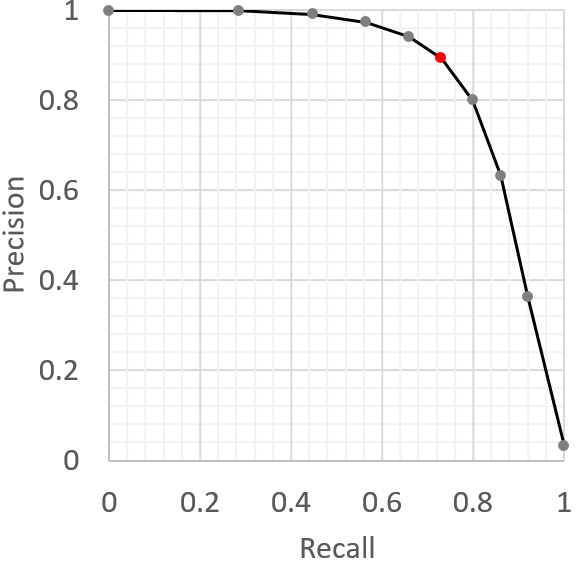}
\captionof{figure}[\mbox{Precision-recall} curve, generated using the $D_E$ split of the CDNet2014 dataset]{\mbox{Precision-recall} curve, generated using the $D_E$ split of the \mbox{CDNet2014} dataset.}
\label{PRcurve}
\end{minipage}
\hfill
\begin{minipage}{0.48\textwidth}
\includegraphics[width=\textwidth]{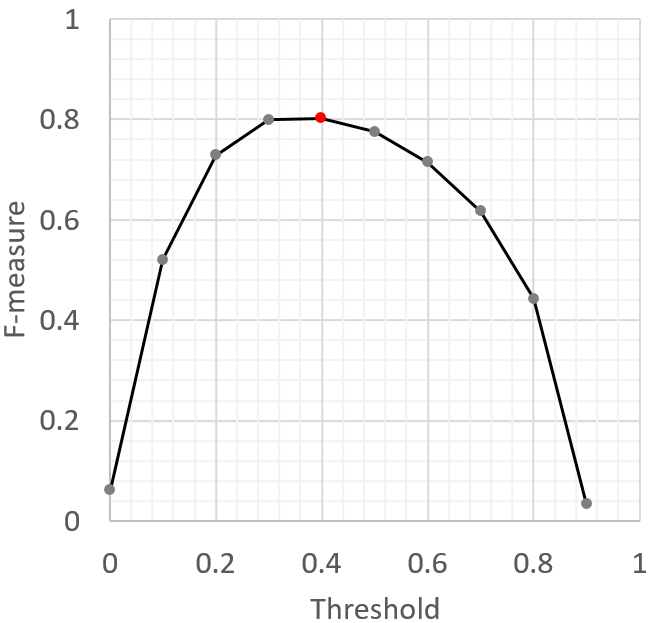}
\captionof{figure}[F-measure for different values of the threshold $T$]{F-measure for different values of the threshold $T$, generated using the $D_E$ split of the CDNet2014 dataset.}
\label{EvalFMeasure}
\end{minipage}
\FloatBarrier

\begin{minipage}{0.56\textwidth}
Figure \ref{OverallMetric} shows the precision, recall and F-measure for a threshold $T$ of $0.4$ in detail. The MOSNET with reaches an overall F-measure of $0.803$ on the evaluation data $D_E$, and a \ac{PWC} of $1.138\%$. Figure \ref{fig:f_ped} to Figure \ref{fig:f_therm} show example frames of the data $D_E$ with the corresponding predicted motion in green color. The following sections bring the performance of the MOSNET in line with the related work.
\end{minipage}
\hfill
\begin{minipage}{0.4\textwidth}
\includegraphics[width=\textwidth]{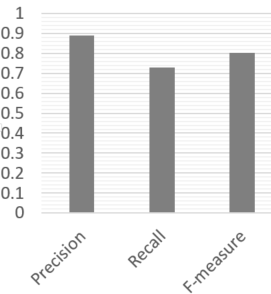}
\captionof{figure}[Precision, recall, and F-measure of the MOSNET]{Precision, recall, and F-measure of the MOSNET, evaluted on the data $D_E$ with a threshold $T= 0.4$.}
\label{OverallMetric}
\end{minipage}
\FloatBarrier

\begin{minipage}{0.4\textwidth}
\includegraphics[width=\textwidth]{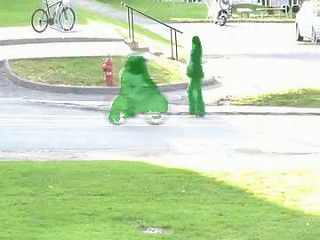}
\captionof{figure}[Example frame with predicted motion of the \textit{pedestrians} scene]{Example frame with predicted motion of the $pedestrians$ scene of $baseline$ category.}
\label{fig:f_ped}
\end{minipage}
\hfill
\begin{minipage}{0.4\textwidth}
\includegraphics[width=\textwidth]{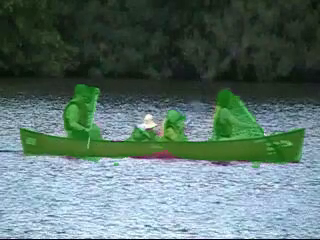}
\captionof{figure}[Example frame with predicted motion of the \textit{canoe} scene]{Example frame with predicted motion of the $canoe$ scene of the \textit{dynamic background} category.}
\label{fig:f_dyn}
\end{minipage}
\FloatBarrier

\begin{minipage}{0.4\textwidth}
\includegraphics[width=\textwidth]{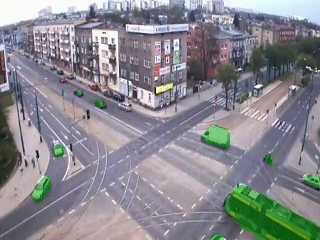}
\captionof{figure}[Example frame with predicted motion of the \textit{tram\_crossroad\_1fps} scene]{Example frame with predicted motion of the \textit{tram\_crossroad\_1fps} scene of the \textit{low framerate} category.}
\label{fig:f_low}
\end{minipage}
\hfill
\begin{minipage}{0.4\textwidth}
\includegraphics[width=\textwidth]{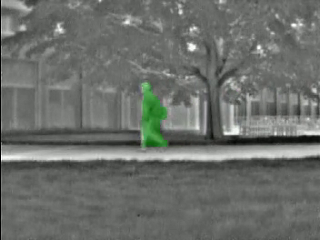}
\captionof{figure}[Example frame with predicted motion of the \textit{park} scene]{Example frame with predicted motion of the \textit{park} scene of the \textit{thermal} category.}
\label{fig:f_therm}
\end{minipage}
\FloatBarrier

\section{Comparison to Early Approaches}
The na{\"i}ve approach of using dense optic flow without any pre- and post-processing for segmenting moving objects results in a poor performance on the CDNet2014 dataset. The algorithm of Kröger et al. \cite{DISopticalflow} leads to an overall F-measure of $0.085$, while the algorithm of Farneback \cite{Farneback} leads to an overall F-measure of $0.127$. Figure \ref{fig:highwayF} and Figure \ref{fig:highwayD} show exemplary results of using both algorithms to segment moving objects in scenes of the CDNet2014 dataset. The MOSNET beats both algorithms with an overall F-measure of $0.803$.\\

\begin{minipage}{0.40\textwidth}
\includegraphics[width=0.9\textwidth]{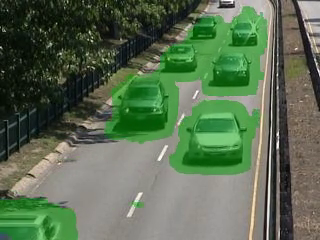}
\captionof{figure}[Example frame of using the Farneback algorithm on the $highway$ scene]{Using the algorithm of Farneback \cite{Farneback} to detect moving objects on the CDNet2014 scene $highway$.}
\label{fig:highwayF}
\end{minipage}
\hfill
\begin{minipage}{0.40\textwidth}
\includegraphics[width=0.9\textwidth]{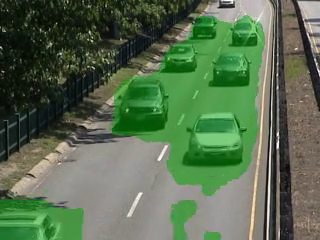}
\captionof{figure}[Example image of using DIS Optical Flow on the $highway$ scene]{Using DIS Optical Flow algorithm of Kröger et al. \cite{DISopticalflow} to detect moving objects on the CDNet2014 scene $highway$.}
\label{fig:highwayD}
\end{minipage}
\FloatBarrier

\section{Comparison to the State-of-the-Art}
\label{Comparisson}

The CDNet2014 dataset does not provide a unified evaluation protocol and therefore no pre-defined data split.
This impedes an unambiguous comparison with the state-of-the-art introduced in the Chapter \ref{RelatedWork}. The fact that no uniform evaluation split $D_E$ is specified for the CDNet2014 dataset leads to some methods evaluating in an \ac{SDE} and some in an \ac{SIE} manner\footnote{In an \ac{SDE} the data used during development and the data used during evaluation may originate of the same scene. In an \ac{SIE} the data used to develop and the data used to evalute the approach do not originate of the same scene (see Section \ref{Datasetsplit}).} (see Table \ref{tab:OverallF1}).\\

In order to achieve a better assessment of the MOSNET's performance into the state-of-the-art, the performance is first compared with the methods listed in the CDNet2014 ranking in Section \ref{CompCDNet}, and then with state-of-the-art methods that use \ac{SIE} in Section \ref{CompSIE}.

\subsection{Comparison to the CDNet2014 ranking}
\label{CompCDNet}

Table \ref{tab:OverallF1} shows the ten best ranked approaches of the CDNet2014 ranking (Accessed: \today), and points out whether \ac{SIE} is used or not. Except IUTIS-3, IUTIS-5 \cite{Bianco}, and SWCD \cite{SWCD}, all method in Table \ref{tab:OverallF1} are based on deep learning. IUTIS-3 and IUTIS-5 are frameworks, which aim at boosting the overall performance by combining several approaches. SemanticBGS is a framework, which combines any background subtraction or motion segmentation method with the information of a semantic segmentation algorithm. Therefore, IUTIS-3, IUTIS-5, and SemanticBGS are not based on a particular network architecture, but are rather frameworks to improve a developed deep learning approach. Discussing such frameworks in combination with the MOSNET is part of Chapter \ref{Conclusion}.\\

{\rowcolors{2}{gray!3!}{gray!30!}
\begin{table}[h!]
\begin{center}
\caption[Methods with the highest overall F-measure on the CDNet2014-dataset]{Methods with the highest overall F-measure on the CDNet2014-dataset \cite{CDNet2014} (Accessed: \today).}
\begin{tabular}{ccc}
\toprule 
\rule[-1ex]{0pt}{4ex}\textbf{Method} & \textbf{F-measure} & \textbf{\ac{SIE}}\\ 
\midrule
\rule[-1ex]{0pt}{4ex}FgSegNet\_v2 \cite{Lim_v2} & 0.985 & -\\
\rule[-1ex]{0pt}{4ex}FgSegNet\_S \cite{LimS} & 0.980 & -\\ 
\rule[-1ex]{0pt}{4ex}FgSegNet\_M \cite{Lim}& 0.977 & -\\ 
\rule[-1ex]{0pt}{4ex}BSPVGAN \cite{BSPVGAN} & 0.950 & -\\ 
\rule[-1ex]{0pt}{4ex}BSGAN \cite{BSGAN}& 0.934 & -\\ 
\rule[-1ex]{0pt}{4ex}Cascade CNN \cite{CascadeCNN} & 0.921 & -\\ 
\rule[-1ex]{0pt}{4ex}SemanticBGS \cite{Braham2} & 0.789 & \checkmark\\ 
\rule[-1ex]{0pt}{4ex}IUTIS-5 \cite{Bianco} & 0.772 & \checkmark\\ 
\rule[-1ex]{0pt}{4ex}SWCD \cite{SWCD} & 0.758 & \checkmark\\ 
\rule[-1ex]{0pt}{4ex}IUTIS-3 \cite{Bianco} & 0.755 & \checkmark\\ 
\bottomrule
\end{tabular}
\label{tab:OverallF1}
\end{center}
\end{table}
}
\FloatBarrier

\begin{figure}[h!]
\centering
\includegraphics[width=\textwidth]{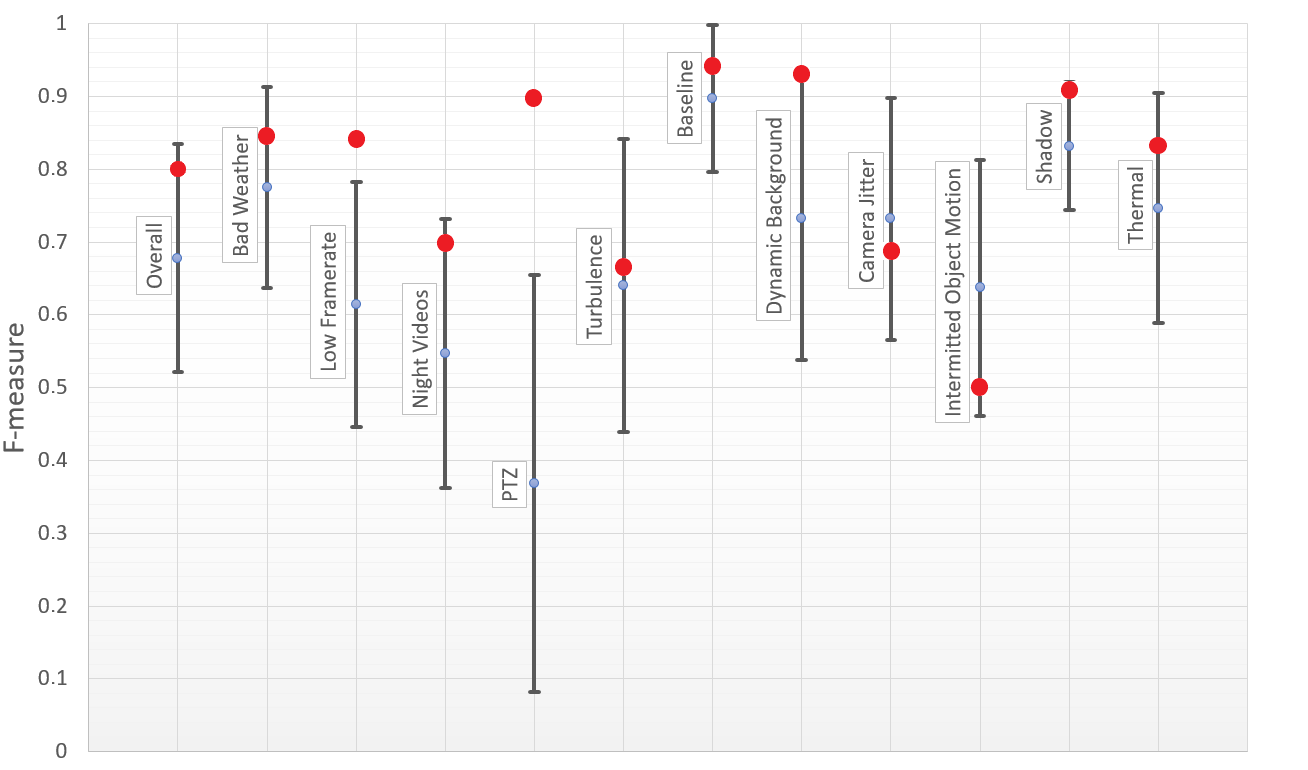}
\caption [Mean and standard deviation of F-measures of the 40 best ranked algorithms]{Mean and standard deviation of F-measures of the 40 best ranked algorithms across all categories of the CDNet2014-dataset. The F-measure, obtained by the MOSNET in each category is marked with \textcolor{red}{$\bullet$}.}
\label{fig:FmeasuresOverview_MOSNET}
\end{figure}
\FloatBarrier

\subsection{Comparison to State-of-the-Art Methods that use SIE}
\label{CompSIE}

For a better comparison of methods of the related work (see Section \ref{RelatedWork}), which use \ac{SIE}, Table \ref{tab:F1_SIE} lists six approaches evaluated with \ac{SIE}.  Ozan et al., who introduced the BSUV-net, trained the FgSegNet\_v2 anew in a \ac{SIE} manner, using the same data split as they used for the BSUV-net \cite{BSUV}. Since there is no uniform evaluation protocol for the CDNet2014, which defines which data is used for evaluation, each approach in Table \ref{tab:F1_SIE} uses different evaluation data.

{\rowcolors{2}{gray!3!}{gray!30!}
\begin{table}[h!]
\centering
\caption[F-measure in the CDNet2014 dataset with methods using SIE]{F-measure in the CDNet2014 dataset with methods, using \ac{SIE}. The best result for each category is written in \textbf{bold}. The numbers originate from the papers \cite{BSUV, SWCD, Bianco, 3DFR} and are rounded to three decimal places.}
\label{tab:F1_SIE}
\begin{tabular}{ccccccc}
\toprule
\textbf{Category}                                            & \textbf{BSUV-net} & \textbf{SWCD} & \textbf{IUTIS-5} & \textbf{3DFR} & \textbf{FgSegNet\_2} & \textbf{MOSNET} \\ \midrule
\begin{tabular}[c]{@{}c@{}}Bad \\ Weather\end{tabular}       & 0.871             & 0.823         & 0.825            & \textbf{0.950}         &  0.328               & 0.846\\ 
\begin{tabular}[c]{@{}c@{}}Low\\ Framerate\end{tabular}      & 0.680             & 0.737         & 0.774            & \textbf{0.920}         &  0.245               & 0.841\\ 
\begin{tabular}[c]{@{}c@{}}Night\\ Videos\end{tabular}       & 0.699             & 0.581         & 0.529            & \textbf{0.790}         &  0.280               &0.698\\ 
PTZ                                                          & 0.628             & 0.455         & 0.4282           & -             &  0.350               &\textbf{0.898}\\ 
Thermal                                                      & 0.858             & 0.858         & 0.830            & \textbf{0.890}         &  0.604               &0.833\\ 
Shadow                                                       & \textbf{0.923}             & 0.878         & 0.908            & 0.790         &  0.530               &0.908\\ 
IOM                                                          & 0.750             & 0.709         & 0.730            & \textbf{0.870}          &  0.200               &0.500\\ 
\begin{tabular}[c]{@{}c@{}}Camera\\ Jitter\end{tabular}      & 0.774             & 0.741         & \textbf{0.833}            & 0.650         &  0.427               &0.688\\ 
\begin{tabular}[c]{@{}c@{}}Dynamic\\ Background\end{tabular} & 0.797             & 0.865         & 0.890            & 0.900         &  0.363               &\textbf{0.926}\\ 
Baseline                                                     & \textbf{0.969}             & 0.921         & 0.957            & 0.930         &  0.693               & 0.941\\ 
Turbulence                                                   & 0.705             & 0.774         & 0.783            & \textbf{0.860}         &  0.064               & 0.665\\ \midrule
Overall                                                      & 0.787             & 0.758         & 0.772            & \textbf{0.860}          &  0.372               & 0.803\\ \bottomrule
\end{tabular}
\end{table}
}

\section{Cross-Evaluation with the LASIESTA dataset}
\label{ExpCrossEval}

In a cross-evaluation, the data $D_T$ and $D_E$ originate of different datasets. In order to show its capability to generalize, a cross-evaluation of the MOSNET is performed by using the datasplit $D_T$ of the CDNet2014 for training and using the LASIESTA dataset for evaluation (see Section \ref{LASIESTA}). Furthermore, the LASIESTA dataset consists of multiple scenes with a different intensity levels of camera motion. This enables to evaluate the MOSNET's performance on scenes captured by a moving camera in more detail.\\

\begin{figure}[h!]
\centering
\includegraphics[width=0.75\textwidth]{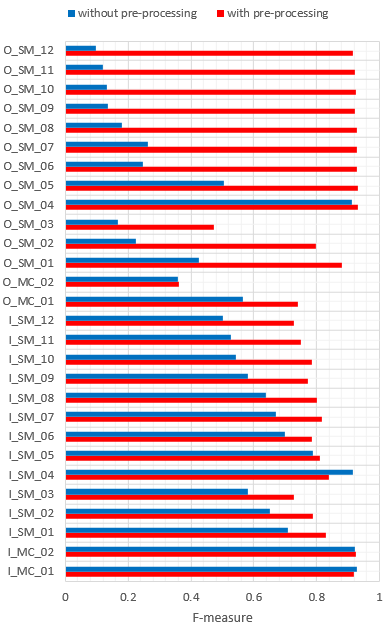}
\caption [Cross-evaluation for scenes of the LASIESTA dataset, captured by a moving camera]{Cross-evaluation for scenes of the LASIESTA dataset captured by a moving camera. Comparing the resulting F-measure of the MOSNET with and without pre-processing.}
\label{fig:LASIESTAMOVING}
\end{figure}

Figure \ref{fig:LASIESTAMOVING} shows the result of the cross-evaluation of scenes, captured by a moving camera. The scenes O\_MC\_01, O\_MC\_02, I\_MC\_01, I\_MC\_02, O\_SM\_01, O\_SM\_02, O\_SM\_03, I\_SM\_01, I\_SM\_02, and I\_SM\_03 are captured with a continuous panning and titling camera, while the other scenes in Figure \ref{fig:LASIESTAMOVING} contain camera jitter, which gets stronger as the scene number goes up. A description of the scene names of the LASIESTA dataset is given in Section \ref{LASIESTA}. 

\FloatBarrier

The larger the camera movement, the greater the improvement that can be achieved by pre-processing. The best improvement can be observed in the O\_SM\_12 scene with a factor of approximately ten. To get a visual impression, an example frame of the O\_SM\_12, predicted with and without pre-processing is shown in Figure \ref{fig:SM_ohne} and Figure \ref{fig:SM_mit}.\\

\begin{minipage}{0.4\textwidth}
\includegraphics[width=\textwidth]{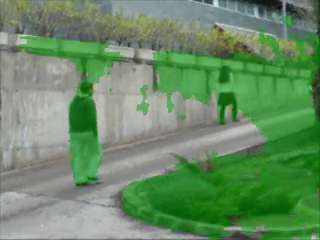}
\captionof{figure}[Example frame of the O\_SM\_12 scene without pre-processing]{Example frame of the O\_SM\_12 scene of the LASIESTA dataset without pre-processing.}
\label{fig:SM_ohne}
\end{minipage}
\hfill
\begin{minipage}{0.4\textwidth}
\includegraphics[width=\textwidth]{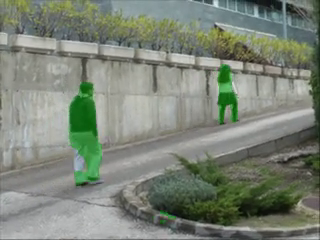}
\captionof{figure}[Example frame of the O\_SM\_12 scene with pre-processing]{Example frame of the O\_SM\_12 scene of the LASIESTA dataset with pre-processing.}
\label{fig:SM_mit}
\end{minipage}
\FloatBarrier

\begin{figure}[h!]
\centering
\includegraphics[width=0.66\textwidth]{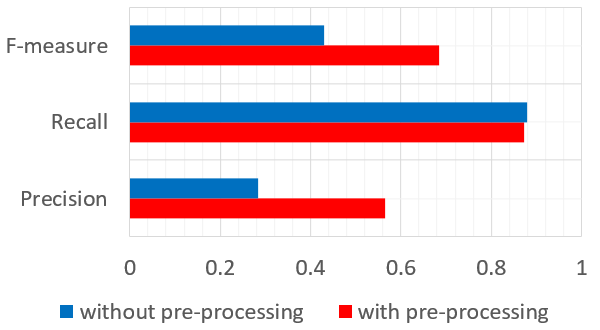}
\caption [Cross-evaluation on all scenes of the LASIESTA dataset]{Cross-evaluation on all scenes of the LASIESTA dataset. Comparing the resulting precision, recall, and F-measure of the MOSNET with and without pre-processing.}
\label{fig:LASIESTAOVERALL}
\end{figure}

Evaluating the MOSNET trained on the training data $D_T$ of the CDNet2014 dataset and evaluated on all scenes of the LASIESTA dataset leads to the results shown in Figure \ref{fig:LASIESTAOVERALL}. The recall does not change significantly when adding pre-processing, which indicates that both times nearly the same amount of actual moving pixels are predicted as being in motion. Adding pre-processing changes the precision from $0.284$ to $0.564$, which shows that less of the predicted moving pixels are actually static pixels. This gain in precision leads to an improvement of the F-measure from $0.429$ to $0.685$ when adding pre-processing.

\FloatBarrier

\begin{figure}[h!]%
  \centering
  \subfloat[][I\_CA\_01]{\includegraphics[width=0.3\textwidth]{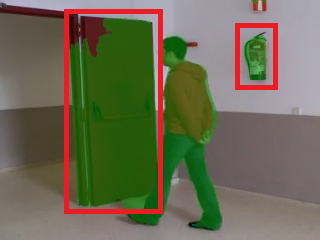}}%
  \qquad
  \subfloat[][O\_SU\_02]{\includegraphics[width=0.3\textwidth]{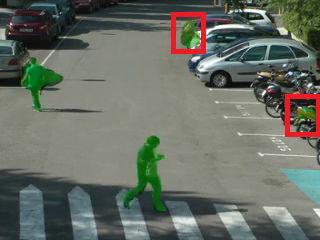}}%
  \qquad
  \subfloat[][O\_SN\_02]{\includegraphics[width=0.3\textwidth]{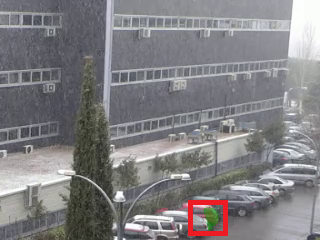}}%
  \caption[Cross-evaluating the MOSNET shows a tendency of mistakenly detecting static red colored objects]{Cross-evaluating the MOSNET shows a tendency of mistakenly detecting static red colored objects as being in motion. Images (a), (b), and (c) shows three examples of different scenes, where this effect can be observed.}%
  \label{LASIESTA_RED}
\end{figure}

Cross-evaluation with the LASIESTA dataset shows a tendency of the trained MOSNET to erroneously predicting red colored static objects for being in motion. Figure \ref{LASIESTA_RED} shows examples of three different scenes, where the predicted moving pixels belonging to red objects are marked with boxes.

\FloatBarrier

Furthermore, since the CDNet2014 dataset is also used for change detection it contains some scenes with static objects labeled as motion that start moving during the scene. As change detection tries to segment objects that change the appearance of the scene, this annotation is suitable.\\

\begin{minipage}{0.56\textwidth}
 In the case of motion segmentation, these moving objects, remain static for a certain period of time do not represent the task tackled in this thesis. Since the relabeled \mbox{LASIESTA} dataset does not contain moving objects that partially remain static during the sequence, salient objects are sometimes recognized as being in motion. Figure \ref{fig:red1} shows an example of the I\_MB\_02 scene, where the salient backpack is mistakenly predicted as motion.\\
Overcoming these problems shown in Figure \ref{LASIESTA_RED} and Figure \ref{fig:red1} is part of a potential future work and is further discussed in Chapter \ref{Conclusion}.
\end{minipage}
\hfill
\begin{minipage}{0.4\textwidth}
\includegraphics[width=\textwidth]{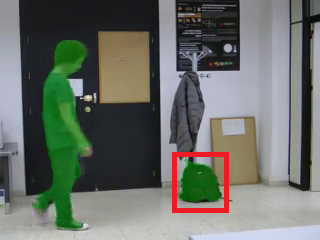}
\captionof{figure}[Example frame of the I\_MB\_02 scene]{Example frame of the I\_MB\_02 scene, which shows that the MOSNET tends to detect salient objects as being in motion.}
\label{fig:red1}
\end{minipage}
\FloatBarrier

\chapter{Conclusion and Future Work}
\label{Conclusion}

This thesis introduced a new end-to-end deep learning based approach that uses a set of five cohesive frames to segment motion in video sequences. Evaluating the developed approach on previously unseen scenes of the CDNet2014 dataset leads to an F-measure of $0.803$. Using the LASIESTA dataset for a cross-evaluation leads to an F-measure of $0.429$. When adding feature based image alignment as an optional pre-processing step, the performance on the LASIESTA dataset increases by about $60\%$ to a F-measure of $0.685$. Due to the small temporal window needed to obtain a prediction, the approach shows an outstanding performance at scenes captured by non-static cameras.\\ 

The network architecture combines an encoder-decoder neural network architecture with a set of 3D convolutions in two different parts of the network. This enables the capturing of spatio-temporal features with a low level of abstraction at high resolution, as well as spatio-temporal features with a high level of abstraction at low resolution. Additionally, a new multi feature map fusion technique enables the network a better handling of moving objects with a different size. Furthermore, implementing Batch Normalization to accelerate the training and a weighted objective function in order to tackle the class imbalance improves the performance of the developed method.\\

The deep learning based approach was compared with the state-of-the-art, by bringing it in line with the published ranking of the CDNet2014 dataset. The performance, which is significantly above average, beats all methods of the CDNet2014 dataset ranking that evaluate only on previously unseen scenes. Comparing the developed approach with further methods, which perform an evaluation on previously unseen scenes shows that the approach is state-of-the-art for such methods. To show the capability to generalize across datasets and handling of scenes captured by a moving camera, a cross-evaluation with the LASIESTA dataset was performed. In order to improve the performance of the deep learning approach on scenes with a strong camera movement, feature based image alignment as an optional pre-processing step was implemented.\\

Evaluating the developed method shows an imbalance in precision and recall. As a result, less false alarms or to be more precise, less \acp{FP} are generated. However, if the MOSNET shall become more sensitive, i.e. it shall not miss moving objects the recall must become higher. One way to do this is to insert further convolution layers in the encoder part of the network in order to enlarge the receptive field and handle moving objects which make up a large part of the video frame. Additionally, a post-processing step can be implemented, which combines the information of semantic segmentation and motion segmentation in order to improve the overall performance. A related approach by Ozan et al. \cite{BSUV} improved their F-measure by about $0.012$ by adding semantic information as initially introduced by Braham et al. \cite{Braham2}. When cross-evaluating the developed approach, it can be observed that salient static objects and especially static red colored objects are likely to be mistakenly segmented as being in motion. In order to tackle this, moving objects that remain static, which are labeled as motion in the CDNet2014 dataset, must be relabeled as being static when training the network to adapt the CDNet2014 dataset to the task of motion segmentation. Furthermore, pre-training the network with the DAVIS 2016 dataset \cite{Perazzi} could enhance the diversity in the semantic of moving objects and could be a way to prevent the network getting biased on a special attribute of occurring objects such as being red colored.

\bibliography{literatur}{}
\bibliographystyle{ieeetr}
\addcontentsline{toc}{chapter}{Bibliography}
\addcontentsline{toc}{chapter}{Appendix}
\newpage
\chapter*{Appendix}
\begin{appendix}

\chapter{Technical Foundations}

\section{Activation functions}
\label{App:Activation}
\begin{minipage}{0.45\textwidth}
\begin{tikzpicture}
\begin{axis}[
    axis lines=middle,
    xmax=10,
    xmin=-10,
    ymin=-0.05,
    ymax=1.05,
    xlabel={$z$},
    ylabel={$\sigma(z)$}
]
\addplot [domain=-9.5:9.5, samples=100,
          thick, blue] {1/(1+exp(-x)};

\end{axis}
\end{tikzpicture}
\end{minipage}
\hfill
\begin{minipage}{0.45\textwidth}
\begin{tikzpicture}
\begin{axis}[
    axis lines=middle,
    xmax=10,
    xmin=-10,
    ymin=-1.05,
    ymax=1.05,
    xlabel={$z$},
    ylabel={$\tanh(z)$}
]
\addplot [domain=-9.5:9.5, samples=100,
          thick, blue] {tanh(x)};

\end{axis}
\end{tikzpicture}
\end{minipage}
\FloatBarrier

\begin{minipage}{0.45\textwidth}
\begin{tikzpicture}
\begin{axis}[
    axis lines=middle,
    xmax=1,
    xmin=-1,
    ymin=-0.05,
    ymax=1.05,
    xlabel={$z$},
    ylabel={$ReLu(z)$}]
\addplot [domain=-5.5:5.5, samples=1000, thick, blue] {max(0, x)};
\end{axis}
\end{tikzpicture}
\end{minipage}
\hfill
\begin{minipage}{0.55\textwidth}
\begin{itemize}
    \item Upper left: Sigmoid function $\sigma(z) = \frac{1}{1-e^{-z}}$
    \item Upper right: Hyperbolic tangent function $tanh(z) =2\sigma(2z)-1$
    \item Lower left: Rectified Linear Unit $ReLu(z) = max(0,z)$
\end{itemize}{}
\end{minipage}
\FloatBarrier

\chapter{Network Architecture}

\section{Evolution of the receptive field in the encoder part of the MOSNET}
\label{Appevolution}

{\rowcolors{2}{gray!3!}{gray!30!}
\begin{table}[h!]
\centering
\caption[Evolution of the receptive field regarding the input of the network in the encoder of the MOSNET]{Evolution of the receptive field regarding the input of the network in the encoder of the MOSNET. Layers used for the fusion of multiple featue maps are written in \textbf{bold}.}
\label{tab:receptivefield}
\begin{tabular}{cc}
\toprule
\textbf{Layer} & \textbf{Size of receptive field}  \\ \midrule
Conv2D\_1      & $3 \times 3$                                         \\ 
Conv2D\_2      & $5 \times 5$                                            \\ 
MaxPooling\_1  & $6 \times 6$                                           \\ 
BatchNorm\_1   & $6 \times 6$                                           \\ 
Conv3D\_1      & $10 \times 10$                                        \\ 
Conv3D\_2      & $14 \times 14$                                          \\ 
BatchNorm\_2   & $14 \times 14$                                          \\ 
Conv2D\_3      & $18 \times 18$                                          \\ 
Conv2D\_4      & $22 \times 22$                                          \\ 
\textbf{MaxPooling\_2}  & \textbf{$24 \times 24$}                                          \\
BatchNorm\_3   & $24 \times 24$                                         \\ 
Conv2D\_5      & $32 \times 32$                                          \\ 
Conv2D\_6      & $40 \times 40$                                          \\ 
\textbf{Conv2D\_7}      & \textbf{$48 \times 48$}                                         \\ 
BacthNorm\_4   & $48 \times 48$                                       \\ 
Conv2D\_8      & $56 \times 56$                                          \\ 
Dropout\_1     & $56 \times 56$                                          \\ 
Conv2D\_9      & $64 \times 64$                                          \\ 
Dropout\_2     & $64 \times 64$                                          \\ 
Conv2D\_10     & $72 \times 72$                                         \\ 
\textbf{Dropout\_3}     & \textbf{$72 \times 72$}                                         \\ 
Conv2D\_11     & $80 \times 80$                                         \\ 
\textbf{Conv2D\_12}     & \textbf{$88 \times 88$}                                         \\
\textbf{MaxPooling\_3}     & \textbf{$16 \times 16$}                                          \\ \bottomrule
\end{tabular}
\end{table}
}

\chapter{Experiments and Results}

\section{MOSNET architecture without the Low-Level Conv3D network}
\label{AppohneLLConv3D}

\begin{figure}[h!]
\centering
\includegraphics[width=0.7\textwidth]{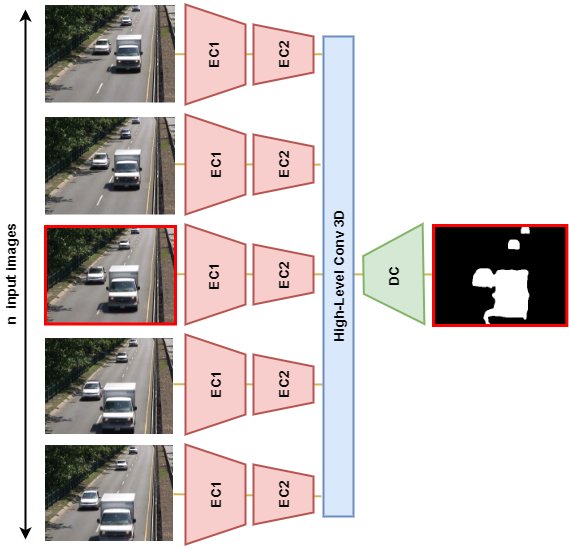}
\caption [Architecture of the MOSNET without the Low-Level Conv3D networks]{Architecture of the MOSNET without the Low-Level Conv3D network, used in Section \ref{ExpLLConv3D}.}
\label{fig:AppEC2architecture}
\end{figure}
\FloatBarrier

\section{EC2 network without multi feature map fusion}
\label{AppE2ohnemulti}

\begin{figure}[h!]
\centering
\includegraphics[width=0.34\textwidth]{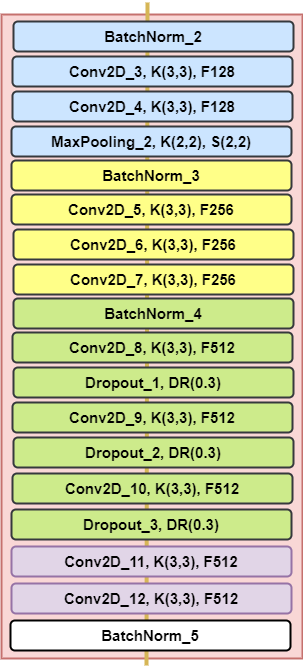}
\caption [EC2 network architecture without fusing multiple feature maps]{EC2 network architecture without fusing multiple feature maps, used in Section \ref{ExpFeaturemapfusing}.}
\label{fig:AppEC2architecture}
\end{figure}
\FloatBarrier

\end{appendix}

\end{document}